\documentclass[12pt,BSc,wordcount,oneside]{muthesis}

\usepackage{verbatim}
\usepackage{enumitem}
\usepackage{graphicx}
\usepackage{url}
\usepackage{listings}
\usepackage[colorlinks=true,linkcolor=blue,citecolor=blue]{hyperref}
\usepackage{makecell}
\usepackage{amssymb}
\usepackage{pifont}
\usepackage{xcolor}
\usepackage{amsmath}
\usepackage{pslatex}
\usepackage{algorithm}
\usepackage{algorithmic}
\usepackage{tablefootnote}
\usepackage{subcaption}
\usepackage[sorting=none, style=nature]{biblatex}
\bibliography{refs}

\begin{document}
\copyrightfalse

\title{Secure Control Systems\\
for Autonomous Quadrotors\\Against Cyber-Attacks}
\author{Samuel Belkadi}
\stuid{10834666}
\principaladviser{Dr. Wei Pan}

\beforeabstract

\prefacesection{Abstract}
%
{\singlespacing
The problem of safety for robotic systems has been extensively studied. However, little attention has been given to security issues for three-dimensional systems, such as quadrotors. Malicious adversaries can compromise robot sensors and communication networks, causing incidents, achieving illegal objectives, or even injuring people.

This study first designs an intelligent control system for autonomous quadrotors. Then, it investigates the problems of optimal false data injection attack scheduling and countermeasure design for unmanned aerial vehicles. Using a state-of-the-art deep learning-based approach, an optimal false data injection attack scheme is proposed to deteriorate a quadrotor's tracking performance with limited attack energy. Subsequently, an optimal tracking control strategy is learned to mitigate attacks and recover the quadrotor's tracking performance. 

We base our work on Agilicious, a state-of-the-art quadrotor recently deployed for autonomous settings. This paper is the first in the United Kingdom to deploy this quadrotor and implement reinforcement learning on its platform. Therefore, to promote easy reproducibility with minimal engineering overhead, we further provide (1) a comprehensive breakdown of this quadrotor, including software stacks and hardware alternatives; (2) a detailed reinforcement-learning framework to train autonomous controllers on Agilicious agents; and (3) a new open-source environment that builds upon PyFlyt for future reinforcement learning research on Agilicious platforms. 

Both simulated and real-world experiments are conducted to show the effectiveness of the proposed frameworks in section \ref{section:evaluation}.

}

\chapter{Introduction}
\label{cha:introduction}
Nowadays, robotic systems are becoming increasingly present in society and industry as an alternative to human labour for repetitive and dangerous tasks. Along with the recent developments of computers, communication networks, and intelligent sensing devices; modern robots can eventually be equipped with high-end technologies to produce improved throughput and better accuracy than human workers. However, this increase in autonomous robots exposes sensitive systems to adversaries. In fact, Upstream\footnote{A cybersecurity and data management platform.} has recorded a soar of 225\% in malicious attacks on autonomous systems in the past three years \cite{soar2255}. This danger arises from within-system communications between intelligent sensors and electrical control units over shared transmission networks. By executing cyber attacks,  malicious attackers can compromise the integrity of robotic systems and deteriorate their performance \cite{Clark2017CybersecurityII}. Such vulnerabilities open the door to adversarial entities that seek to exploit these systems, which can compromise people's safety.

Malicious attacks can be executed on various modules, such as the sensors and actuators, the communication network, and the physical interface of a system. For instance, Iranian air forces captured an American Lockheed Martin RQ-170 Sentinel unmanned aerial vehicle (UAV) in 2011 by spoofing its Global Position System (GPS) data \cite{csmonitor2011iran}. Two years later, some attackers tricked a yacht's navigation system by spoofing its GPS and putting it off course \cite{rutkin2013spoofers}. The authors of \cite{maggi2017rogue} described in fine detail how a hacker may attack an industrial robot and the consequences this may have on surrounding entities.

Although robotic systems' stability and safety issues have received much attention in recent years, with promising results being reported \cite{YANG201631, HUANG2021172}, previous works could not solve some security problems. To protect robots' performance from deterioration or even destruction, the study of designing robotic systems with high assurance is a hot topic, attracting researchers from the control and computer science communities. \\

This study intends to establish a secure robot learning framework for cyber attack scheduling and countermeasures. It analyses how an adversary with limited attack energy can construct an optimal false data injection attack scheme to disrupt a robot's tracking performance and proposes a secure control algorithm against such adversaries. Finally, we hope our paper will open the way to future research on this hot security challenge. The source code for our learning-based controller and open-source environment for reinforcement learning research on Agilicious can be found in \url{https://github.com/SamySam0/BaBee}. 

\section{Motivations}
In the control community, none of the existing methods adequately address optimal attack and countermeasure design problems on underactuated nonlinear complex systems. The method in \cite{han2020actor} focused on typical control problems without attack, while the method in \cite{hu2020lyapunov} focused on state estimation problems and cannot be applied to the secure control problem. We draw inspiration from \cite{wu2023secure}, which deals with ground vehicles of greater linearity. However, under attacks, the secure control problem for underactuated dynamical systems cannot be solved by applying the methods in \cite{wu2023secure}.

\section{Objectives and Contributions}
In this paper, we develop an autonomous control system, called a nominal controller, and two reinforcement learning-based methods to solve the optimal false data injection attack and countermeasure design problems for underactuated nonlinear systems. To this end, our study provides \textbf{(a)} a nominal controller for a robot under no attacks. Such a controller is capable of flying and stabilising the system. Then, \textbf{(b)} a learning-based malicious adversary employing false data injection attacks is constructed to perturb the control signals sent by the nominal controller and deteriorate its tracking performance with minimal cost. Finally, \textbf{(c)} a learning-based countermeasure is developed to mitigate attacks with minimum control cost and recover the tracking performance disturbed by the attacker. All the above algorithms are provided and described following the proximal policy optimisation approach. \\

We base our work on unmanned aerial vehicles, particularly on X-shaped quadrotors. As a study case, we deploy our framework on the Agilicious quadrotor, a state-of-the-art quadcopter for autonomous flying deployed in 2023 by the University of Zurich. In fact, \textbf{our team was the first in the United Kingdom to deploy this quadrotor and implement reinforcement learning on its platform}. Therefore, on top of the proposed adversary attack and defence countermeasure frameworks, our work introduces \textbf{(d)} a comprehensive breakdown of Agilicious quadrotors, including software designs and hardware alternatives, \textbf{(e)} a detailed reinforcement-learning framework to train autonomous controllers on Agilicious-based agent, and \textbf{(f)} a new open-source environment that builds upon PyFlyt for future reinforcement learning research on Agilicious platforms. These aim to promote easy reproducibility with minimal engineering overhead for future research on this promising quadrotor. As a little caviar, our team pre-designed various tasks, including hovering, obstacle avoidance, and trajectory tracking, to allow for straightforward experimental work on this framework.

The three learning-based controllers are evaluated against their respective objectives through a series of experiments. This includes experimenting in simulation to obtain metrical results, and testing their ability to be deployed in real-world scenarios. Results are compared to the state-of-the-art when they exist in order to provide the reader with a clear comparison to successful works.

\section{Report Structure}
The rest of this paper is organised as follows. Chapter \ref{cha:background} introduces background knowledge in quadrotor dynamics, cyber-attacks and countermeasures, and deep reinforcement learning. Chapter \ref{cha:background} Section \ref{section:prev-works} describes and refers the reader to related works. Chapter \ref{cha:setup} provides the experimental setup for our experiments. This includes our simulation and the assembly of the physical quadrotors used, i.e. Agilicious and Crazyflie, with detailed information on their hardware components and software stacks. Chapter \ref{cha:design} introduces our objectives and formulates the problems to be solved. Then Subsection \ref{section:nominal} presents the design of our autonomous control system, Subsection \ref{section:false-data} gives the learning-based false data injection attack algorithm, and Subsection \ref{section:countermeasure} the secure robot learning framework for mitigating attacks. Chapter \ref{cha:experiments} provides simulation and real-world experimental results with some in-depth discussion. This report is concluded in Chapter \ref{cha:conclusion} with a note on potential future works to extend our discoveries.
\chapter{Background}
\label{cha:background}

\section{Quadrotor Dynamics} \label{section:bg-dynamics}
In order to properly design aerial platforms according to our objectives, it is essential to understand the geometry and mechanics of a quadrotor. This section describes the dynamics of an X-shaped quadrotor and defines its kinematic model for later calculations. \\

\begin{figure}[h!]
  \centering
  \includegraphics[width=0.6\textwidth]{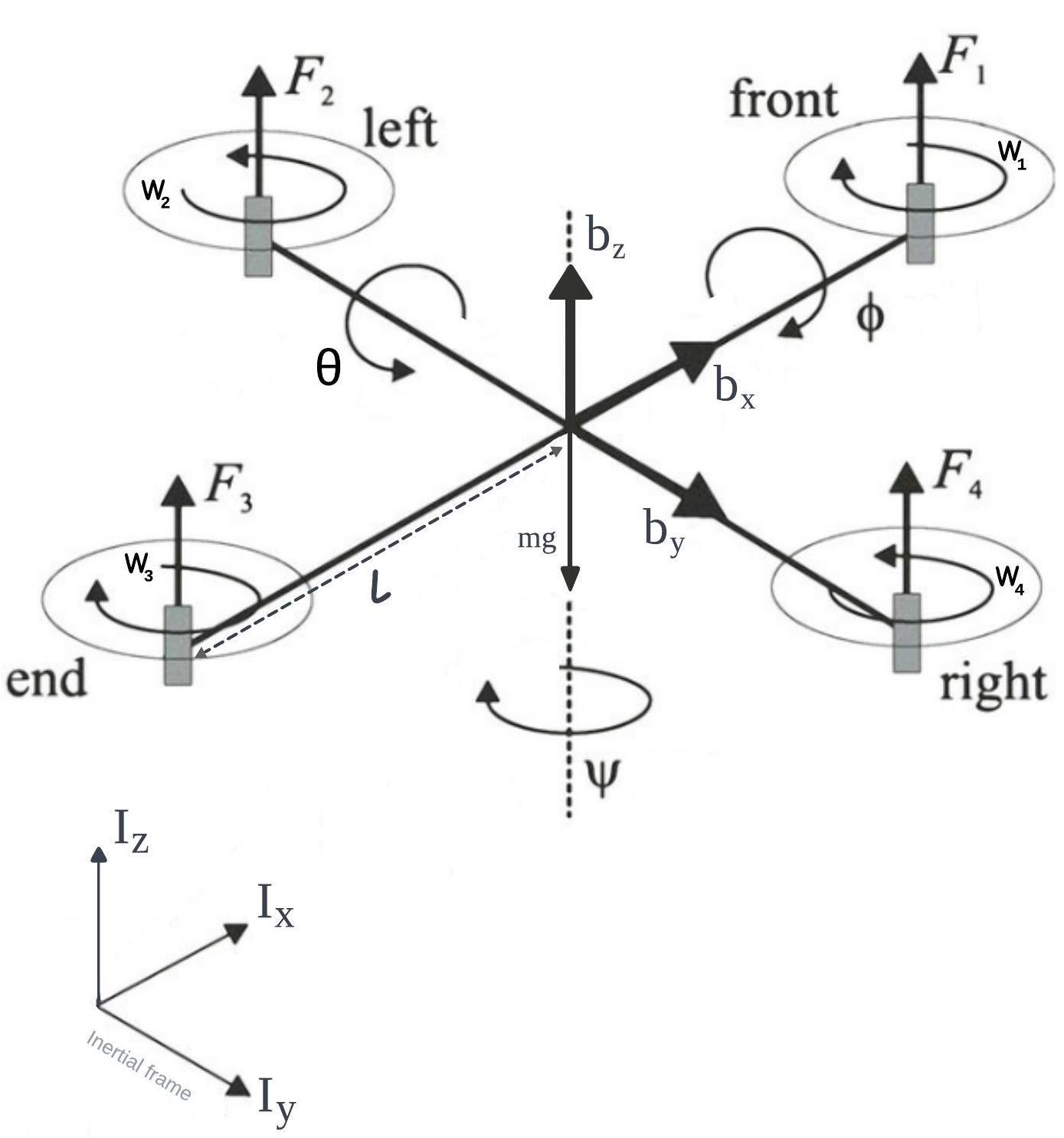}
  \caption[Quadrotor dynamics and reference frames.]{Quadrotor model with illustrated degrees of freedom. $F_i = k\cdot w^{2}_{i}$ is the force generated by motor $i$, and $mg$ is the force due to gravity. In a hovering state, $\sum_{i=1}^{4} F_i$ should balance $mg$.}
  \label{fig:quad_dynamics}
\end{figure}

A quadrotor is a \textit{rigid} aircraft with hardware components at its centre and four engines at the end of equally spaced arms. By definition, its motion has six degrees of freedom which are defined as follows: $\xi = [x,y,z]$ represents translation motions, i.e. the position of the quadrotor's centre, and $\Theta = [\theta, \phi, \psi]$ denotes rotation motions, i.e. the orientation of the quadrotor, also known as pitch, roll and yaw angles. An illustration of a quadrotor model is displayed in Figure \ref{fig:quad_dynamics}.

\subsection{Coordinate System and Reference Frames}
In our work, we use two reference frames to describe the motions of a quadrotor, where the first is fixed and the second is mobile. 
The mobile coordinate system called the \textit{intertial} frame $(I_x, I_y, I_z)$ describes an earth-fixed coordinate system with its origin located on the ground at a chosen location. In this system, we consider the first Newton’s law to be valid. 
On the other hand, the mobile coordinate system called the \textit{body} frame $(b_x, b_y, b_z)$ has its origin located at the centre of gravity of the quadrotor.
Note that we attribute linear positions and velocities to the inertial frame, and angular positions and velocities to the body frame.

\subsection{Thrust and Moments}
The quadrotor's attitude and position can be controlled by altering the speeds of its four motors to control thrust and rotations. The following forces and moments can be performed on the quadrotor: the thrust generated by spinning rotors, the pitching and rolling moments resulting from the speed difference between the four rotors, the gravity, gyroscopic effects, and yawing moments. 
However, the gyroscopic effect can be ignored as it is only noticeable in lightweight quadrotor constructions. Additionally, the yawing moments arise from imbalances in the rotational speeds of the four rotors, which can be countered by having two rotors spinning in opposite directions. 

Therefore, the propellers are grouped into pairs, with two diametrically opposed motors in each group, easily identifiable by their rotation direction. Namely, we group:
\begin{itemize}
    \item the front and rear propellers, rotating counterclockwise; and
    \item the left and right propellers, rotating clockwise.
\end{itemize}

By varying the rotational speeds of each motor, we can control the motion in the six degrees of freedom including forward and backward movements, lateral movement, vertical motion, roll motion, pitch motion, and yaw motion. However, we can directly influence only four of those six motions.

\begin{figure}[h!]
  \centering
  \includegraphics[width=\textwidth]{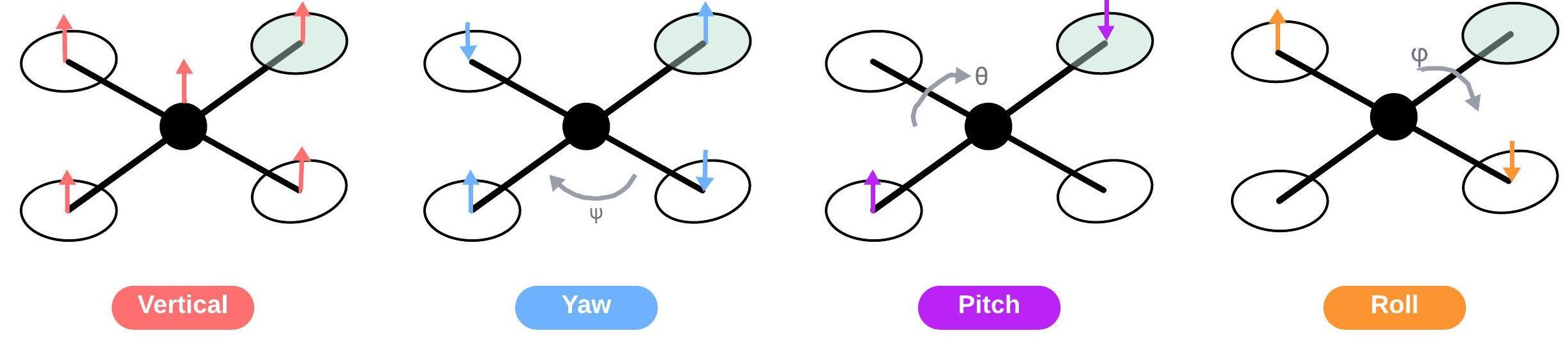}
  \caption[Quadrotor motions.]{Simplified illustrations of Altitude (vertical), Yaw, Pitch and Roll motions. The front propeller is highlighted in green for convenience.}
  \label{fig:rotation_motions}
\end{figure}

Depending on the rotational speed of each propeller, it is possible to identify the four basic movements of the quadrotor illustrated in Figure \ref{fig:rotation_motions}. Those are \textbf{(a)} the roll motion ($\phi$), being when left and right motors do not rotate with equal speed. Rolling increases when the left motor spins faster than the right one. 
\begin{equation} \begin{aligned}
    \phi &= l\cdot(F_2 - F_4) \\
         &= l\cdot k\cdot(w_{2}^{2} - w_{4}^{2})
\end{aligned} \end{equation}

\textbf{(b)} The pitch motion ($\theta$) when the front and rear motors do not rotate with equal speed. Pitching increases when the front motor spins faster than the rear one.
\begin{equation} \begin{aligned}
    \theta &= l\cdot(F_1 - F_3) \\
           &= l\cdot k\cdot(w_{1}^{2} - w_{3}^{2})
\end{aligned} \end{equation}

\textbf{(c)} The yaw motion ($\psi$) when the two motor groups do not sum to the exact same rotational speed. Yawing increases when the left and right motors spin faster than the front and rear ones.
\begin{equation}
    \psi = b\cdot(w_{4}^{2} + w_{2}^{2} - w_{1}^{2} - w_{3}^{2})
\end{equation}

Finally, it is assumed that \textbf{(d)} the torque and thrust caused by each rotor act particularly in the body's z-direction. Accordingly, the vertical motion ($F^z$) is the net propulsive force in the z-direction given by the sum of forces produced by each motor,
\begin{equation} \begin{aligned}
F^{z} &= F_1 + F_3 + F_2 + F_4 - mg \\
      &= k\cdot(w_{1}^{2} + w_{3}^{2} + w_{2}^{2} + w_{4}^{2}) - mg
\end{aligned} \end{equation}

where $k$ and $b$ are the lift and drag constants defined empirically, $mg$ is the force of gravity, and $l$ is the distance between any rotor and the centre of the quadrotor. As we assume our rigid body is symmetrical, all four arms must have the same length.

More generally, we can define the Thrust ($F$) and Moment ($M$) vectors as
\[
F = \begin{bmatrix}
    F^x \\
    F^y \\
    F^z
\end{bmatrix} = \begin{bmatrix}
    0 \\
    0 \\
    F_1 + F_2 + F_3 + F_4 - mg
\end{bmatrix}
\] 
and
\[
M = \begin{bmatrix}
    \phi \\
    \theta \\
    \psi
\end{bmatrix} = \begin{bmatrix}
    l\cdot k\cdot(w_{2}^{2} - w_{4}^{2}) \\
    l\cdot k\cdot(w_{1}^{2} - w_{3}^{2}) \\
    b\cdot(w_{4}^{2} + w_{2}^{2} - w_{1}^{2} - w_{3}^{2})
\end{bmatrix}.
\]

We can observe that the quadrotor has six degrees of freedom combining positions and orientations, but only four actuators. As a result, we must infer the change of the other two from the four we command. Therefore, such a quadrotor is considered an \textit{underactuated nonlinear complex system}.

\subsection{Kinematic Model of a Quadrotor}
\label{cha:quad-kinematics}
From the definitions above, we can infer the kinematic model of our quadrotor, i.e. the physical model on which we can act to control the vehicle. This section aims to show how the quadrotor's orientation and position change with respect to its angular velocities and (collective) thrust.

The translation of the quadrotor in the inertial frame is given by its position vector $\xi = [x,y,z]$. The change in position over time, i.e. the linear velocity in the inertial frame, can be expressed as the time-derivative of the linear positions,

\[
\frac{d\xi}{dt} = \nu.
\]

Here $\nu = [\Dot{x}, \Dot{y}, \Dot{z}]$ represents the inertial frame velocities. In order to relate these velocities to those in the body frame, we must apply a rotation transformation. \\

\textbf{Euler angles.} We will use ZYX Euler angles \cite{villanibruno} to describe the orientation of a frame of reference relative to another, and transform the coordinates of a point in a reference frame into the coordinates of the same point in another frame. Euler angles represent a sequence of three elemental rotations, i.e. rotations about the three axes of a coordinate system. The sequence is formed by the following rotation matrices \cite{lee2009robust}:

\[
\textbf{R}_x(\phi) = \begin{bmatrix}
    1 & 0 & 0 \\
    0 & \cos(\phi) & -\sin(\phi) \\
    0 & \sin(\phi) & \cos(\phi)
\end{bmatrix},
\]

\[
\textbf{R}_y(\theta) = \begin{bmatrix}
    \cos(\theta) & 0 & \sin(\theta) \\
    0 & 1 & 0\\
    -\sin(\theta) & 0 & \cos(\theta)
\end{bmatrix},
\]

\[
\textbf{R}_z(\psi) = \begin{bmatrix}
\cos(\psi) & -\sin(\psi) & 0 \\
\sin(\psi) & \cos(\psi) & 0 \\
0 & 0 & 1
\end{bmatrix}.
\]

Therefore, the inertial position coordinates and the body reference
coordinates are related by the rotation matrix $\textbf{R}_{zyx}(\phi, \theta, \psi)$, which describes the rotation from the body reference system to the inertial one as

\begin{equation}
\textbf{R}_{zyx}(\phi, \theta, \psi) = \textbf{R}_z(\psi) \cdot \textbf{R}_y(\theta) \cdot \textbf{R}_x(\phi) \\
\end{equation}
\[
= \begin{bmatrix}
c(\theta)c(\psi) & s(\phi)s(\theta)c(\psi) - c(\phi)s(\psi) & c(\phi)s(\theta)c(\psi) + s(\phi)s(\psi) \\
c(\theta)s(\psi) & s(\phi)s(\theta)s(\psi) + c(\phi)c(\psi) & c(\phi)s(\theta)s(\psi) - s(\phi)c(\psi) \\
-s(\theta) & s(\phi)c(\theta) & c(\phi)c(\theta)
\end{bmatrix},
\]

where 
$c(\psi)=\cos(\psi)$, $s(\psi)=\sin(\psi)$,
$c(\phi)=\cos(\phi)$, $s(\phi)=\sin(\phi)$,
$c(\theta)=\cos(\theta)$, and $s(\theta)=\sin(\theta)$. \\

\begin{figure}[h!]
  \centering
  \includegraphics[width=0.6\textwidth]{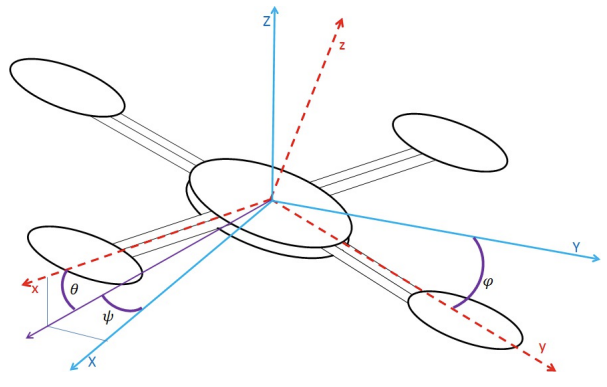}
  \caption[Euler angles.]{Illustration of Euler Angles.}
  \label{fig:euler_angles}
\end{figure}

Next, we analyse the rotation of the quadrotor in the body frame which is given by its orientation vector $\Theta = [\theta, \phi, \psi]$. The change in position over time, i.e. angular velocity in the body frame, can be expressed as the time-derivative of the orientation

\[
\frac{d\Theta}{dt} = \omega_b.
\]

Here $\omega_b = [\Dot{\theta}, \Dot{\phi}, \Dot{\psi}]$ represents angular velocities in the body frame. In order to relate these velocities to those in the inertial frame, we must apply the following angular transformation

\begin{equation}
\textbf{T} = \begin{bmatrix}
1 & \sin(\phi)\tan(\theta) & \cos(\phi)\tan(\theta) \\
0 & \cos(\phi) & -\sin(\phi) \\
0 & \frac{\sin(\phi)}{\cos(\theta)} & \frac{\cos(\phi)}{\cos(\theta)}
\end{bmatrix}.
\end{equation}

Therefore, from 3D body dynamics, it follows that the two reference frames (inertial and body) are linked by the following relations:

\begin{equation} \begin{aligned}
    \nu &= \textbf{R} \cdot \nu_b, \\
    \omega &= \textbf{T} \cdot \omega_b,
\end{aligned} \end{equation}

where $\nu$ and $\nu_b$ are linear velocities in the inertial and body frames, respectively, and $\omega$ and $\omega_b$ are angular velocities in the inertial and body frames, respectively. \\

This shows that thrust and orientations in the body frame, which are controlled through angular velocities, directly influence the quadrotor's position and orientation in the inertial frame.

\subsection{Differential Flatness}
Although the previous section shows that the quadrotor's orientation and position change with respect to its thrust and angular velocities, it does not prove that any trajectory in the space of flat outputs will be dynamically feasible for the quadrotor. Specifically, it does not show that controlling thrust and angular velocities guarantees that it can take any desired trajectory in 3D space. To prove that, we must show that the quadrotor system is differentially flat; that is, the states and inputs can be written as algebraic functions of the flat outputs and their time-derivatives. This proof is way beyond the scope of our work but can be read in \cite{faessler2017differential}. \\

The proof shows that the extended dynamical model of a quadrotor subject to rotor drag with four controlled inputs is differentially flat. In fact, the quadrotor states $[\xi, \nu_b, \Theta, \omega]$ and inputs $[F^{z}, \omega_w]$ can be written as algebraic functions of four selected flat outputs and a finite number of their derivatives. They chose the flat outputs to be the quadrotor’s position $\xi$ and its heading $\psi$, and proved that the commanded orientations and collective thrust are functions of the flat outputs. \\

This concludes on the fact that commanded orientations and collective thrust are not only directly influencing the quadrotor's position and orientation with respect to the inertial frame, but can also represent any realistic trajectory in 3D space.

\section{Quadrotor Attacks and Countermeasures}
\label{section:bg-cyber-attack}
In this section, we define cyber-attacking in the context of quadrotor tracking disruption and, more specifically, introduce false data injection as a man-in-the-middle approach. Additionally, we describe countermeasures as a way to mitigate malicious attacks and preserve the quadrotor's tracking performance.\\

In the context of quadrotor tracking and control, we refer to cyber-attacking as a malicious attempt to compromise the availability and integrity of transmitted information. Nonetheless, it is essential to mention that system faults and attacks differ fundamentally. That is, while faults manifest from a system's unintended error, attacks are concealed and designed elaborately \cite{bezzo2014attack, jin2018average, sun2020fault}. Such disruptions can be particularly concerning for quadrotors as they mainly rely on real-time data for stable flight and navigation. 

One specific form of cyber-attack is the false data injection, where incorrect or misleading data are inserted into the quadrotor's data streams. In our study, the method used to execute false data injections is the man-in-the-middle approach, where an attacker intercepts and alters the communications between the quadrotor and its control system. This involves altering actuator signals and sending disrupted control commands to the flight control system. Such attacks may result in minor disruptions or even complete loss of control when no countermeasure is executed, posing significant risks to safety and task integrity.\\

Two types of strategies are typically employed to effectively address and counter false data injections. The first are \textit{passive defences}, which involve designing more robust control systems. For example, such methods could implement redundant systems or fail-safes that can detect discrepancies in data and initiate appropriate procedures. On the other hand, \textit{active defences} involve systems that actively monitor for signs of cyber-attacks and respond in real-time to mitigate them. These can include anomaly detection systems that use statistical models to identify unusual data, or intrusion detection systems that monitor network traffic for signs of malicious activity.

\section{Reinforcement Learning}
This section introduces reinforcement learning as a mathematical solution to autonomous quadrotor control, optimal false data injections and countermeasure design. Moreover, it explains how deep reinforcement learning can handle the high-dimensional state spaces and complex environments required to deal with three-dimensional quadrotor controls. Finally, it goes in-depth with the Proximal Policy Optimization algorithm implemented to train our three controllers. \\

\begin{figure}[h!]
  \centering
  \includegraphics[width=0.8\textwidth]{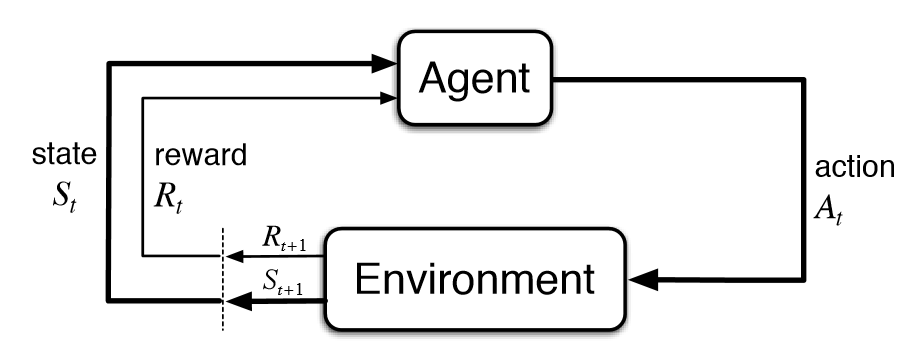}
  \caption[Interaction between an agent and its environment.]{Simplified interaction between an agent and its environment.}
  \label{fig:rl-env}
\end{figure}

Reinforcement learning is a subset of Machine Learning where an intelligent system, referred to as an agent, learns through trial and error by interacting in an unknown environment and receiving feedback on its actions \cite{sutton2018reinforcement}. The mechanism is the following: In any given state of an environment, an agent uses its policy to choose which action to take based on the observed state, and receives a reward from the environment for doing so. The reward tells the agent ``how good'' his action was. This process continues with the agent repeatedly taking actions and receiving rewards until the environment terminates or the agent reaches a final state. Reinforcement learning uses the rewards obtained through training episodes to update the agent’s policy and progress toward the optimal policy in an environment. Essentially, reinforcement learning allows an agent with no prior information about the environment to learn a strategy about how to interact with the environment and maximize accumulated rewards. An illustration of the interaction between an agent and its environment is displayed in \ref{fig:rl-env}.

There exist two main challenges with reinforcement learning. These are \textbf{(1)} to write a qualitative reward function to receive accurate feedback from the agent's actions, and \textbf{(2)} to design an environment that accurately describes the environment in which the agent will act when deployed. Indeed, this latter point is at the heart of the Simulation-to-Real-World transfer challenge.

\subsection{Markov Decision Processes} \label{section:mdp}
Formally, reinforcement learning can be described as Markov decision processes (MDP). We represent these with the tuple $(\mathcal{S}, \mathcal{A}, \mathcal{P}, \mathcal{R}, \gamma)$, where $\mathcal{S}$ is the state space, $\mathcal{A}$ is the action space, $\mathcal{P}$ is the transition probability matrix from a state-action pair at time $t$ onto a distribution of possible states at time $t+1$, $\mathcal{R}(s_t, a_t, s_{t+1})$ is the immediate reward function, and $\gamma \in [0, 1)$ is the discount factor for future rewards, where lower values place more emphasis on immediate rewards.

\subsubsection{Policies}
In general, the policy $\pi$ is a mapping from states to a probability distribution over actions: $\pi : S \to p(\mathcal{A} = a|\mathcal{S})$. 
In our case, the Markov decision processes are episodic, meaning that the state is reset after each episode of length $T$. Therefore, the sequence of states, actions and rewards in an episode constitutes a policy \textit{rollout}. Every rollout of a policy accumulates rewards from the environment, resulting in the return 

\[
R = r_0 + \gamma r_1 + \gamma^2r_2 ... = \sum_{t=0}^{T-1}\gamma^{.t}r_{t+1}.
\]

The goal of reinforcement learning is to find an optimal policy $\pi^*$ which maximizes the expected return from all states:

\begin{equation}
    \pi^* = \underset{\pi}{\text{argmax}} \, \mathbb{E}[R|\pi].
\end{equation}

Therefore, the agent must deal with long-range time dependencies, as the consequences of an action often only materialise after many transitions in the environment. This is known as the \textit{temporal credit assignment problem} \cite{sutton1984temporal}.

\subsection{Traditional Reinforcement Learning}

With Markov decision processes as the key formalism of reinforcement learning, there exist two main approaches to solving reinforcement problems: methods based on value functions and methods based on policy search. \\

Value function methods are based on estimating the value (expected outcome) of being in a given state. The state-value function $V^\pi(s)$ is the expected outcome of being in state $s$ and following $\pi$ henceforth:
\begin{equation}
V^\pi(s) = \mathbb{E}[R|s, \pi].
\end{equation}

The optimal policy $\pi^*$ has a corresponding state-value function $V^*(s)$, and vice-versa, the optimal state-value function can be defined as
\begin{equation}
V^*(s) = \max_\pi V^\pi(s) \quad \forall s \in \mathcal{S}.
\end{equation}

If we could obtain $V^*(s)$, the optimal policy could be retrieved by choosing among all actions available from $s_t$ and picking the action $a$ that maximises the expected future reward from state $s_t$ according to transition probabilities $\mathcal{P}$, as $\mathbb{E}_{s_{t+1} \sim \mathcal{P}(s_{t+1}|s_t,a)}[V^*(s_{t+1})]$.

However, this assumes that transition dynamics $\mathcal{P}$ are available, which is not the case in the context of reinforcement learning. Therefore, we construct another function, the state-action-value function $Q^\pi(s, a)$, which is similar to $V^\pi$ except that the initial action $a$ is provided, and $\pi$ is only followed from the succeeding state onwards:
\begin{equation}
Q^\pi(s, a) = \mathbb{E}[R|s, a, \pi].
\end{equation}

The best policy can be found given $Q^\pi(s, a)$ by choosing action $a$ greedily at every state, following $arg\max_a \, Q^\pi(s, a)$. Under this policy, we can also define $V^\pi(s)$ by maximising $Q^\pi(s, a)$ as $V^\pi(s) = \max_{a} Q^\pi(s, a)$.

\subsubsection{Dynamic Programming}
To actually learn $Q^\pi$, we exploit the Markov property and define the function as a Bellman equation \cite{bellman1966dynamic}, which has the following recursive form:
\begin{equation}
Q^\pi(s_t, a_t) = \mathbb{E}_{s_{t+1}}[r_{t+1} + \gamma Q^\pi(s_{t+1}, \pi(s_{t+1}))].
\end{equation}

This means that $Q^\pi$ can be improved by \textit{bootstrapping}, i.e., we can use the current values of our estimate of $Q^\pi$ to improve our estimate. This is the foundation of Q-learning \cite{watkins1992q} where updates occur as follows:
\begin{equation}
Q^\pi(s_t, a_t) \leftarrow Q^\pi(s_t, a_t) + \alpha [r_t + \gamma\max_{a'} Q(s_{t+1},a') - Q(s_t, a_t)],
\end{equation}
where $\alpha$ is a constant or decreasing learning rate, and $\alpha \in (0,1]$.

\subsection{Deep Reinforcement Learning}
\begin{figure}[h!]
  \centering
  \includegraphics[width=0.8\textwidth]{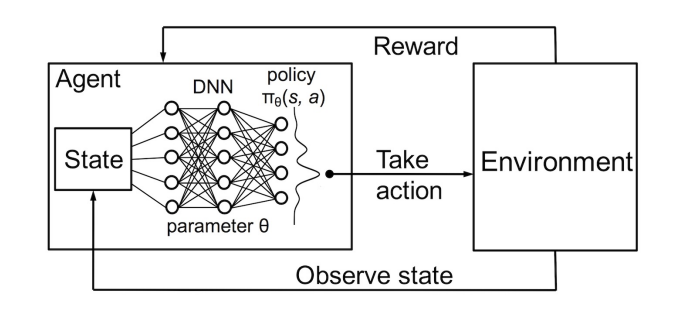}
  \caption[End-to-end deep reinforcement learning framework.]{Simplified illustration of an end-to-end deep reinforcement learning framework. Neural network architectures may vary according to the specific task.}
  \label{fig:drl-env}
\end{figure}

Deep reinforcement learning combines reinforcement learning and deep neural networks by using deep neural networks as function approximators to help agents learn how to achieve their goals. Where tabular learning previously struggled to represent such problems, this approach supports high-dimensional and continuous state spaces, as well as complex environments.
Figure \ref{fig:drl-env} shows how deep learning and reinforcement learning can be used together.

\begin{figure}[th!]
  \centering
  \includegraphics[width=0.5\textwidth]{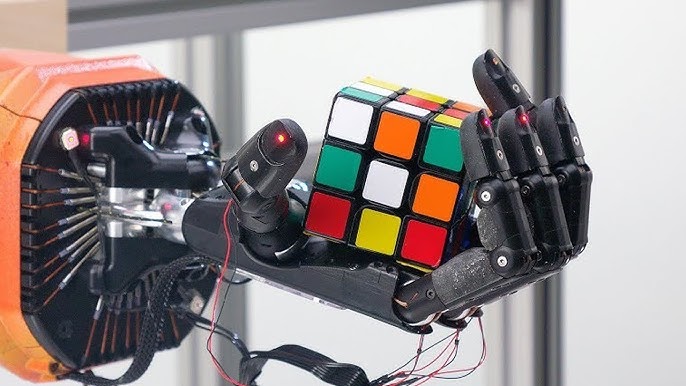}
  \caption[Dactyl 2019: OpenAI's robotic hand.]{Dactyl 2019: OpenAI's robotic hand solving a Rubik's Cube.}
  \label{fig:openai-dactyl}
\end{figure}

In recent works, deep neural networks have been used to approximate policy functions, resulting in superhuman performances at tasks such as board games \cite{silver2017mastering}, computer vision \cite{he2016deep} and robotics \cite{finn2016guided}. For instance, Figure \ref{fig:openai-dactyl} showcases Dactyl, OpenAI's robotic hand capable of dexterous manipulations.

The learning process for deep learning algorithms is the following: During training, the agent iteratively observes a state and selects an action based on the neural network's predictions. The neural network aims at approximating the policy and predicting the future outcomes of a given action. According to the chosen algorithm, the action may be selected deterministically or stochastically. Note that a balance must be maintained between exploration and exploitation throughout training.

Deep reinforcement learning has been used extensively in quadrotor controls \cite{app13031275, 5d43d183cab84e4894ea7cef68892ef6, han2020actor} and in a few works on optimal false data injections and countermeasures for unmanned ground vehicles \cite{wu2023secure}. 

\subsection{Proximal Policy Optimization Algorithm (PPO)}
\label{section:ppo}
\begin{figure}[th!]
  \centering
  \includegraphics[width=0.83\textwidth]{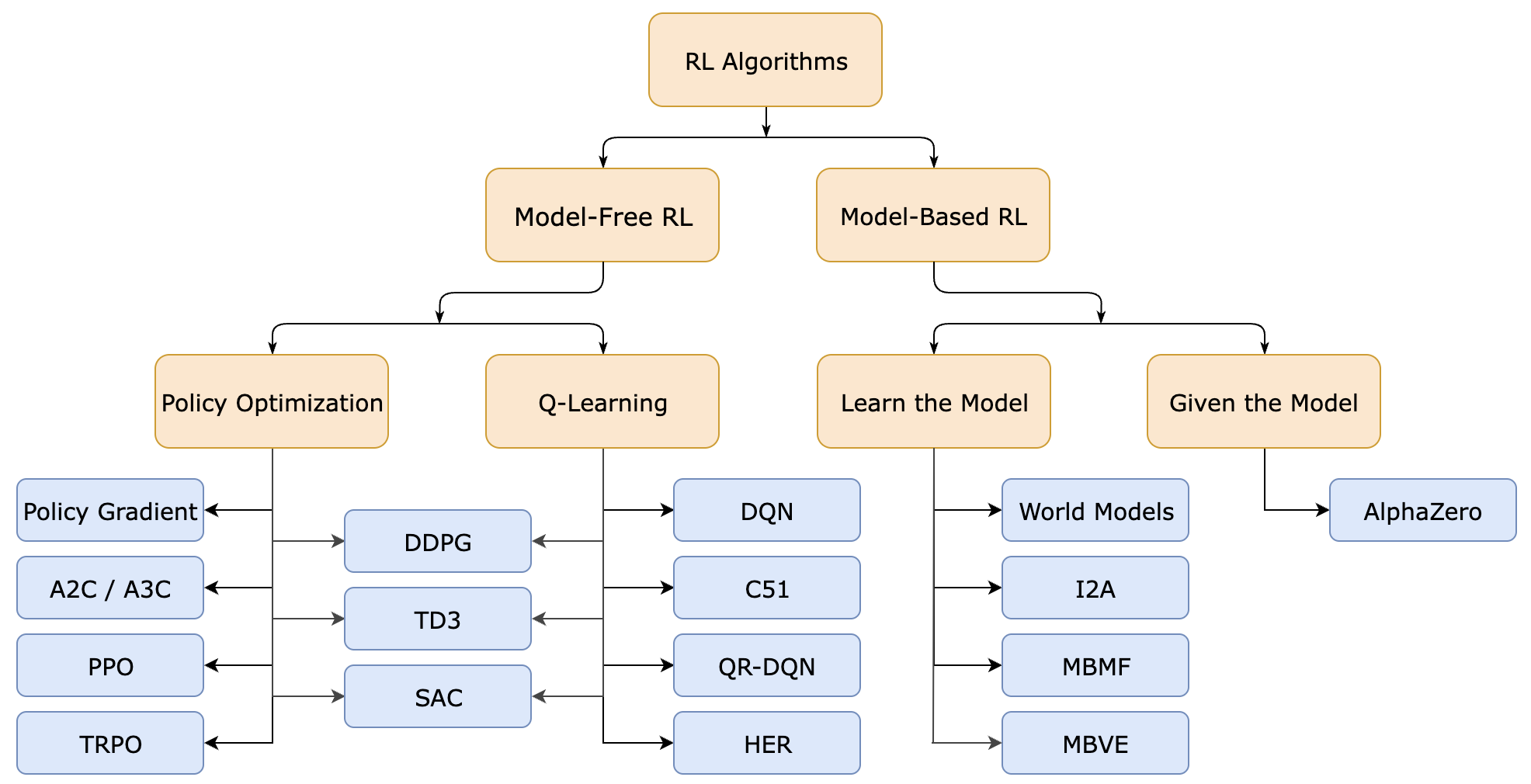}
  \caption[Most widely used reinforcement learning algorithms.]{Most widely used reinforcement learning algorithms to date.}
  \label{fig:RL-algos}
\end{figure}

Although numerous reinforcement learning algorithms exist (Figure \ref{fig:RL-algos}), the selection of a method depends entirely on the specific working task. In this section, we describe the Proximal Policy Optimization algorithm (PPO) \cite{schulman2017proximal} as a deep learning algorithm widely used in quadrotor control \cite{lopes2018intelligent, koch2019reinforcement} due to its fast and easy implementation and improved training stability. 

As a model-free method, PPO can efficiently deal with quadrotors' high complexity and nonlinear dynamics by learning a control policy directly from interactions with the environment and without requiring an explicit model of the vehicle dynamics. Additionally, it can operate in continuous high-dimensional state and action spaces without an exponential increase in complexity. For instance, PPO has become OpenAI's default reinforcement learning algorithm due to its ease of use and good performance.\\ 

PPO is composed of an actor $\pi_\theta(\cdot | s_t)$ which outputs a probability distribution for the next action given the state at timestamp $t$, and a critic $V(s_t)$ which estimates the expected cumulative reward from that state. Since both the actor and critic take the states as input, we can save computation by sharing the neural network architecture between the policy and the value functions.
Therefore, \cite{schulman2017proximal} proves that we must use a loss function that combines the policy surrogate and a value function error term. Furthermore, as suggested in \cite{williams1992simple}, this objective can be further augmented by adding an entropy bonus to ensure sufficient exploration. Combining these terms, we obtain the following objective maximized at each iteration
\begin{equation}
    L_t^{\text{CLIP+VF+S}}(\theta) = \hat{\mathbb{E}}_t \left[ L_t^{\text{CLIP}}(\theta) - c_1 L_t^{\text{VF}}(\theta) + c_2 S[\pi_{\theta_{old}}(s_t)] \right],
\end{equation}

where $c_1$ and $c_2$ are coefficients which balance the importance of the accuracy of the critic against the exploration capabilities of the policy, $S$ denotes an entropy bonus, and $L_t^{\text{VF}}$ is a squared-error loss $(V_\theta(s_t) - V_t^{\text{targ}})^2$. As suggested by Stable Baseline\footnote{\nolinkurl{https://github.com/DLR-RM/stable-baselines3/blob/master/stable_baselines3}}'s implementation, we will use $c_1 = 0.5$ and $c_2 = 0.05$.

\subsubsection{Clip term}
The \textit{clip term} $L_t^{\text{CLIP}}(\theta)$ maximizes the probability of actions, given by the actor $\pi_\theta(\cdot | s_t)$, that resulted in an advantage. Additionally, it tries to improve the policy's training stability by avoiding large policy updates between training epochs. The motivation behind the latter is based on empirical analysis which proved that smaller policy updates during training were more likely to converge to an optimal solution \cite{lillicrap2015continuous}. The clip term is defined as
\begin{equation}
    L^{\text{CLIP}}(\theta) = \hat{\mathbb{E}}_t \left[ \min(r_t(\theta) \hat{A}_t, \text{clip}(r_t(\theta), 1 - \epsilon, 1 + \epsilon) \hat{A}_t) \right],
\end{equation}

with
\begin{equation} \begin{aligned}
    r_t(\theta) &= \frac{\pi_\theta(a_t, s_t)}{\pi_{\theta_\text{old}}(a_t, s_t)}, \\
    \hat{A}_t &= -V(s_t) + r_t + \gamma r_{t+1} + \dots + \gamma^{T-t+1} r_{T-1} + \gamma^{T-t} V(s_T),
\end{aligned}\end{equation}

where the \textit{ratio function} $r_t(\theta)$ tells how likely we are to take action $a_t$ in the current policy as opposed to the previous policy. In order to stabilise the training, we do not allow for drastic changes in policy and, therefore, take the minimum of the new policy and its clipped version between $[1-\epsilon, 1+\epsilon]$. In their paper, \cite{schulman2017proximal} recommend using $\epsilon=0.2$ so that the ratio can only vary from $0.8$ to $1.2$.

The advantage $\hat{A}_t$ measures how wrong the critic was for the given state $s_t$. It does this by running the policy for $T$ timesteps and measuring the difference between the value at the initial state $V(s_t)$ and the true cumulative reward obtained from following policy $\pi_\theta$ over the next $T$ steps. 

\subsubsection{Value-function term}
To have a good estimate of the advantage, we need a critic which can accurately predict the value of a given state. The \textit{value-function term} accounts for learning such a function with a simple mean square error loss between its predicted expected reward and the observed cumulative reward $(V_\theta(s_t) - V_t^{\text{targ}})^2$ computed as
\begin{equation}
    L_t^{VF} = \text{MSE}\left(r_t + \gamma r_{t+1} + \ldots + \gamma^{T-t+1}r_{T-1} + V(s_T), V(s_t)\right) = \| \hat{A}_t \|_2^2.
\end{equation}

\subsubsection{Entropy term}
Finally, PPO encourages exploration with a small bonus on the entropy of the output distribution of the policy. For the \textit{entropy term}, we consider the standard entropy
\begin{equation}
    S[\pi_{\theta}](s_t) = - \int \pi_{\theta}(a_t|s_t) \log(\pi_{\theta}(a_t|s_t)) da_t.
\end{equation}

\subsubsection{Algorithm}
An implementation of the proximal policy optimization algorithm that uses fixed-length trajectory segments is given in Algorithm \ref{algo:ppo}. In every iteration, each of $N$ parallel actors collects $T$ timesteps of data. Then, we construct the surrogate loss on these $NT$ timesteps of data and optimize it with Adam \cite{kingma2014adam} gradient descent for $K$ epochs.

\begin{algorithm}
\caption{Proximal Policy Optimization Algorithm}
\label{algo:ppo}
\begin{algorithmic}[1]
\FOR{iteration = 1, 2, \dots}
    \FOR{actor = 1, 2, \dots, N}
        \STATE Run policy $\pi_{\text{old}}$ in environment for $T$ timesteps
        \STATE Compute advantage estimates $\hat{A}_1, \dots, \hat{A}_T$
    \ENDFOR
    \STATE Optimize surrogate $L$ wrt $\theta$, with $K$ epochs and minibatch size $M \leq NT$
    \STATE $\theta_{\text{old}} \leftarrow \theta$
\ENDFOR
\end{algorithmic}
\end{algorithm}

\section{Related Works} \label{section:prev-works}
Autonomous quadrotor control systems have been investigated considerably over the past few years. Initiating from mathematical models such as PD \cite{erginer2007modeling} or MPC \cite{torrente2021data}, recent studies have tried to train learning-based controllers using reinforcement learning \cite{app13031275, 5d43d183cab84e4894ea7cef68892ef6, han2020actor}. As a result, although mathematical models provide the safest and most predictable behaviour for autonomous flying, learning-based controllers have demonstrated strong abilities to solve non-linear control problems that control theory could not solve. \\

On that account, although learning-based controllers may have limited flying abilities, combining them with mathematical models could indeed solve significant complex problems that occur during quadrotor flying operations. That is notably the case with malicious attacks, which typically can be categorised into four distinct types: the denial of service (DoS) attack, the false data injection attack, the replay attack, and the zero-dynamics attack. For the reader's convenience, all are described in \cite{teixeira2012attack}. Although securing robotic systems in the presence of malicious attacks is a new challenge, a few results have already been reported. For example, \cite{kerns2014unmanned} showed how to capture and control a UAV by spoofing GPS data. Regarding stealthy attacks on ground vehicle sensors, \cite{bezzo2014attack} proposed a robust control system that can estimate system states while under attack by utilising redundant sensor measurements. \\

Following is a description of the state of research in this area. In \cite{guo2018roboads}, \textit{Guo, P. et al.} designed an attack detection scheme based on the non-linear dynamics of a mobile robot to warn the system in case of actuator attacks. In their work, the proposed scheme was successfully verified on two types of robots with various attack scenarios. Furthermore, \cite{lee2021distributed} proposed a defence scheme that switches distributed control between multiple robots to avoid actuator DoS attacks and falsified data injections. However, these two solutions do not directly preserve tracking performance but instead suggest ways to detect or improve the robustness of robotic systems.
Over the past two decades, alternative secure algorithms have emerged from schemes previously developed for cyber-physical systems, based on control-theoretical approaches. Some examples include a switching observer-based estimate scheme \cite{lu2019secure}, a linear quadratic secure controller \cite{an2018lq}, and a learning-based secure tracking control algorithm \cite{wu2021learning}.

In addition, false data injection attacks have been widely investigated due to their stealthy characteristic. As a result, many systems have been proposed, including attack detection schemes \cite{luo2020interval}, secure state estimate algorithms \cite{hu2018state, kazemi2020secure}, and resilient controllers \cite{lucia2016set}. Unfortunately, these programs either cannot be applied to quadrotor systems or do not commit to preserving a confident level of stability in agile settings. From an attacker's perspective, researchers have investigated how to construct effective attack sequences to deteriorate system performance, such as DoS attack scheduling \cite{qin2020optimal} and false data injection attack scheduling \cite{gao2021class}, based on which can be designed more effective countermeasures. \\

In general, although progress has been made in securing cyber-physical systems, the results mentioned above rely on exact system knowledge, which may not be obtained easily. Using reinforcement learning may allow us to design systems policies without using such knowledge. In fact, a few reinforcement learning algorithms have been proposed \cite{mnih2015human, lillicrap2015continuous, schulman2015trust} to this end, based on which learning-based autonomous control with stability guaranteed \cite{han2020actor} have been developed. More recently, considering the complexity and scale of cyber-physical systems, some attempts have been made at utilising deep reinforcement learning to solve security problems. Although not directly designed for unmanned aerial vehicles, encouraging results have been demonstrated on problems such as anomaly detection, secure control, attack detection schemes and other security-related applications \cite{nguyen2021deep}.

Similarly, adversaries may also use deep reinforcement learning approaches to construct attack schemes \cite{lee2021query}. However, how to design an optimal attack scheme and optimal countermeasure for \textbf{quadrotors} under false data injection attacks is still a hot challenge awaiting some solutions. This study provides a solution employing deep reinforcement learning and opens the way for some future research.
\newcommand{\greencheck}{{\textcolor{green}\checkmark}}
\newcommand{\xmark}{\ding{55}}
\newcommand{\redcross}{{\textcolor{red}\xmark}}

\chapter{Experimental Setup}
\label{cha:setup}

\section{Quadrotor Hardware Assembling}
\label{section:qaudrotor-models}
This section introduces the physical quadrotors used in our experiments and the reasons behind our choices. Furthermore, it reports previous results obtained on these models in the context of tracking performance and reinforcement-learning-based control.

\subsection{Agilicious} \label{section:agilicious}
\begin{figure}[h!]
  \centering
  \includegraphics[width=0.95\textwidth]{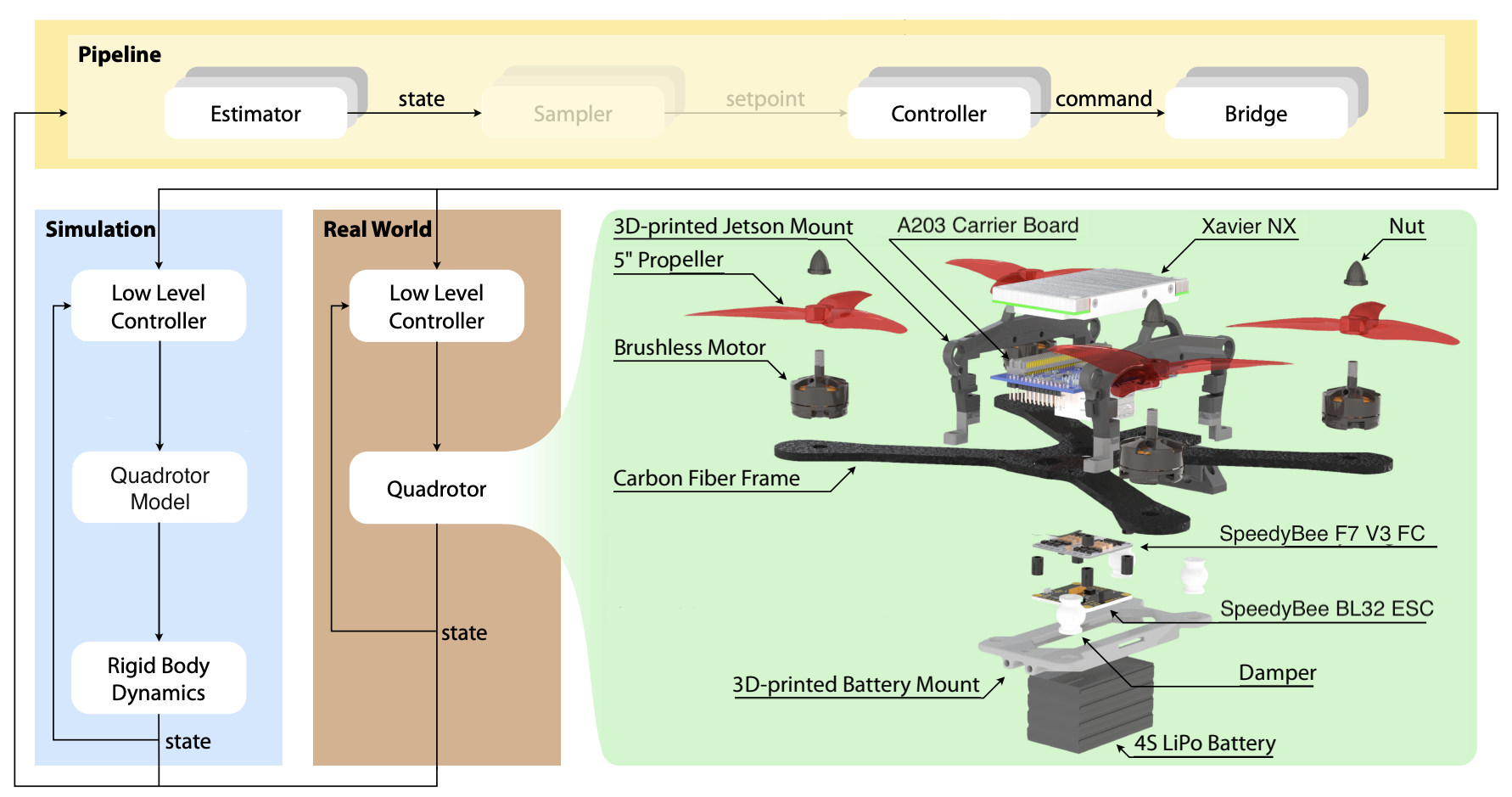}
  \caption[Customised framework of Agilicious.]{Customised framework of Agilicious used in our work. We provide two pipelines: one \textit{with} a sampler and an MPC-based controller, and one \textit{without} a sampler which utilises a neural network as its controller. Our hardware is designed for positional sensing rather than visual sensing as originally suggested in \cite{drone:Agilicious}. Finally, our simulation is based on PyFlyt with the PyBullet engine and is described in section \ref{section:bg-env}.}
  \label{fig:agilicious_pipeline}
\end{figure}

Agilicious is a co-designed hardware and software framework tailored to autonomous and agile quadrotor flight. It was deployed by the University of Zurich in 2023 as an open-source and open-hardware quadrotor supporting both model-based and neural-network–based controllers. It provides high thrust-to-weight and torque-to-inertia ratios for improved agility, and GPU-accelerated compute hardware for efficient neural network inference. Our customised Agilicious framework is described in Figure \ref{fig:agilicious_pipeline}.

\subsubsection{Motivation}
\begin{figure}[h!]

\includegraphics[width=\textwidth]{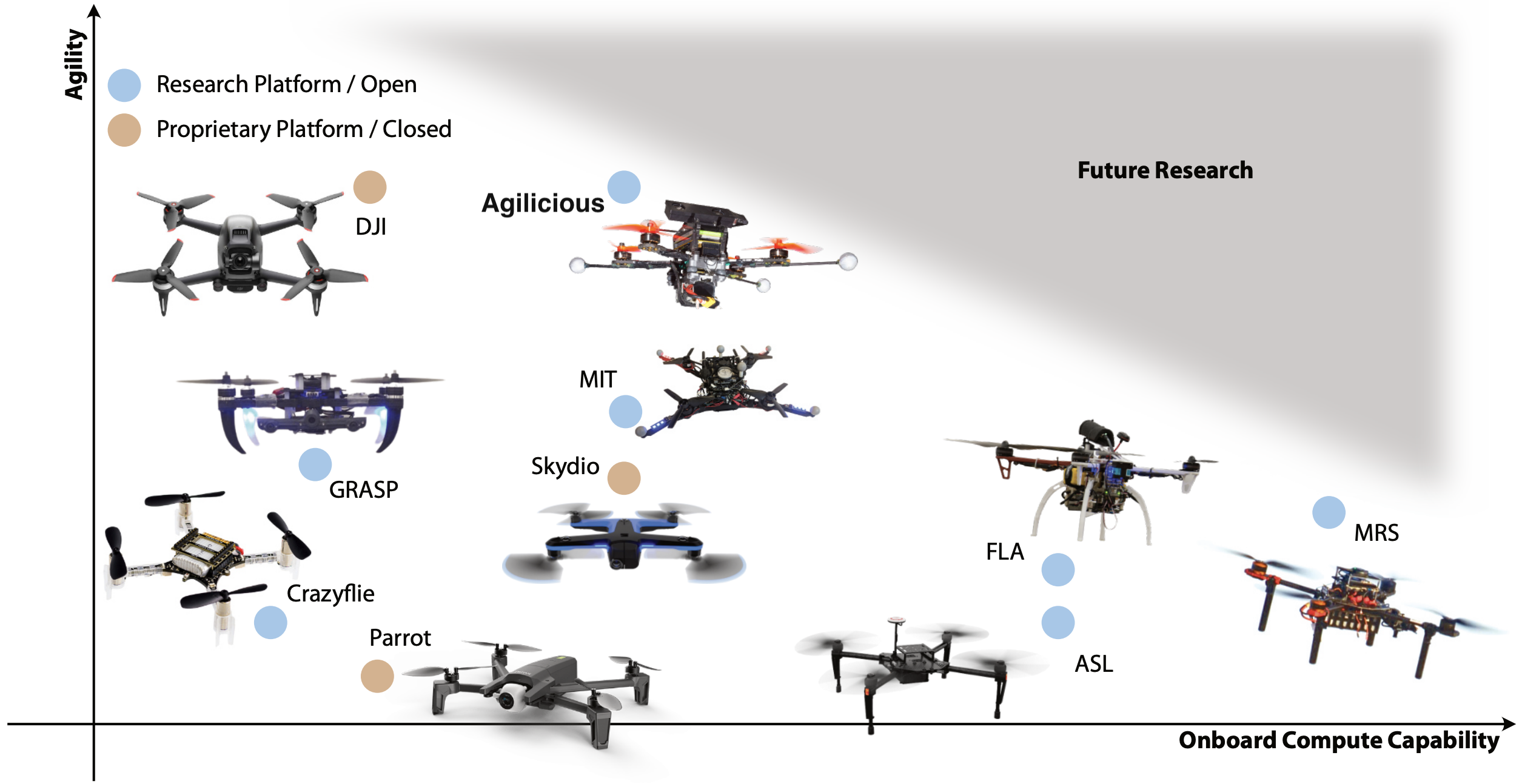}
\vspace{10pt}

\centerline{
    \begin{tabular}{l|l|p{2cm}|p{2.5cm}|l|p{1.7cm}|p{1cm}}
    \hline
    \textbf{Framework} & Open-source & Onboard Computer & CPU mark (higher better) & GPU & Maximum speed & Thrust to weight \\ \hline
    PX4 \cite{drone:PX4} & \greencheck & \redcross & - & \redcross & - & - \\
    Paparazzi \cite{drone:Paparazzi} & \greencheck & \redcross  & - & \redcross & - & - \\
    DJI \cite{drone:DJI} & \redcross & \redcross & - & \redcross & 140km/h & 4.43 \\
    Skydio \cite{drone:Skydio} & \redcross & \redcross & - & \redcross & 58km/h & - \\
    Parrot \cite{drone:Parrot} & \redcross & \redcross & - & \redcross & 55km/h & - \\
    Crazyflie \cite{drone:Crazyflie} & \greencheck & \redcross & - & \redcross & - & 2.26 \\
    FLA-Quad \cite{drone:FLA-Quad} & \greencheck & \greencheck & 3.383 & \redcross & - & 2.38 \\
    GRASP-Quad \cite{drone:GRASP-Quad} & \redcross & \greencheck & 625 & \redcross & - & 1.80 \\
    MIT-Quad \cite{drone:MIT-Quad} & \redcross & \greencheck & 1.343 & \greencheck & - & 2.33 \\
    ASL-Flight \cite{drone:ASL-Flight} & \greencheck & \greencheck & 3.383 & \redcross & - & 2.32 \\
    RPG-Quad \cite{drone:RPG-Quad} & \greencheck & \greencheck & 633 & \redcross & - & 4.00 \\
    MRS UAV \cite{drone:MRS-UAV} & \greencheck & \greencheck  & 8.8846 & \redcross & - & 2.50 \\ \hline
    \textbf{Agilicious} \cite{drone:Agilicious} & \greencheck & \greencheck & 1.343 & \greencheck & 131km/h & 5.00 \\ \hline 
    \end{tabular}}
    \caption[Comparison of different quadrotor models.]{A comparison from Agilicious work on different available consumer and research platforms with respect to available onboard compute capability and agility. The sequence of experience to obtain the above results is described in \cite{drone:Agilicious}. The PX4 \cite{drone:PX4} and the Paparazzi \cite{drone:Paparazzi} are rather low-level autopilot frameworks without high-level computation capability. The open-source frameworks FLA \cite{drone:FLA-Quad}, ASL \cite{drone:ASL-Flight}, and MRS \cite{drone:MRS-UAV} have relatively large weight and low agility. The DJI \cite{drone:DJI}, Skydio \cite{drone:Skydio}, and Parrot \cite{drone:Parrot} are closed-source commercial products that are not intended for research purposes. The Crazyflie \cite{drone:Crazyflie} does not allow for sufficient onboard compute or sensing, while the MIT \cite{drone:MIT-Quad} and GRASP \cite{drone:GRASP-Quad} platforms are also not open-source. Instead, the Agilicious \cite{drone:Agilicious} framework provides agile flight performance, onboard GPU-accelerated compute capability, as well as open-source and open-hardware availability.}
    \label{fig:agilicious_results}
    
\end{figure}

In order to build an agile quadrotor and experiment with autonomous tracking tasks, the quadrotor should meet the following pair of design requirements. It must \textbf{(1)} carry the computing hardware needed for autonomous operation and \textbf{(2)} be capable of agile flight. To meet the first requirement on computing resources needed for true autonomy, a quadrotor should carry sufficient computing capacity to run estimation, planning, and control algorithms concurrently; hence requiring heavy hardware. 
However, the physical model must deliver adequate thrust-to-weight and torque-to-inertia ratios to permit agile flight. The thrust-to-weight ratio can often be enhanced using more powerful motors, which in turn require larger propellers and, thus, a larger platform size.
On the other hand, the torque-to-inertia ratio decreases with higher weight and size. That is due to the moment of inertia increasing quadratically with the size and linearly with the weight of the platform. 
As a result, it is desirable to achieve the best trade-off between agility (i.e., maximising both thrust-to-weight and torque-to-inertia ratios) and available compute resources onboard. In their work, Agilicious proved to be the most appropriate quadrotor solution for autonomous and agile flights. For convenience, we display the results of their experiments in Figure \ref{fig:agilicious_results}.

In addition to its hardware capabilities, Agilicious proposes a software stack to quickly transfer algorithms from simulation to real-world deployment. This modularity allows us to test and experiment with our research code without the need to develop an entire flight stack, facilitating development and reproducibility.
Moreover, it supports the popular Robot Operating System (ROS) used throughout our development. 

Ultimately, Agilicious remains quite unexplored due to its recency. In fact, our team is the first in the United Kingdom to build upon this quadrotor and implement reinforcement learning on its platform. Consequently, we aim to support the community by providing detailed insights into its capabilities for reinforcement learning while promoting reproducibility and minimal engineering overhead.

\clearpage

\subsubsection{Components}
\begin{figure}[hb!]
  \centering
  \includegraphics[width=1\textwidth]{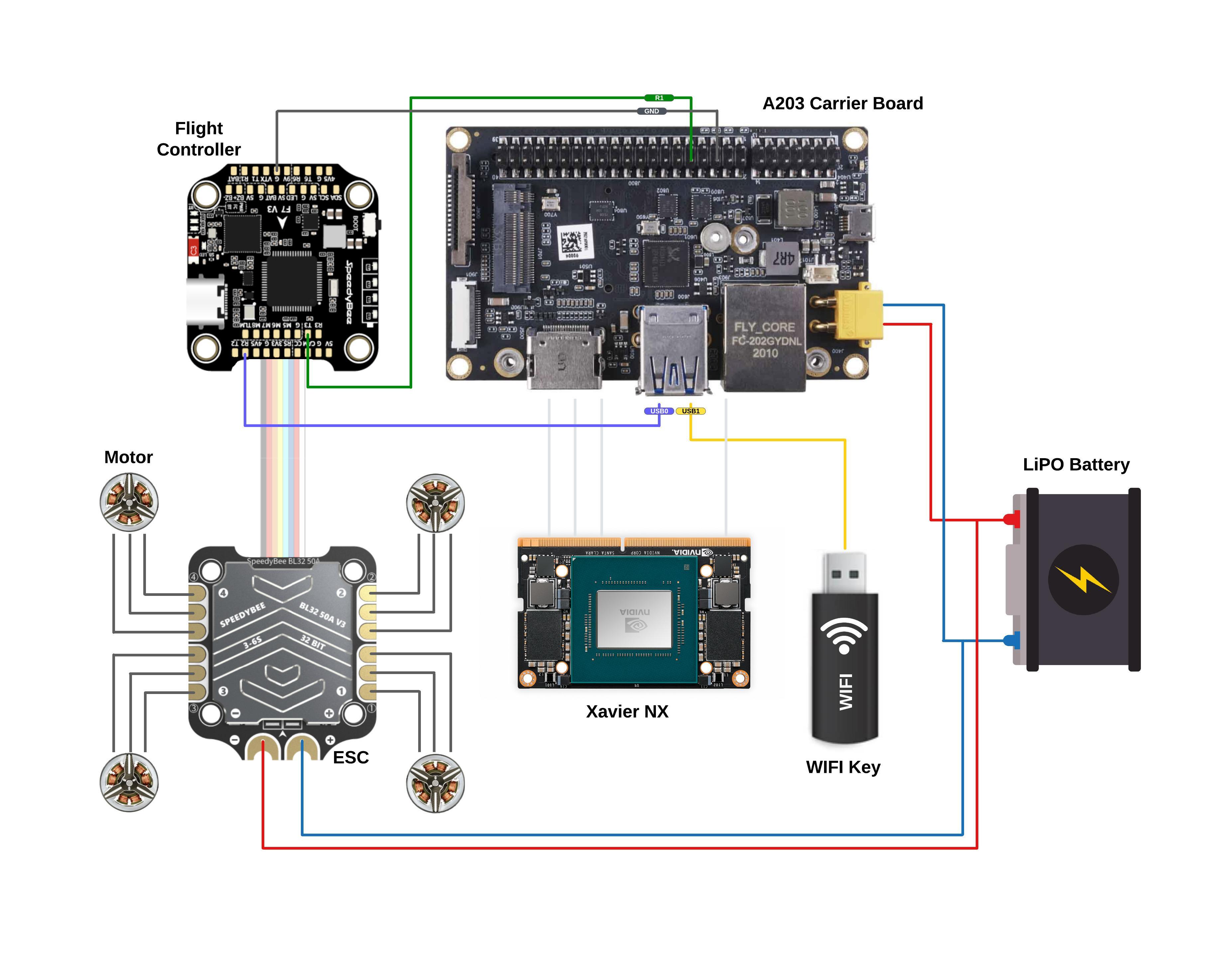}
  \caption[Inter-communications between Agilicious' hardware components.]{Top view of the hardware system describing the communications between components. \textbf{(a)} The Nvidia Xavier NX (Compute Unit) is embedded at the back of the A203 Carrier Board and connected through a 260-pin Connector. The Compute Unit runs all calculations on board from the A203 Carrier Board, supplied by a LiPO battery. To access a router or network, the A203 Carrier Board is also equipped with a Wifi Key on its USB1 port. \textbf{(b)} The Carrier Board is connected to the Flight Controller to communicate by sending control commands through its USB0-RX (transmitting) port and receiving feedback on its UART1-RXD (receiving) port. In order to use USB0 as a transmitting port, we utilise a USB-to-TTL cable (note that the USB-to-TTL cable inverts RX and TX; therefore, we use RX instead of TX on USB0 for transmission). \textbf{(c)} The Flight Controller is connected to the Carrier Board to communicate by sending feedback on its T3 (transmitting) port and receiving control commands on its R2 (receiving) port. It is also connected to the ESC through an 8-pin connector (on Bi-directional protocol) to send the processed Control Commands to the ESC and receive the root feedback from the latter. \textbf{(d)} The ESC receives processed control commands and varies the speed and direction of the motor based on these commands. ESC's power is also provided on board by the LiPO battery.}
  \label{fig:hardware_system}
\end{figure}

The hardware system of the quadrotor aims to carefully manage the trade-off between onboard computing capability and platform agility. We describe below the components used to build the flight and compute hardware, as well as why we decided to use them. A detailed description of the hardware system and inter-components communications is displayed in Figure \ref{fig:hardware_system}. \\

\textbf{High-level Controller:} The high-level controller is responsible for providing the necessary computing power to run the full flight stack, including estimation, planning, optimization-based control and neural network inference. Therefore, it must provide \textbf{strong computing power} while preserving \textbf{limited weight onboard}. 
It combines two components: \textbf{(1)} a Carrier Board which serves as an interface between the computing unit and the various peripherals and connectors (such as USB, Ethernet, sensors...) through serial ports; and \textbf{(2)} a Compute Unit which provides the necessary computing power onboard and is directly embedded onto the Carrier Board. For the latter, although we recommend using the Nvidia Jetson TX2 as it possesses the best performance-to-size ratio, it is hardly available to order. Therefore, we decided to use the Nvidia Xavier NX as it performs almost as good as the former and supports accelerated inference from the Nvidia CUDA general-purpose GPU architecture. As for the carrier board, we used the A203 Carrier Board, which is compatible with the Nvidia Xavier NX and offers the minimal configuration required to communicate with the low-level controller, i.e., two UART ports.\\

\textbf{Low-level Controller:} The low-level controller provides real-time low-latency interfacing and control. It is made up of two components: \textbf{(1)} The first is a Flight Controller (FC), which is a microcontroller equipped with various sensors (gyroscope, accelerometer and magnetometer) responsible for processing inputs from the high-level controller and sending control commands to the ESC, as well as getting feedback from the ESC and sharing them with the high-level controller over the digital bi-directional DShot protocol. \textbf{(2)} The second is the Electronic Speed Controller (ESC). It is the interface between the Flight Controller and the quadrotor’s motors. Each motor on the quadrotor is connected to the ESC, which controls the speed and direction of the motor based on commands received from the Flight Controller by adjusting the power supplied to the motors. The ESC sends feedback to the Flight Controller, such as rotor speed, IMU, battery voltage and flight mode information over a 1 MBaud serial bus at 500 Hz. For our Flight Controller and ESC, we used the SpeedyBee F7 V3 FC with BetaFlight Firmware and the SpeedyBee BL32 50A 4-in-1 ESC, respectively. \\

\textbf{Flight Hardware:} To maximize the agility of the quadrotor, it must be designed as lightweight and small as possible \cite{doi:10.1177/0278364912455954} while still being able to accommodate the Xavier NX compute unit. Therefore, we selected the following off-the-shelf drone components, summarised in Table \ref{tab:hardware_components}.
The Armattan Chameleon 6-inch frame is used as a base plate since it is one of the smallest frames with enough space for computing hardware. Its carbon fibre makes it durable and lightweight (86g). For propulsion, four 5.1-inch three-bladed propellers are used with fast-spinning DC motors rated at a high maximum power of 750W. The chosen motor-propeller combination achieves a continuous static thrust of $4\times9.5$N on the quadrotor and consumes about 400W of power per motor. To match the high power demand of the motors, a lithium-polymer battery with 2000 mAh and a rating of 120C is used. Therefore, the total peak current of 110A is well within the limit of the battery. \\

\begin{table}[h!]
\centerline{
    \begin{tabular}{l|l|p{4.6cm}}
    \hline
    \textbf{Component}     & \textbf{Product} & \textbf{Specification} \\ \hline
    Carrier Board          & A203 Carrier Board            & 87mm x 52mm x 26mm, 100 g \\
    Compute Unit           & Nvidia Jetson Xavier NX       & 70mm x 45mm, 80 g  \\ \hline
    Flight Controller (FC) & SpeedyBee F7 V3 FC            & 41 x 38 x 8.1mm, 9 g, BetaFlight firmware \\
    Motor Controller (ESC) & SpeedyBee BL32 50A 4-in-1 & 45.6 x 40 x 8.8mm, 12 g, 4x 50A \\ \hline
    Frame                  & Armattan Chameleon 6 inch     & 4 mm carbon fiber, 86 g \\
    Motor                  & TMotor F40 Pro V              & 23x6 mm stator, 2150 kV, 750 W, 4x 34 g  \\
    Propeller              & T5147 POPO Racing Tri-blade          & 5.1 inch length and 4.8 inch pitch, 4x 4.4 g \\
    Battery                & Tattoo R-Line v3.0 2000       & 4× 3.7 V, 2000 mAh, 217 g \\                                                 
    \hline \end{tabular}}
    \caption{Overview of the components of the flight hardware design.}
    \label{tab:hardware_components}
\end{table}

\begin{figure}[h!]
  \centering
  \includegraphics[width=0.6\textwidth]{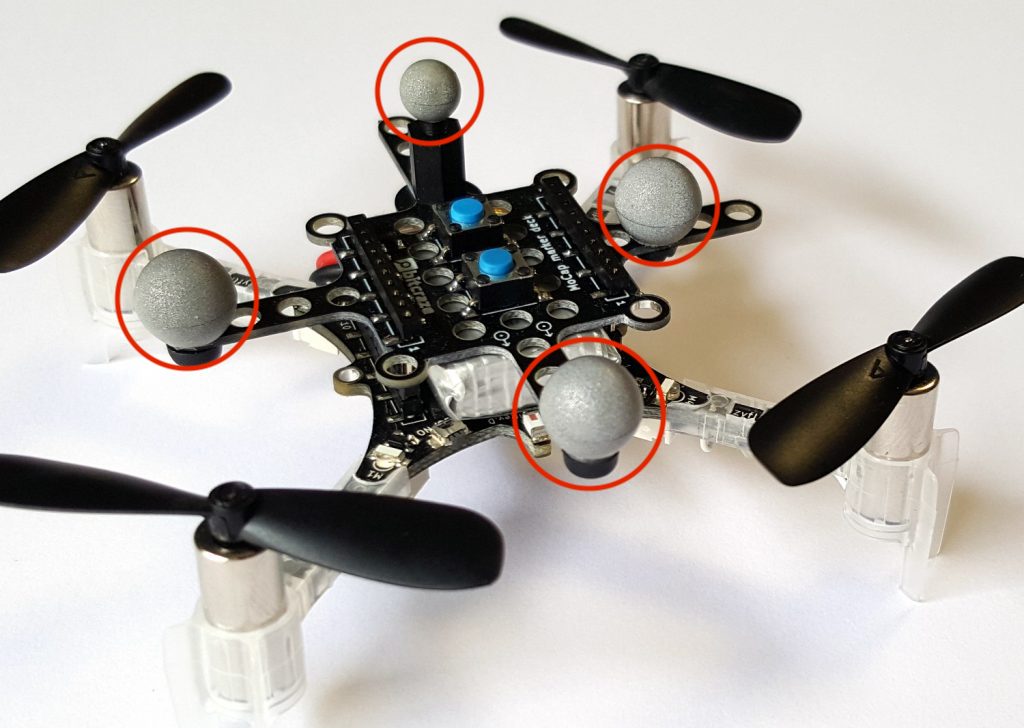}
  \caption[Example of a suitable reflective markers configuration.]{Example of a suitable reflective markers configuration on a Crazyflie quadrotor. The four reflective markers are circled in red.}
  \label{fig:markers}
\end{figure}

\textbf{Sensors:} To arbitrarily navigate into an environment, quadrotors need some way to measure their absolute or relative locations and orientations. For that, many odometry solutions exist to estimate state changes over time. The most famous \textit{positional sensing} techniques are GPS for outdoor localisation and Motion Capture systems for indoor and controlled environments. Alternatives exist with \textit{visual sensing} methods, such as camera-based visual-inertial odometry, but we decided not to use that one in order to match our current research flow. In our work, we utilise Motion Capture with eight cameras as a means to obtain odometry estimates of the quadrotor’s position (in m) and linear velocity (in m/s) expressed in the inertial frame, as well as orientation (in °) and body rates (in °/s) in the body frame. First, we placed some reflective markers on the quadrotor’s body. An example of a suitable configuration is displayed in Figure \ref{fig:markers}. It is essential to follow the requirements below to obtain optimal results:
\begin{itemize}
    \item \textit{Visibility:} Markers should be distributed such that most of them are visible by the motion capture cameras no matter what the quadrotor's orientation is.
    \item \textit{Avoid Occlusion:} Markers should be placed such that they are less likely to be occluded by the drone's body or arms during flight. Solutions imply placing some markers on stalks or elevated positions.
    \item \textit{Orientation Tracking:} To track orientation, markers should not be placed in a uniform pattern. We must use an asymmetrical arrangement to help the motion capture system distinguish the drone's orientation efficiently.
\end{itemize} 

Furthermore, the motion capture system must be calibrated so that each camera understands the geometry, position and orientation it captures in space and infers a reference state with respect to the world. After calibration, the motion capture system can accurately estimate the odometry of a quadrotor in the inertial frame. 

\begin{figure}[h!]
  \centering
  \includegraphics[width=\textwidth]{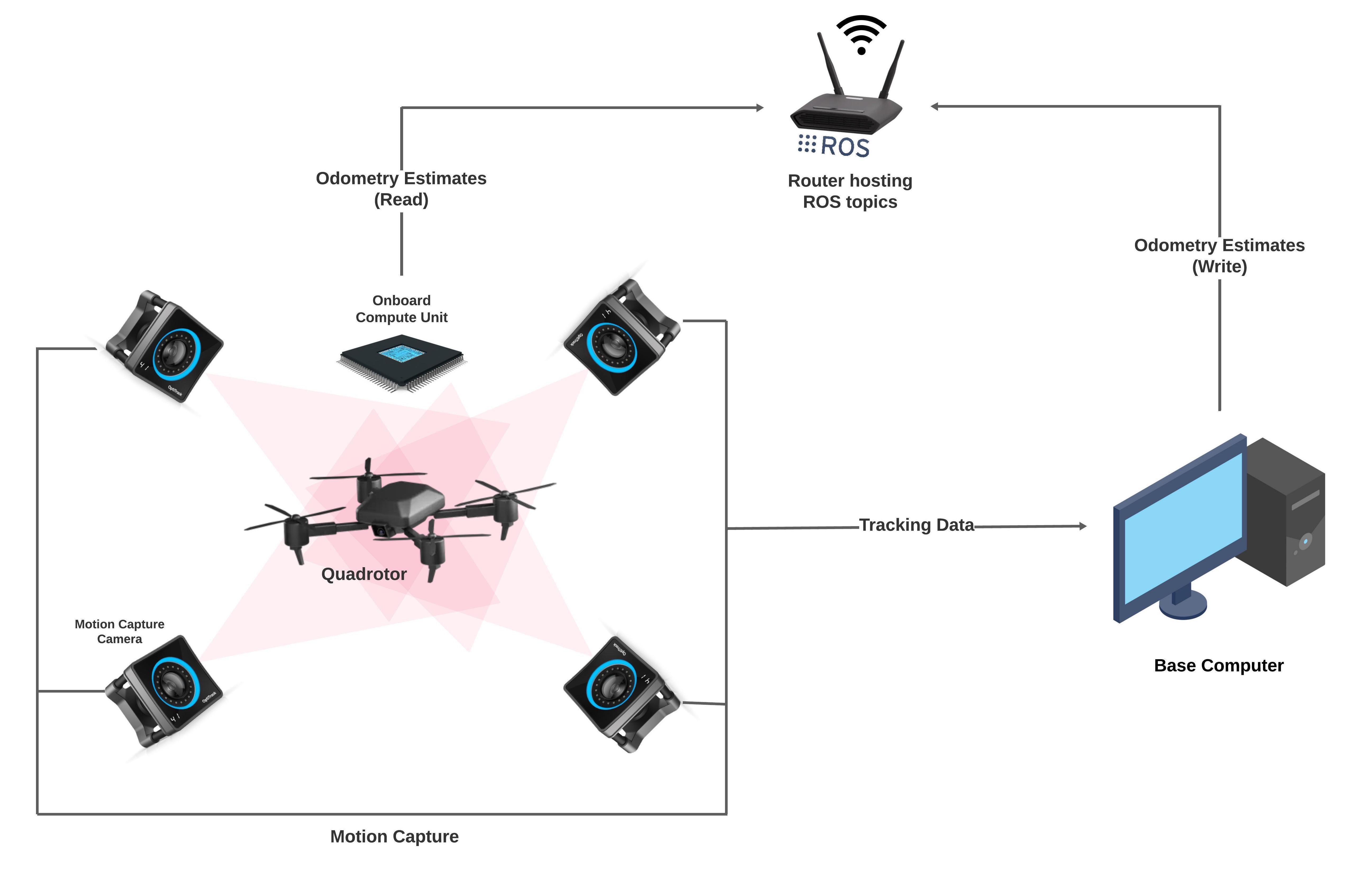}
  \caption[Sketch of the Motion Capture system used for localisation.]{Entire motion capture system illustrating how the Onboard Compute Unit can obtain the quadrotor's state estimates, shared by the Base Computer and derived through motion capture. The motion capture cameras send individual tracking data to the Base Computer. The latter calculates and publishes odometry estimates to the appropriate ROS topic hosted on the router. The quadrotor's Onboard Computer Unit subscribes to the ROS topic on the router's IP to read odometry estimates in real time and produce control commands.}
  \label{fig:mocap_system}
\end{figure}

Finally, the motion capture system sends these estimates to a Base Computer. In turn, the Base Computer publishes them in real-time on a ROS topic accessible by the onboard Compute Unit, which will utilise them to calculate control commands. Figure \ref{fig:mocap_system} provides an illustration depicting the entire system in action.

\subsubsection{Pilot}
Agilicious uses modular software components unified in a \textit{pipeline} and orchestrated by a control logic: the \textit{pilot}. The modules consist of an estimator, a sampler, a controller, and a bridge. All are working together to complete a given task. These modules are executed sequentially (illustrated in Figure \ref{fig:agilicious_pipeline}) within a forward pass of the pipeline, corresponding to one control cycle. The pilot contains the main logic needed for flight operation and handling of the individual modules. At its core, it loads and configures the software modules and runs the control loop. \\

\textbf{Pipeline:} A pipeline is a distinct configuration of a defined module sequence, for example, an estimator followed by a controller and a bridge. These pipeline configurations can be switched at any time. In our work, we experiment on two different pipelines: \textbf{(1)} The first uses an intermediate sampler, which, given estimator states, generates a subset of points along the desired trajectory to pass to a state-of-the-art MPC controller. \textbf{(2)} The second pipeline directly exploits our trained neural network as the controller without the need for a sampler. \\

\textbf{Estimator:} The first module in the pipeline is an estimator, which provides a time-stamped state estimate for the subsequent software modules in the control cycle. A state estimate is a set $x = [p,q,\nu,\omega,a,\tau,j,s,b_\omega,b_a,f_d,f]$ with quadrotor's position $p$, orientation unit quaternion $q$, velocity $\nu$, body rate $\omega$, linear $a$ and angular $\tau$ accelerations, jerk $j$, snap $s$, gyroscope and accelerometer bias $b_\omega$ and $b_a$, and desired and actual single rotor thrusts $f_d$ and $f$.
We use an Extended Kalman Filter (EKF) as our feed-through estimator, which fuses the data from the Inertial Measurement Unit (IMU) with the state estimates recovered from the motion capture system to provide a more accurate and reliable estimate of the quadrotor's state. In parallel, because sensors operate at different frequencies, the estimator synchronizes these signals to provide a consistent final state estimate. \\

\textbf{Sampler (Optional):} When utilising a mathematical control system such as a PD or MPC, the controller module needs to be provided with a subset of trajectory points that encode the desired progress along it. The sampler is responsible for providing that sequence. In our experiment on an MPC-based controller, we use a position-based sampling scheme that selects trajectory progress based on the current position of the quadrotor and its target location. Note that such a sampler is unnecessary when using our neural-network-based controller since it takes raw state estimates as inputs to directly predict control commands as expected body rates and collective thrusts. \\

\textbf{Controller:} The next module in the chain is the controller. It outputs control commands given state estimates (or given a set of trajectory points if a sampler is used). In our experiment, as an initial approach, we utilise a state-of-the-art MPC that uses the full non-linear model of the platform to generate body rate controls. In the context of reinforcement learning, we use our neural network as a controller, which, given state estimates, outputs desired body rates and collective thrusts $[\dot{\phi}, \dot{\theta}, \dot{\psi}, F]$. Therefore, this latter approach does not need a sampler to generate intermediate trajectory points. \\

\clearpage
\textbf{Bridge:} The final module of our pipeline is a bridge. It serves as an interface between the hardware and software layers to send the control commands to the low-level controller. We decided to use a bridge provided by Agilicious, which communicates via an SBUS protocol. As a standard in the quadrotor community, this protocol further promotes reproducibility. Additionally, it allows for interfacing with BetaFlight’s flight controllers, as required for our system.

\subsection{Crazyflie} \label{section:crazyflies}
Being eight times smaller and twenty times lighter than Agilicious, Crazyflies are a series of small quadrotors developed by Bitcraze AB, promoting lightweight design and versatility. Additionally, thanks to their compact and accurate architecture, Crazyflies represent good baselines for quick real-world deployment post-training. However, their design comes to the cost of lower onboard computing capacity and poor agility.

\subsubsection{Motivation}
In some of our experiments, we chose Crazyflie as an alternative to Agilicious. This is done to deploy our trained controllers to real-world scenarios without the security and safety concerns that Agilicious raises.

\subsubsection{Components}
\begin{figure}[h!]
  \centering
  \includegraphics[width=0.8\textwidth]{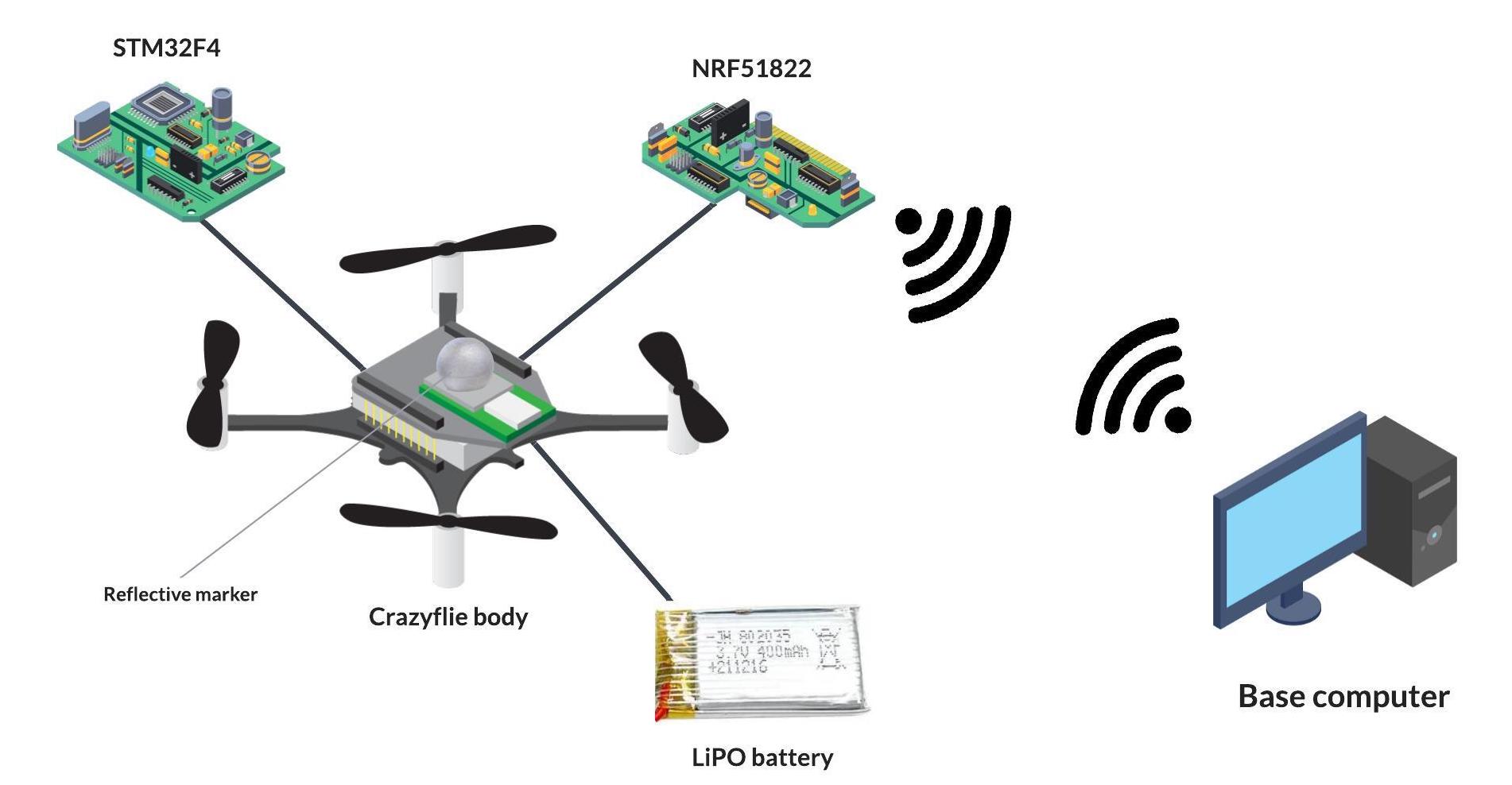}
  \caption[Illustration of Crazyflie's hardware system.]{Illustration of the hardware system of a Crazyflie quadrotor and its communication with the base computer.}
  \label{fig:crazyflie-hardware}
\end{figure}

As indicated above, the hardware of a Crazyflie is much lighter than that of Agilicious. For convenience, an illustration of its hardware is depicted in Figure \ref{fig:crazyflie-hardware}. Crazyflie has two microprocessors: \textbf{(a)} the \textbf{STM32F4} handles the firmware stack, including low-level and high-level controllers; and \textbf{(b)} the \textbf{NRF51822}, which handles all the radio communications. Due to limited computation capacities onboard, calculations are done off-board from the base computer, with control commands transmitted through radio signals to the Crazyflie. Finally, the system is alimented by a LiPO battery directly connected to the body’s bolt. Reflective markers are also attached to the quadrotor’s body to obtain state estimates in the same way as with Agilicious (Figure \ref{fig:mocap_system}).

\subsubsection{Pilot}
Regarding the pilot, Crazyflie uses the same pipeline as Agilicious, except for a few details omitted here for simplicity, as they have no impact on the understanding or reproducibility of our work. 

The reinforcement learning controllers used to fly the Crazyflie quadrotor have been trained using the same parameters as those used for Agilicious. The only changes made to the environment are linked to the physical properties of the quadrotors, such as weight, size, rotor power, etc. To this end, all physical properties have been directly taken from the constructor’s documentation. Note that, as opposed to Crazyflie, the physical properties of Agilicious quadrotors differ from one assembly to another and must be measured by ourselves. This may often result in light measurement errors and, thus, reduced performance.

\section{Simulation and Environment} \label{section:bg-env}
In this section, we introduce the methodology used to set up our environment and describe its potential to support future reinforcement learning research on Agilicious quadrotors. \\

Throughout our investigations, we developed a custom environment based on the open-source PyFlyt simulator \cite{tai2023pyflyt}. Although most related works were conducted on the famous pybullet-drone \cite{panerati2021learning} environment, our strategy involved modifying PyFlyt to accommodate the specific physical properties of the Agilicious quadrotor. The main reason why PyFlyt was chosen over alternatives such as pybullet-drone, MuJoCo \cite{todorov2012mujoco}, AirSim \cite{shah2018airsim} and Flightmare \cite{song2021flightmare} is that it offers greater adaptability and customisation options. This allows us to modify the environment in aspects such as state and action spaces, physical properties, kinematics, and reward functions.

Additionally, the environment proposed by Agilicious, named Argviz, does not support reinforcement learning integration. Therefore, this study is the first to propose an open-source environment that builds upon PyFlyt for future reinforcement learning research on Agilicious models. To that end, we have pre-designed various tasks, including hovering, obstacle avoidance and trajectory tracking, to allow future deployments without extensive engineering overhead.

\subsubsection{PyBullet}
As its physics engine, PyFlyt utilises PyBullet, a free open-source simulator used in numerous reinforcement learning research over the past decade \cite{brunke2022safe, oroojlooy2023review, hanover2023autonomous}. It provides a fast real-time simulation built in C++ and thus supports realistic collisions and aerodynamics at a level other physics simulators may lack. The relative performance of PyBullet compared to other popular simulators has been demonstrated in prior works from Korber et al. \cite{korber2021comparing}.

\subsubsection{OpenAI Gym}
In addition to its adaptability, PyFlyt incorporates the OpenAI Gym interface, providing a user-friendly and training-efficient experience by allowing CUDA-based GPU usage. Moreover, this allows simultaneous and parallel execution of environments, resulting in fast and parallelised training. As a result, this is estimated to save us about half the training time and 20\% of the estimated energy consumption.
\chapter{Design}
\label{cha:design}

\section{Preliminaries and Problem Formulation}
\begin{figure}[h!]
  \centering
  \includegraphics[width=0.8\textwidth]{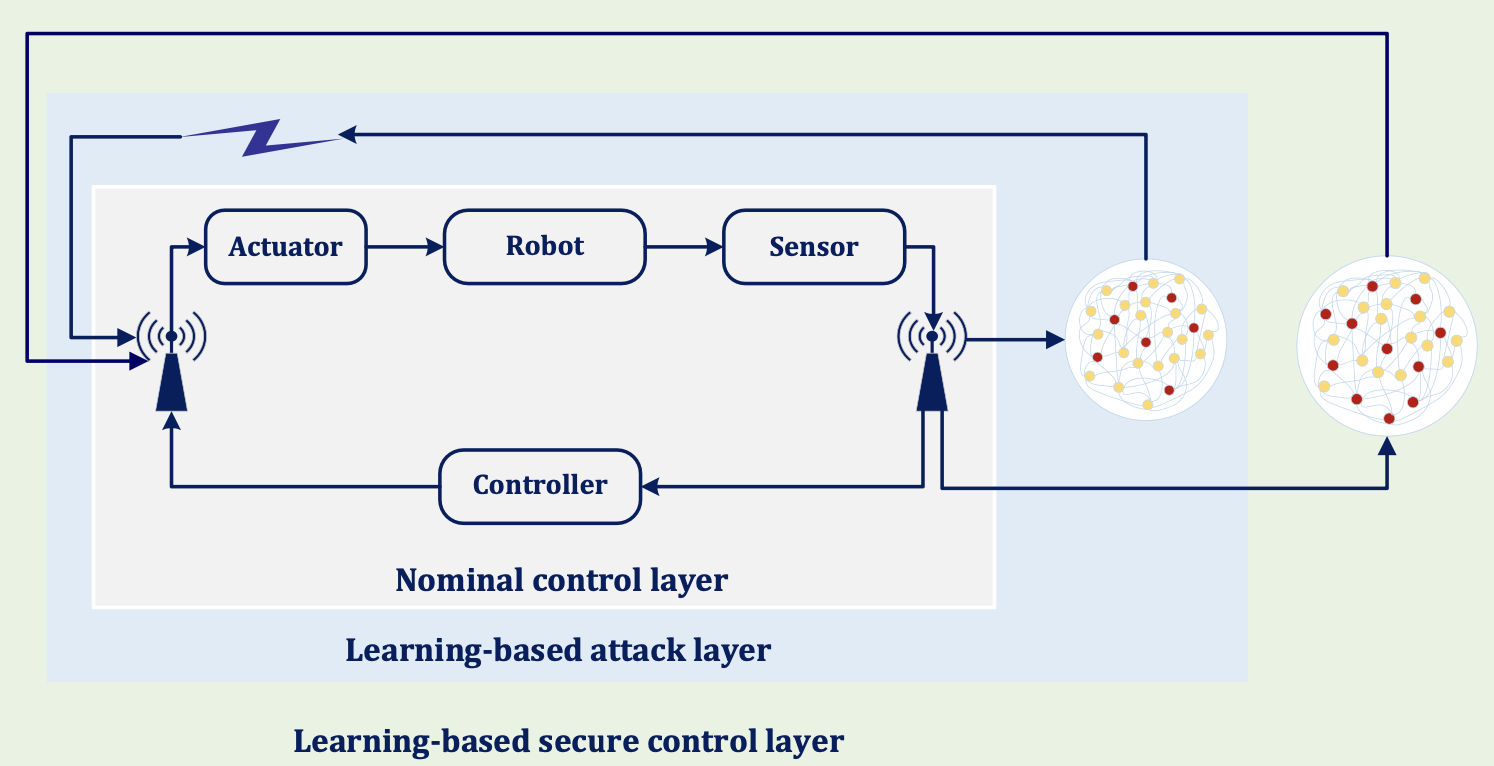}
  \caption[Proposed learning framework.]{Our proposed secure framework which includes (a) a nominal controller layer, (b) an optimal attack layer, and (c) a learning-based secure control layer. All three layers are designed using deep reinforcement learning.}
  \label{fig:framework}
\end{figure}

Our study mainly focuses on designing a secure learning framework for a quadrotor under malicious false-data injection attacks. The blueprint of our framework is described in Figure \ref{fig:framework}, where we introduce the three layers developed in this paper. The first is a \textit{nominal controller} responsible for controlling the quadrotor in total autonomy. The second is an \textit{attacker}, which monitors the sensor data transmitted on the communication network and learns optimal injection attacks to insert into the control signals of the nominal controller. The third is a learning-based \textit{secure control algorithm} designed to mitigate attacks by adjusting control signals within the secure control layer.

\begin{figure}[h!]
  \centering
  \includegraphics[width=0.5\textwidth]{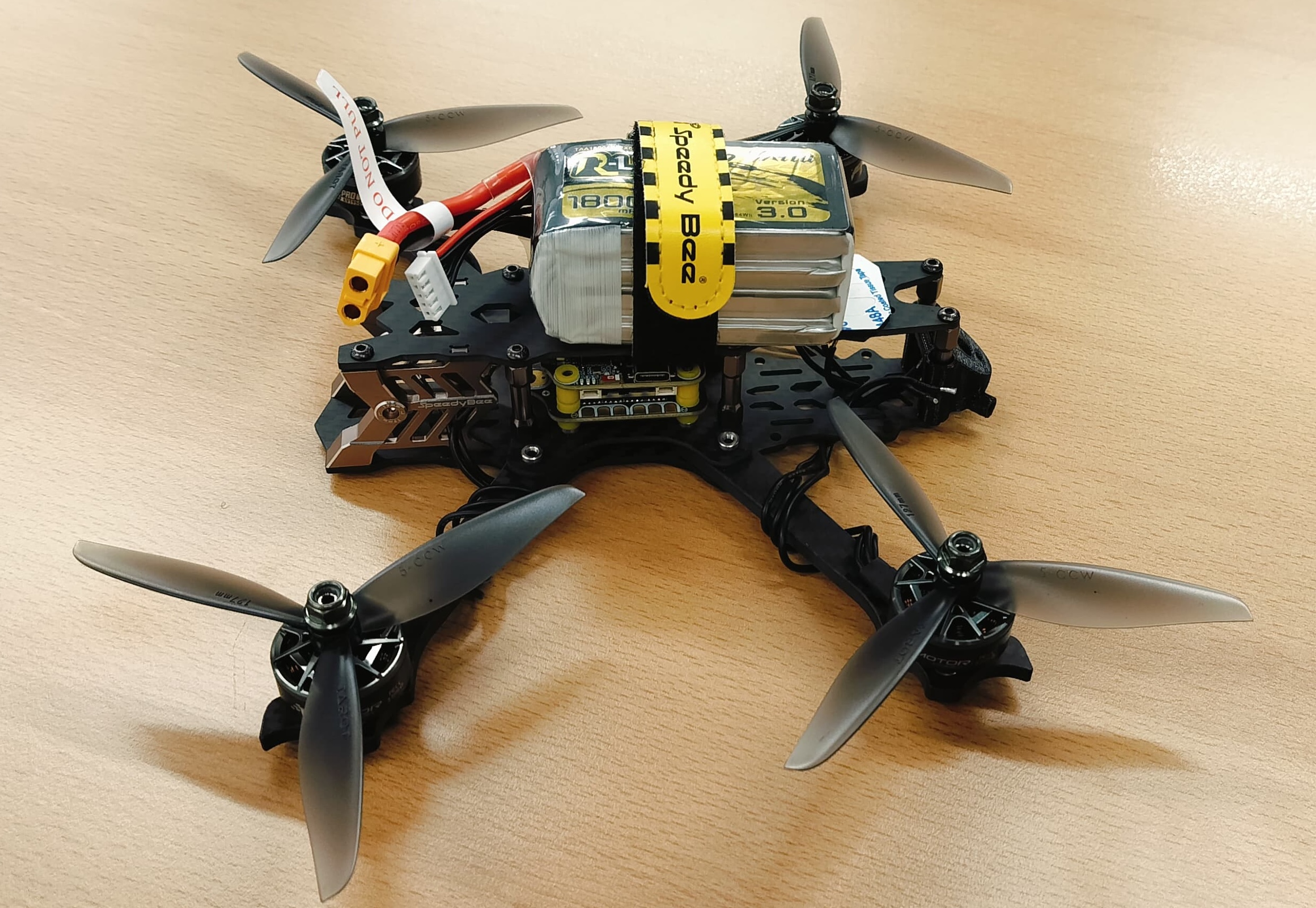}
  \caption{Agilicious: the main quadrotor used in this work.}
  \label{fig:agilicious-quad}
\end{figure}

We use an X-shaped quadrotor (displayed in Figure \ref{fig:agilicious-quad}) to describe our design process and the effectiveness of the proposed secure framework. Details on our quadrotor, including hardware, software stack and communication layers, are discussed in \ref{section:qaudrotor-models}. Specifically, we design a nominal controller trained using reinforcement learning to move the quadrotor towards a given location and hover. Furthermore, to address the secure control problem, we design both a malicious attacker and a mitigating defender using equivalent reinforcement learning approaches.

\subsection{Kinematic Model of our Quadrotor}
The motion of our quadrotor is defined as $[x,y,z,\phi,\theta,\psi]$, where $(x,y,z)$ represents the position of the quadrotor in the inertial frame, and $(\phi,\theta,\psi)$ its orientation in the body frame. Its kinematic model is described in subsection \ref{cha:quad-kinematics} as
\begin{equation} \label{eq:kinematics}
    [\Dot{x},\Dot{y},\Dot{z},\Dot{\phi},\Dot{\theta},\Dot{\psi}],
\end{equation}

which implies the collective thrust of the quadrotor $v$ and its body rates $[\Dot{\phi},\Dot{\theta},\Dot{\psi}]$.
We assume that the velocity $v$ satisfies $v_{max} < v < v_{min}$ with $v_{max}$ and $v_{min}$ the maximal and minimal velocities, respectively. For simplicity, as we do not aim to do acrobatic flying, the orientations of the quadrotor in its body frame are constrained to $\phi \in ]-\frac{\pi}{3},\frac{\pi}{3}[$, $\theta \in ]-\frac{\pi}{3},\frac{\pi}{3}[$, and $\psi \in ]-\frac{\pi}{3},\frac{\pi}{3}[$. 

For later calculations, we describe the virtual quadrotor generating the desired trajectories with
\begin{equation} \label{eq:reference}
\Dot{x}_r,\Dot{y}_r,\Dot{z}_r,\Dot{\phi}_r,\Dot{\theta}_r,\Dot{\psi}_r,
\end{equation}

and which follows the kinematic model described in \ref{section:bg-dynamics}.

\subsection{Attack Model}
\begin{figure}[h!]
  \centering
  \includegraphics[width=0.6\textwidth]{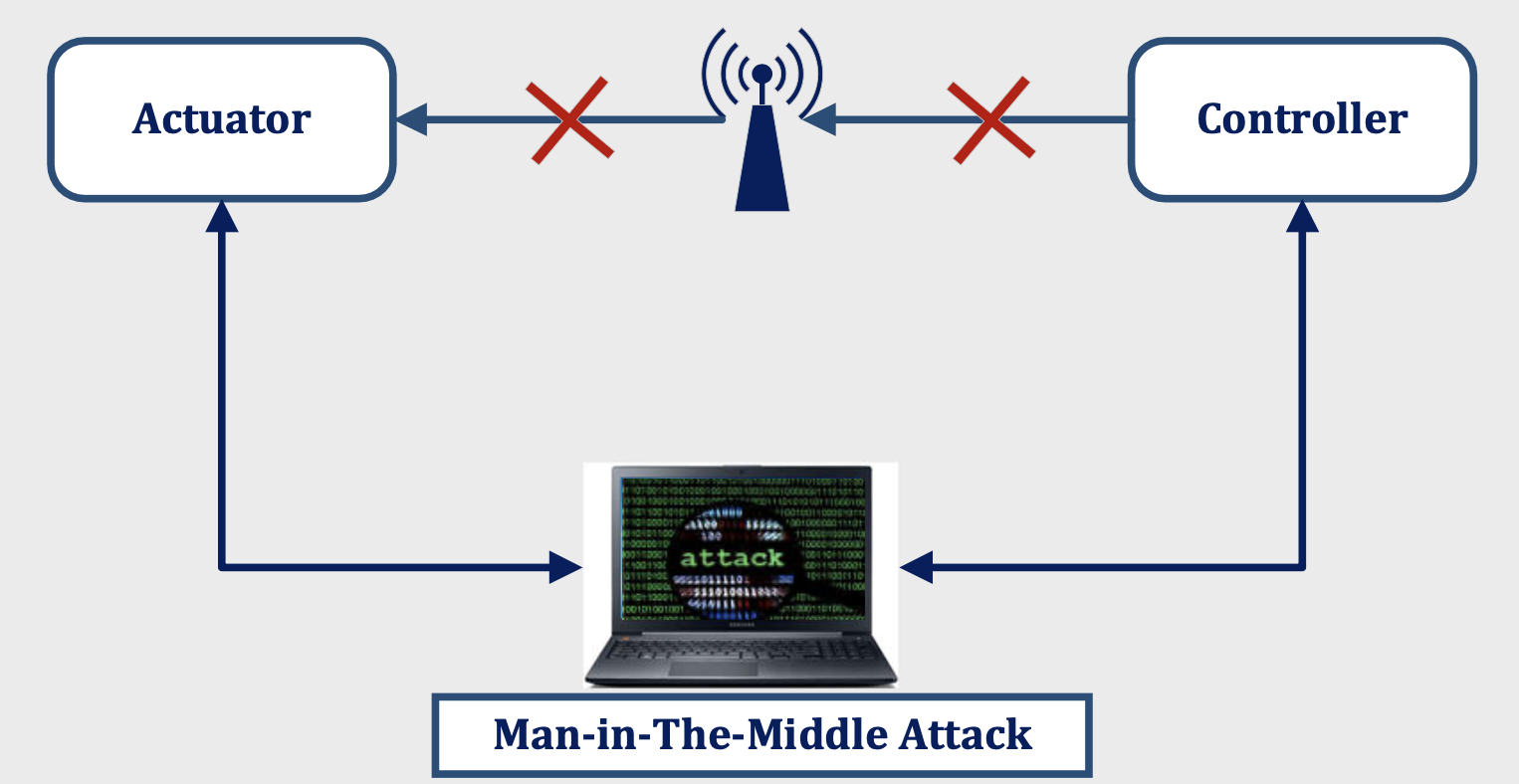}
  \caption[Illustration of a man-in-the-middle attack.]{Illustration of a man-in-the-middle attack. An adversary acts secretly as a middleman. It intercepts the signals sent from the controller to the quadrotor's actuators, modifies the commands, and sends them back to the actuators.}
  \label{fig:man-in-the-middle}
\end{figure}

As illustrated in Figure \ref{fig:man-in-the-middle}, a malicious adversary interrupts control commands sent from the nominal controller to the quadrotor's actuators via the communication network and injects false data to disrupt the quadrotor's trajectory. Under such attacks, the control commands are described as follows:
\begin{equation} \begin{aligned}
v(k) &= v_a(k) + v_b(k) \\
\Dot{\phi}(k) &= \Dot{\phi}_a(k) + \Dot{\phi}_b(k) \\
\Dot{\theta}(k) &= \Dot{\theta}_a(k) + \Dot{\theta}_b(k) \\
\Dot{\psi}(k) &= \Dot{\psi}_a(k) + \Dot{\psi}_b(k)
\end{aligned} \end{equation}

where $v_a(k)$, $\Dot{\phi}_a(k)$, $\Dot{\theta}_a(k)$ and $\Dot{\psi}_a(k)$ are the false data attacks designed in section \ref{section:false-data}; and $v_b(k)$, $\Dot{\phi}_b(k)$, $\Dot{\theta}_b(k)$ and $\Dot{\psi}_b(k)$ are control signals sent by the nominal controller designed in \ref{section:nominal}. 
We assume that the following knowledge is known to our adversary:
\begin{itemize}
    \item Adversaries inject malicious attacks without violating the physical constraints of the quadrotor, by which adversaries can both save attack energy and guarantee the effectiveness of attacks to some degree.
    \item Adversaries can access the communication network through the man-in-the-middle cyber-attack, for which a blueprint is given in Figure \ref{fig:man-in-the-middle}. That is, the adversaries can intercept the communication network between the actuators and the controller and secretly inject false data.
\end{itemize}

\subsection{Motivation Example}
This subsection provides numerical examples to show the necessity of protecting the quadrotor's control from deterioration. The experimental results are given in section \ref{section:evaluation}. We use the nominal controller from section \ref{section:nominal} to demonstrate the tracking performance of our quadrotor with and without attacks.

The initial state of the quadrotor is set as ($x=0, y=0, z=0.5, \phi=0^\circ, \theta=0^\circ, \psi=0^\circ$) and the hovering point at ($x_r=0.85, y_r=0.90, z_r=1.70$) with no constraint on the orientation.

\begin{figure}[ht]
    \centering
    \begin{subfigure}[b]{0.32\textwidth}
        \includegraphics[width=\textwidth]{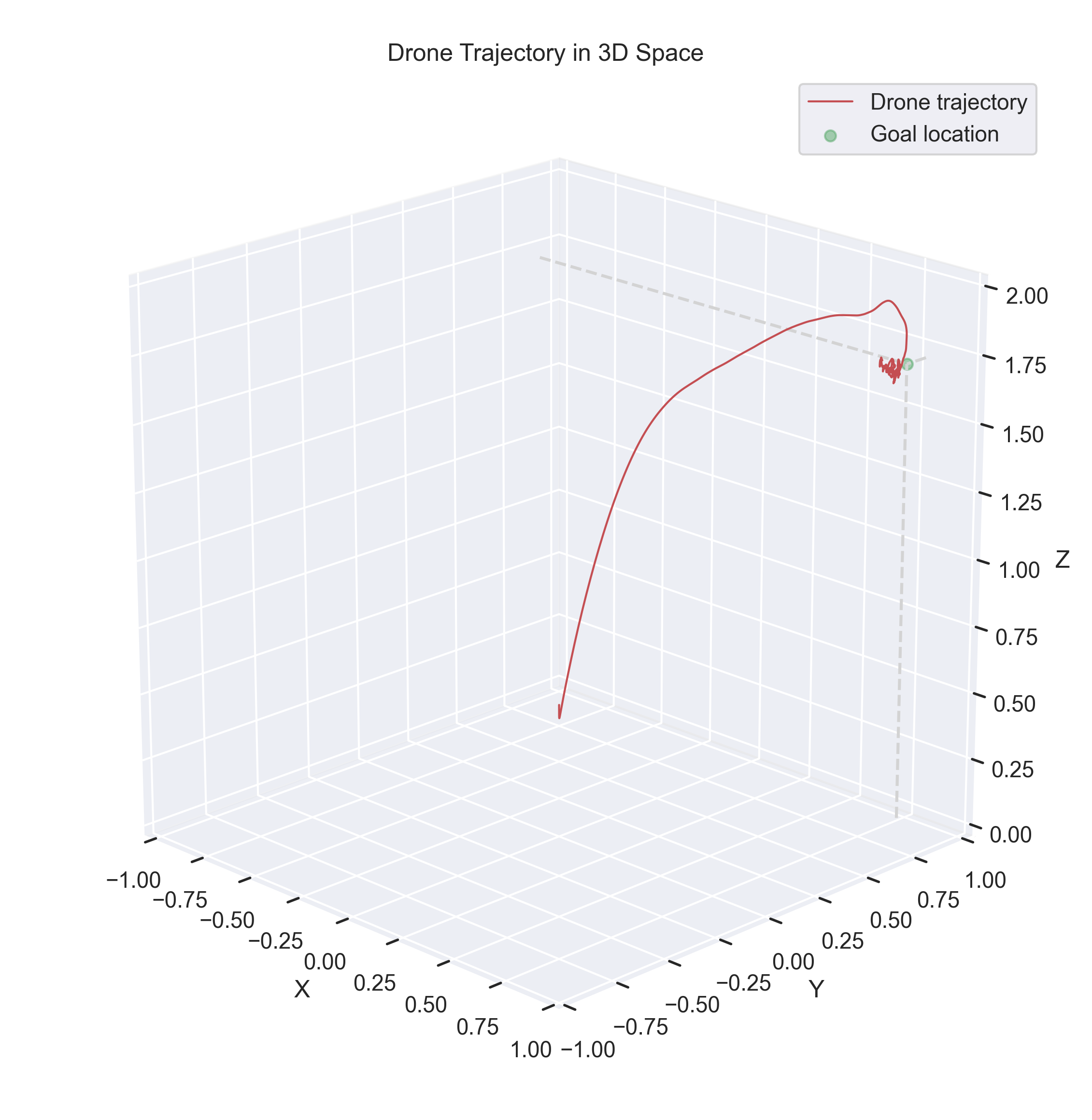}
        \caption{}
        \label{fig:nominal-0.85_0.9_1.7}
    \end{subfigure}
    \hfill
    \begin{subfigure}[b]{0.32\textwidth}
        \includegraphics[width=\textwidth]{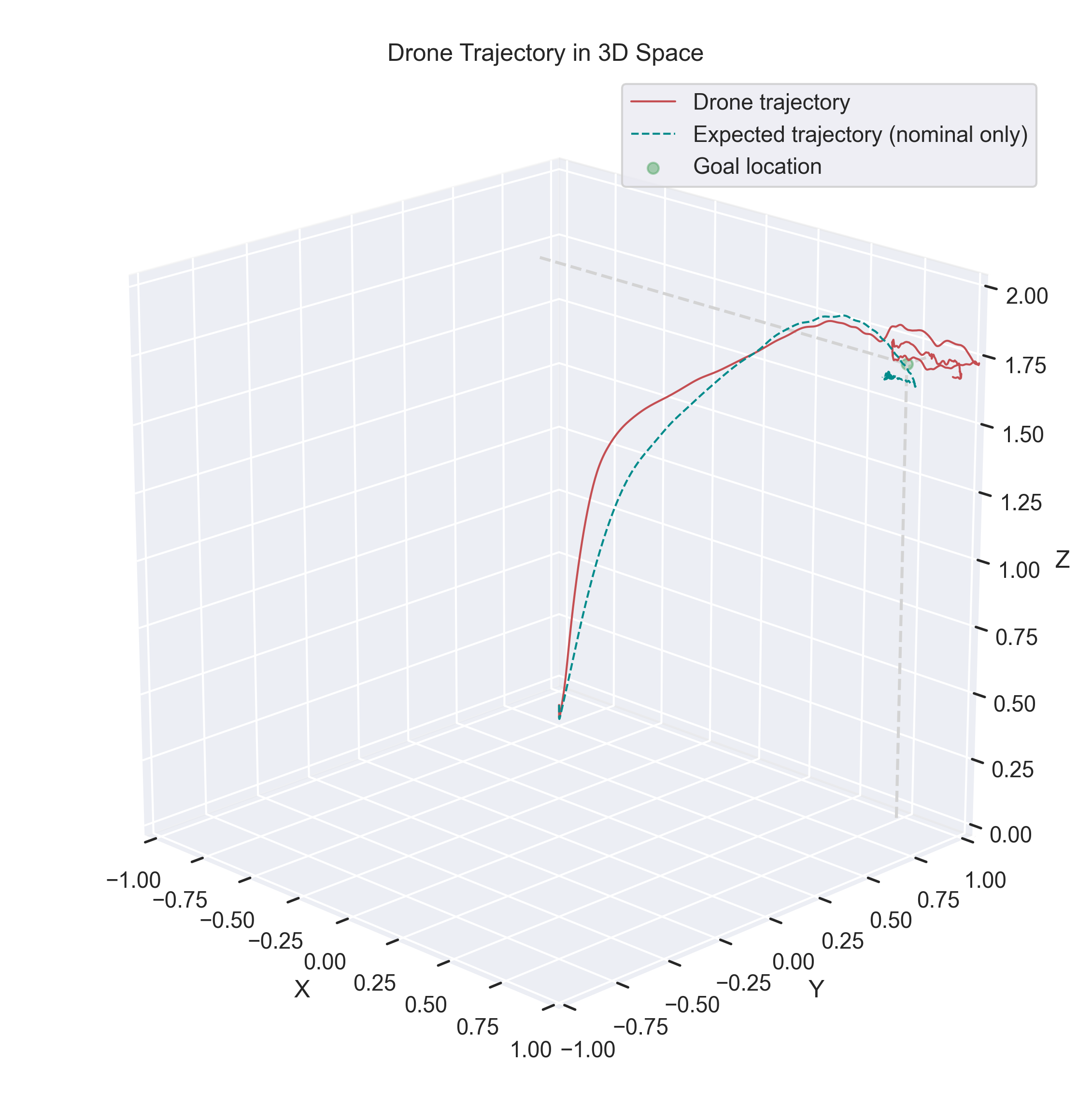}
        \caption{}
        \label{fig:attack_random-0.85_0.9_1.7}
    \end{subfigure}
    \hfill
    \begin{subfigure}[b]{0.32\textwidth}
        \includegraphics[width=\textwidth]{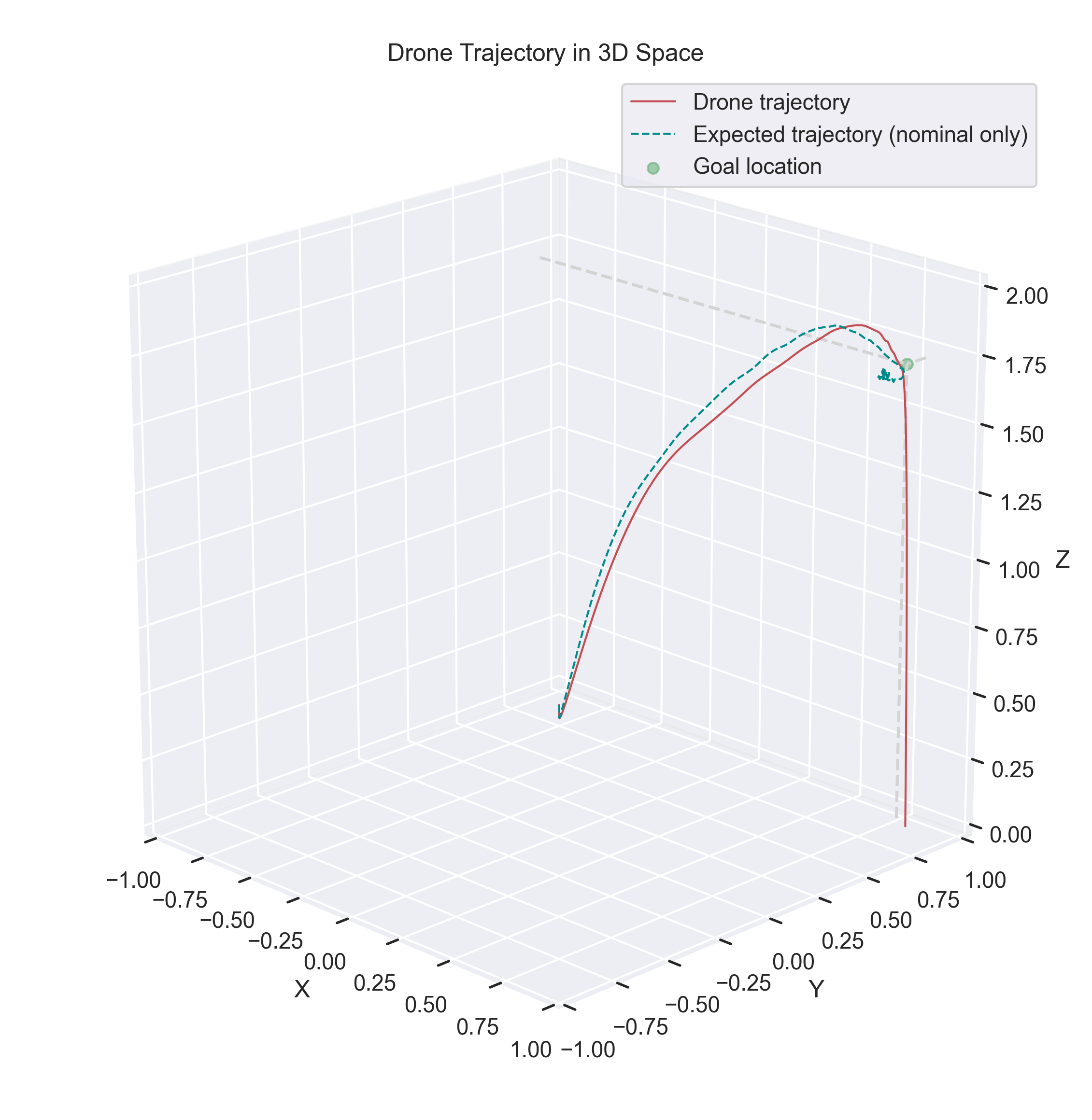}
        \caption{}
        \label{fig:attack_optimal--1.0_-1.0_0.5}
    \end{subfigure}

    \caption[Tracking performance of a quadrotor with vs. without attacks.]{Tracking performance of the quadrotor under (a) no malicious attack, (b) random injection attack, and (c) optimal false data injection attack designed in Algorithm \ref{algo:attacker}. The quadrotor is controlled by a nominal controller designed in Algorithm \ref{algo:nominal}.}
\end{figure}

The tracking trajectory of our quadrotor without any attack is displayed in Figure \ref{fig:nominal-0.85_0.9_1.7}. As a comparison, Figure \ref{fig:attack_random-0.85_0.9_1.7} shows the tracking trajectory of the quadrotor under \textit{random} false data injections. From these simulation results, we can conclude that a quadrotor is vulnerable to malicious adversary attacks. Therefore, providing solutions to prevent quadrotors’ performance from deteriorating is crucial.

Furthermore, as demonstrated in Figure \ref{fig:attack_optimal--1.0_-1.0_0.5}, the attack's performance depends on how adversaries utilise the available information. Comparing \ref{fig:attack_random-0.85_0.9_1.7} and \ref{fig:attack_optimal--1.0_-1.0_0.5} shows that learning-based \textit{optimal} attacks can achieve the best deterioration performance. Consequently, this motivates us to investigate the \textit{optimal} false data injection attack and defence countermeasure design.

\subsection{Problem Formulation}
As shown in the above examples and described in the literature review of section \ref{section:bg-cyber-attack}, quadrotors' performance can be compromised or even destroyed by cyber threats. Our work focuses on solving the following three problems:
\begin{itemize}
    \item How to create a reinforcement-learning-based control system capable of flying a quadrotor in complete autonomy;
    \item How to learn optimal false data injection attacks to deteriorate a quadrotor's performance and disturb its trajectory;
    \item How to design a secure control algorithm for quadrotor systems to mitigate false data injection attacks and recover flying abilities. 
\end{itemize}
The three problems are defined mathematically below. \\

\textit{Problem 1}. A nominal controller aims at generating control commands for the quadrotor to reach a desired location and hover at arrival. The control problem can be defined as follows:
\begin{equation}
    u_b(k) = arg\max_{u_b(k)} [\lim_{N \to T} \frac{1}{N} \sum_{k=1}^{N} \{ k - X^T(k)Q_bX(k) - \omega^T(k) \mathcal{L}_b \omega(k) + k\}],
\end{equation}

signifying that the nominal controller must \textbf{(a)} \textit{minimize} the difference to the reference $X^T(k)Q_bX(k)$ and \textbf{(b)} reach a stable hovering state $\omega_b^T(k) \mathcal{L}_b \omega_b(k)$ as quick as possible. Additionally, a constant $k=1.5$ is added to encourage survival of the quadrotor, i.e., rewarding the quadrotor for not failing over time.
This is subject to \[ \begin{aligned}
    x(k+1) &= x(k) + \Dot{x}(k) \Delta t, &
    y(k+1) &= y(k) + \Dot{y}(k) \Delta t, &
    z(k+1) &= z(k) + \Dot{z}(k) \Delta t, \\
    \phi(k+1) &= \phi(k) + \Dot{\phi}(k) \Delta t, &
    \theta(k+1) &= \theta(k) + \Dot{\theta}(k) \Delta t, &
    \psi(k+1) &= \psi(k) + \Dot{\psi}(k) \Delta t, \\
    && v_{min} &\leq v \leq v_{max},
\end{aligned} \]

where $x(k)$, $y(k)$, $z(k)$, $\phi(k)$, $\theta(k)$ and $\psi(k)$ are the states of the quadrotor under nominal controls, and $\Dot{x}(k)$, $\Dot{y}(k)$, $\Dot{z}(k)$, $\Dot{\phi}(k)$, $\Dot{\theta}(k)$ and $\Dot{\psi}(k)$ are its \textbf{global}\footnote{Global states do not vary whether they are under nominal, attacker or defender controls. They represent overall states of the quadrotor, i.e. the sum of all controllers in action at a given time.} states. 

The quadrotor actions and errors to reference are defined as
\begin{equation} \begin{aligned}
    u_b(k)     &= [ v_b(k), \Dot{\phi}_b(k), \Dot{\theta}_b(k), \Dot{\psi}_b(k) ]^T, \\
    v(k)       &= v_b(k), \Dot{\phi}(k) = \Dot{\phi}_b(k),
    \Dot{\theta}(k)  = \Dot{\theta}_b(k), 
    \Dot{\psi}(k) = \Dot{\psi}_b(k), \\
    X(k) &= [x(k) - x_r(k), y(k) - y_r(k), z(k) - z_r(k)], \\
    \omega(k) &= [\phi, \theta, \psi],
\end{aligned} \end{equation}

and $Q_b \geq 0$ and $\mathcal{L}_b \geq 0$ are weighting matrices. This is subject to the initial constraints on $\Dot{\phi}$, $\Dot{\theta}$, $\Dot{\psi}$, and $v_{min} \leq v(k) \leq v_{max}$.  \\

\textit{Problem 2}. An attacker intends to deteriorate the tracking performance by injecting false data attacks with minimal energy cost. The optimal attack problem can be defined as follows:
\begin{equation}
    u_a(k) = arg\max_{u_a(k)} [ \lim_{N\to T} \frac{1}{N} \sum_{k=1}^{N} \{\Bar{X}^T(k)Q_a\Bar{X}(k) - u_a^T(k)R_au_a(k)\} ],
\end{equation}

meaning that the optimal attack must \textbf{(a)} \textit{maximize} the error of the nominal model to the reference $\Bar{X}^T(k)Q_a\Bar{X}(k)$ while \textbf{(b)} \textit{minimizing} the cost of the attack $u_a^T(k)R_au_a(k)$ with $R_a$ defining the cost of each action in $u_a(k)$. Here, $T$ represents the length of a full episode.
This is subject to \begin{equation} \begin{aligned}
    \Bar{x}(k+1) &= \Bar{x}(k) + \Dot{x}(k) \Delta t, &
    \Bar{y}(k+1) &= \Bar{y}(k) + \Dot{y}(k) \Delta t, &
    \Bar{z}(k+1) &= \Bar{z}(k) + \Dot{z}(k) \Delta t, \\
    \Bar{\phi}(k+1) &= \Bar{\phi}(k) + \Dot{\phi}(k) \Delta t, &
    \Bar{\theta}(k+1) &= \Bar{\theta}(k) + \Dot{\theta}(k) \Delta t, &
    \Bar{\psi}(k+1) &= \Bar{\psi}(k) + \Dot{\psi}(k) \Delta t, \\
    && v_{min} &\leq v \leq v_{max},
\end{aligned} \end{equation}

where $\Bar{x}(k)$, $\Bar{y}(k)$, $\Bar{z}(k)$, $\Bar{\phi}(k)$, $\Bar{\theta}(k)$ and $\Bar{\psi}(k)$ are the states of the quadrotor under attack. The quadrotor actions and errors to reference are
\begin{equation} \begin{aligned}
    u_a(k)     &= [ v_a(k), \Dot{\phi}_a(k), \Dot{\theta}_a(k), \Dot{\psi}_a(k) ]^T, \\
    v(k)       &= v_a(k) + v_b(k), &\Dot{\phi}(k) &= \Dot{\phi}_a(k) + \Dot{\phi}_b(k), \\
    \Dot{\theta}(k)  &= \Dot{\theta}_a(k) + \Dot{\theta}_b(k), &\Dot{\psi}(k) &= \Dot{\psi}_a(k) + \Dot{\psi}_b(k), \\
    \Bar{X}(k) &= [\Bar{x}(k) - x_r(k), \Bar{y}(k) - y_r(k), \Bar{z}(k) - z_r(k)],
\end{aligned} \end{equation}

and $Q_a \geq 0$ and $R_a > 0$ are weighting matrices. This must also fulfil the initial constraints on $\Dot{\phi}$, $\Dot{\theta}$, $\Dot{\psi}$, and $v_{min} \leq v(k) \leq v_{max}$.  \\

\textit{Problem 3}. Under attacks, the defender’s objective is to find an optimal countermeasure to mitigate the attacks with minimal control cost. The optimal secure control framework is defined as follows:
\begin{equation}
    u_d(k) = arg\min_{u_d(k)} [ \lim_{N\to T} \frac{1}{N} \sum_{k=1}^{N} \{- \Bar{X}^T(k)Q_d\Bar{X}(k) - u_d^T(k)R_du_d(k)\} ],
\end{equation}

signifying that the optimal countermeasure must \textit{minimize} \textbf{(a)} the error of the nominal model to the reference $\Bar{X}^T(k)Q_d\Bar{X}(k)$ and \textbf{(b)} the cost of the attack $u_d^T(k)R_du_d(k)$ with $R_d$ defining the cost of each action in $u_d(k)$. 
This is subject to \begin{equation} \begin{aligned}
    \Tilde{x}(k+1) &= \Tilde{x}(k) + \Dot{x}(k) \Delta t, &
    \Tilde{y}(k+1) &= \Tilde{y}(k) + \Dot{y}(k) \Delta t, &
    \Tilde{z}(k+1) &= \Tilde{z}(k) + \Dot{z}(k) \Delta t, \\
    \Tilde{\phi}(k+1) &= \Tilde{\phi}(k) + \Dot{\phi}(k) \Delta t, &
    \Tilde{\theta}(k+1) &= \Tilde{\theta}(k) + \Dot{\theta}(k) \Delta t, &
    \Tilde{\psi}(k+1) &= \Tilde{\psi}(k) + \Dot{\psi}(k) \Delta t, \\
    && v_{min} &\leq v \leq v_{max},
\end{aligned} \end{equation}

where $\Tilde{x}(k)$, $\Tilde{y}(k)$, $\Tilde{z}(k)$, $\Tilde{\phi}(k)$, $\Tilde{\theta}(k)$ and $\Tilde{\psi}(k)$ are the states of the quadrotor under attack. The quadrotor actions and errors to reference are
\begin{equation} \begin{aligned}
    u_d(k)     &= [ v_d(k), \Dot{\phi}_d(k), \Dot{\theta}_d(k), \Dot{\psi}_d(k) ]^T, \\
    v(k)       &= v_a(k) + v_b(k) + v_d(k), &\Dot{\phi}(k) &= \Dot{\phi}_a(k) + \Dot{\phi}_b(k) + \Dot{\phi}_d(k), \\
    \Dot{\theta}(k)  &= \Dot{\theta}_a(k) + \Dot{\theta}_b(k) + \Dot{\theta}_d(k), &\Dot{\psi}(k) &= \Dot{\psi}_a(k) + \Dot{\psi}_b(k) + \Dot{\psi}_d(k), \\
    \Tilde{X}(k) &= [\Tilde{x}(k) - x_r(k), \Tilde{y}(k) - y_r(k), \Tilde{z}(k) - z_r(k)],
\end{aligned} \end{equation}

and $Q_d \geq 0$ and $R_d > 0$ are weighting matrices. This is once again subject to the initial constraints on $\Dot{\phi}$, $\Dot{\theta}$, $\Dot{\psi}$, and $v_{min} \leq v(k) \leq v_{max}$.

\section{Control System for an Autonomous Quadrotor} \label{section:nominal}
In this section, we learn an optimal control system to fly our quadrotor in total autonomy. While one can design such a controller following the control theory, e.g. utilising a PD or MPC controller, we decide to take a reinforcement learning approach to solve \textit{Problem 1}. By taking this approach, we aim to contribute to current research on learning-based controllers and provide comparative resources against mathematical systems.

\subsection{Environment for a Quadrotor under Nominal Control} \label{section:mdp-nominal}
As defined in subsection \ref{section:mdp}, a reinforcement learning setup includes an agent and an environment, which interact to improve the agent's capabilities. The environment here is defined by a Markov decision process as a tuple $(\mathcal{S}, \mathcal{A}, \mathcal{P}, \mathcal{R}, \gamma)$ where $\mathcal{S}$ is the state space, $\mathcal{A}$ is the action space, $\mathcal{P}$ is the transition probability matrix from a state-action pair at time $t$ onto a distribution of possible states at time $t + 1$, $\mathcal{R}$ is the immediate reward function, and $\gamma \in [0, 1)$ is the discount factor.
Given that the quadrotor is controlled by the nominal controller, we describe the MDP as:
\begin{equation}
    s(k+1) \sim \mathcal{P}(s(k+1) | s(k), u_b(k)),
\end{equation}

meaning the transition probability from $s(k)$ to $s(k+1)$ under the nominal controller $u_b(k)$, where $s = [x(k), y(k), z(k), \phi(k), \theta(k), \psi(k)]$ with $s \in \mathcal{S}$, and $u_b(k) \in \mathcal{A}$. \\

\textbf{More details on the environment.} We describe below the MDP defining our nominal environment in more details, including vectorial and constraint definitions.
\begin{itemize}
    \item The \textit{action space} $\mathcal{A} = [v_b, \Dot{\phi}_b, \Dot{\theta}_b, \Dot{\psi}_b]$ with $v \in [0, 3.5]$ and $\Dot{\phi}, \Dot{\theta}, \Dot{\psi} \in [-\frac{\pi}{3}, \frac{\pi}{3}]$;
    
    \item The \textit{state space} $\mathcal{S} = [\Dot{\phi}, \Dot{\theta}, \Dot{\psi}, \phi, \theta, \psi, \Dot{x}, \Dot{y}, \Dot{z}, x, y, z, \Dot{\phi}_{\text{prev}}, \Dot{\theta}_{\text{prev}}, \Dot{\psi}_{\text{prev}}, x_h, y_h, z_h]$, where $\Dot{\phi}_{\text{prev}}, \Dot{\theta}_{\text{prev}}, \Dot{\psi}_{\text{prev}}$ are previous actions, and $x_h, y_h, z_h$ are hover point coordinates;

    \item The \textit{reward function} $\mathcal{R}$ is defined as $\mathcal{R}(k) = k - X^T(k)Q_bX(k) - \omega^T(k) \mathcal{L}_b \omega(k)$\\according to \textit{Problem 1}, where $Q_b \geq 0$ and $\mathcal{L}_b \geq 0$ are weighting matrices.
\end{itemize}

\subsection{Objectives in the Learning of a Nominal Controller}
Through reinforcement learning, we want to \textit{minimize} the action-value function\\$Q_{\pi_b}(s(k), u_b(k))$, where $\pi_b$ denotes the nominal policy to learn which finds an optimal function that maps a state and an action to the expected future reward of taking that action. If we can obtain such a function which ensures that the expected future reward is accurate to the true outcome, we can greedily select an action at time $t$ according to $\pi_b$ and guarantee the best outcome in the future.
The expression of $Q_{\pi_b}(s(k), u_b(k))$ is computed as
\begin{equation} \label{eq:Q-pi-b}
    Q_{\pi_b}(s(k), u_b(k)) = \gamma \mathbb{E}_{s(k+1)}[V_{\pi_b}(s(k+1))] + \mathcal{R}(k),
\end{equation}

where $\gamma \mathbb{E}_{s(k+1)}[V_{\pi_b}(s(k+1))]$ is the expected cumulative reward from state $s(k+1)$, given that action $u_b(k)$ was taken on state $s(k)$ according to the probability distribution $\mathcal{P}_{k+1|k} = \mathcal{P}(s(k+1)|s(k), u_b(k))$ of possible resulting states $s(k+1)$. $\mathcal{R}(k)$ is the immediate reward for state $s(k)$. $\pi_b(u_b(k)|s(k))$ gives the probability of choosing the command $u_b(k)$ at state $s(k)$ according to the nominal policy $\pi_b$.
Moreover,
\begin{equation} \label{eq:V-pi-b} \begin{aligned}
    V_{\pi_b}(s(k)) &= \sum_k^T \sum_{u_b(k)}^{} [\pi_b(u_b(k)|s(k)) \sum_{s(k+1)}^{} \mathcal{P}_{k+1|k} \times (\mathcal{R}(k) + \gamma V_{\pi_b}(s(k+1)))] \\
    &= \sum_k^T \sum_{u_b(k)}^{} [\pi_b(u_b(k)|s(k)) \mathbb{E}_{(k+1|k) \sim \mathcal{P}}[\mathcal{R}(k) + \gamma V_{\pi_b}(s(k+1))]] 
\end{aligned} \end{equation}

is the state estimation function, which represents the probability of taking each action $a_t$ from state $s_t$ using policy $\pi_b$, multiplied by the expected cumulative reward of taking that action $\mathbb{E}_{(k+1|k) \sim \mathcal{P}}[\mathcal{R}(k) + \gamma V_{\pi_b}(s(k+1))]$ (probability of each possible state $s(k+1)$ from transition probabilities $\mathcal{P}$ multiplied by its expected cumulative reward); summed over all time step $k$ of the episode of length $T$.

Therefore, by finding a solution to the following optimal problem (\ref{eq:nominal-solution}),\\ \textit{Problem 1} can be solved:
\begin{equation}
\label{eq:nominal-solution}
    \pi_b^* = arg\max_{\pi_b} Q_{\pi_b}(s(k), u_b(k)),
\end{equation}

where $\pi_b^*$ is the optimal control policy, and $u_b(k)$ samples from that nominal policy. \\

\textbf{Remark 1.} \textit{One might wonder how the controller knows the position of the hover point it should reach. This is done by artificially including the desired hover point's coordinates, denoted by the subset $[x_h, y_h, z_h]$, into its observation states at each timestep.}

\subsection{Autonomous Control Algorithm}
Next, the proximal policy optimization algorithm is introduced to learn the optimal attack policy $\pi_b^*$ and value-function $V_{\pi_b^*}$. We use a shared neural network architecture for both the policy and value-function to learn.
According to subsection \ref{section:ppo}, this reinforcement learning algorithm minimizes the loss 
\begin{equation}
    L_t^{\text{CLIP+VF+S}}(\theta) = \hat{\mathbb{E}}_t \left[ L_t^{\text{CLIP}}(\theta) - c_1 L_t^{\text{VF}}(\theta) + c_2 S[\pi_{\theta_{old}}(s_t)] \right],
\end{equation}

explained in \ref{section:ppo}. In the proximal policy optimization algorithm, the evaluation and improvement steps are executed every $T$ timesteps to learn the optimal control policy $\pi_b^*$. Note that $\mathcal{S}$ guarantees that the entropy of action is maximized, i.e., pushes $\pi_b^*$ to produce a fair amount of exploration.

In every iteration, each $N$ parallel actor collects $T$ timesteps of data. Then, we evaluate the policy based on the surrogate loss $L_t$ on these $NT$ timesteps of data and improve that policy through the Adam \cite{Kingma2014AdamAM} optimizer, which follows a gradient descent approach over $K$ epochs.

According to the above explanations and Algorithm \ref{algo:ppo} defined in \ref{section:ppo}, we derive Algorithm \ref{algo:nominal} to train our deep neural network and learn an optimal nominal control to fly our quadrotor in complete autonomy. \\

\textbf{Remark 2.} \textit{The hover position varies with each training episode to prevent the nominal controller from becoming overly specialized in navigating to a unique hover position in space, i.e. overfitting to its reward. This strategy ensures that the controller actually learns how to ``fly'' towards a given point instead of learning to converge to a static point through the experimentation of arbitrary sequences of actions during training. Furthermore, its spawn (or initial) position is altered at each training episode to guarantee the controller’s ability to fly arbitrarily in three-dimensional space.}

\begin{algorithm}
\caption{Learning-based autonomous control algorithm}
\label{algo:nominal}
\begin{algorithmic}[1]
\FOR{iteration = 1,2,\dots}
    \STATE Initialize replay memory buffer $\mathcal{M}_b$
    \FOR{actor = 1,2,\dots,N}
        \FOR{each data collection step of $T$ total timesteps}
            \STATE Sample $u_b(k)$ from the nominal controller
            \STATE $u(k) \leftarrow u_b(k)$
            \STATE Update the memory $\mathcal{M}_b$ $\leftarrow$ $\mathcal{M}_b \cup < s_t, u_b(k), r_t, V_{\pi_{b_\text{old}}}(s_t), \pi_{b_\text{old}}(a_t|s_t) >$
            \STATE Take the generated action $u(k)$ in the environment
            \IF{episode terminated}
                \STATE Change hovering point coordinates $[x_h, y_h, z_h]$ and quadrotor's\\initial position to prevent overfitting
            \ENDIF
    \ENDFOR 
    \STATE Compute advantage estimates $\hat{A}_1,\dots,\hat{A}_T$
    \ENDFOR
    \STATE Optimize surrogate $L^{\text{CLIP+VF+S}}$ wrt $\pi_b$, with $K$ epochs and minibatch\\size $M \leq NT$ from $\mathcal{M}_b$
    \STATE $\pi_{b_\text{old}} \leftarrow \pi_b$
\ENDFOR
\end{algorithmic}
\end{algorithm}

\section{Optimal False Data Injection Scheduling}
\label{section:false-data}
In this section, we focus on solving \textit{Problem 2} by proposing a learning framework to design the optimal false data injection attacks and deteriorate the performance of a quadrotor under nominal controls.
Many reinforcement learning algorithms have already been used to similar ends, such as the trust region policy optimization \cite{schulman2015trust} and soft actor-critic \cite{haarnoja2018soft} algorithms. However, we decided to use the proximal policy optimization algorithm (PPO) described in section \ref{section:ppo} to solve the optimal attack problem. This choice is motivated by the advantages of PPO over alternative algorithms, as described by OpenAI in a report stating that ``PPO has become the default reinforcement learning algorithm at OpenAI because of its ease of use and good performance.''.

\subsection{Markov Decision Process of a Quadrotor under Attack}
\label{section:mdp-attack}
Similarly to the description in Subsection \ref{section:mdp-nominal}, the Markov decision process of a quadrotor under optimal false data injections consists of five elements, that is, the state space $\Bar{\mathcal{S}}$, the action space $\Bar{\mathcal{A}}$, the transition probability matrix $\Bar{\mathcal{P}}$, the immediate reward function $\Bar{\mathcal{R}}$, and the discount factor $\Bar{\gamma} \in [0, 1)$. However, note that the action space is modified to $\Bar{\mathcal{A}} = [v_a, \Dot{\phi}_a, \Dot{\theta}_a, \Dot{\psi}_a]$ with $v(k) \in [0, 3.5]$ and $\Dot{\phi}, \Dot{\theta}, \Dot{\psi} \in [-\frac{\pi}{3}, \frac{\pi}{3}]$. The involution of such a process is as follows:
\begin{equation}
    \Bar{s}(k+1) \sim \Bar{\mathcal{P}}(\Bar{s}(k+1) | \Bar{s}(k), u_a(k)),
\end{equation}

where $\Bar{s} = [\Bar{x}(k), \Bar{y}(k), \Bar{z}(k), \Bar{\phi}(k), \Bar{\theta}(k), \Bar{\psi}(k)]$ with $\Bar{s} \in \Bar{\mathcal{S}}$, and $u_a(k) \in \Bar{\mathcal{A}}$.

\subsection{Learning-based Objectives for an Optimal Attacker}
This subsection defines the attack cost and action-value function, following which the attack policy to learn is derived. The attacker aims to inject false data into the nominal control commands to optimally deteriorate the quadrotor’s tracking performance and trajectory with minimal control cost. Therefore, the reward function $\Bar{\mathcal{R}}(k)$ is defined according to \textit{Problem 2} as
\begin{equation}
    \Bar{\mathcal{R}}(k) = \Bar{\mathcal{X}}^T(k)Q_a\Bar{\mathcal{X}}(k) - u_a^T(k)R_au_a(k),
\end{equation}

where $Q_a \geq 0$ and $R_a > 0$ are weighting matrices.

Based on the definition of $\Bar{\mathcal{R}}(k)$, the attacker's objective is to \textit{maximize} the action-value function $Q_{\pi_a}(\Bar{s}(k), u_a(k))$, where $\pi_a$ denotes the attack policy to be learned. The definitions of $Q_{\pi_a}(\Bar{s}(k), u_a(k))$ and $V_{\pi_a}(\Bar{s}(k))$ are given below, but explanations are omitted due to similarity with the previous definitions \ref{eq:Q-pi-b} and \ref{eq:V-pi-b}.
\begin{equation} \label{eq:Q-V-pi-a} \begin{aligned}
    Q_{\pi_a}(\Bar{s}(k), u_a(k)) &= \gamma \mathbb{E}_{\Bar{s}(k+1)}[V_{\pi_a}(\Bar{s}(k+1))] + \Bar{\mathcal{R}}(k) \\
    V_{\pi_a}(\Bar{s}(k)) &= \sum_k^T \sum_{u_a(k)}^{} [\pi_a(u_a(k)|\Bar{s}(k)) \sum_{\Bar{s}(k+1)}^{} \Bar{\mathcal{P}}_{k+1|k} \times (\Bar{\mathcal{R}}(k) + \gamma V_{\pi_a}(\Bar{s}(k+1)))] \\
    &= \sum_k^T \sum_{u_a(k)}^{} [\pi_a(u_a(k)|\Bar{s}(k)) \mathbb{E}_{(k+1|k) \sim \Bar{\mathcal{P}}}[\Bar{\mathcal{R}}(k) + \gamma V_{\pi_a}(\Bar{s}(k+1))]] 
\end{aligned} \end{equation}

If we can find a solution to the following optimization problem (\ref{eq:att-solution}), \textit{Problem 2} can be solved:
\begin{equation} \label{eq:att-solution}
    \pi_a^* = arg\max_{\pi_a} Q_{\pi_a}(\Bar{s}(k), u_a(k)),
\end{equation}

where $\pi_a^*$ is the optimal attack policy, and $u_a(k)$ samples from this optimal policy.

\subsection{Learning-based False Data Injection Algorithm}
We employ reinforcement learning following the proximal policy optimization algorithm to learn the optimal attack policy $\pi_a^*$ and value-function $V_{\pi_a^*}$. We maintain a shared neural network architecture for both the policy and value-function to learn, and train it by minimizing the same loss as the nominal controller
\begin{equation} \label{eq:att-ppo-loss}
    L_t^{\text{CLIP+VF+S}}(\theta) = \hat{\mathbb{E}}_t \left[ L_t^{\text{CLIP}}(\theta) - c_1 L_t^{\text{VF}}(\theta) + c_2 S[\pi_{\theta_{old}}(s_t)] \right],
\end{equation}

explained in subsection \ref{section:ppo}. Note that $\mathcal{S}$ still guarantees sufficient exploration.

The parameters of our policy network are trained using Algorithm \ref{algo:attacker}, proposed below. Once the network has been successfully trained, false data injection attacks sampled from policy $\pi_a$ can be applied to deteriorate the quadrotor's tracking performances. \\

\textbf{Remark 3.} \textit{Similarly to Remark 2, variations are introduced in the desired hover point $[x_h, y_h, z_h] \in \Bar{\mathcal{S}}$ and initial quadrotor's position. However here, this is made to learn optimal false data injections over many different trajectories.}

\begin{algorithm}
\caption{Learning-based false data injection attack algorithm}
\label{algo:attacker}
\begin{algorithmic}[1]
\FOR{iteration = 1,2,\dots}
    \STATE Initialize replay memory buffer $\mathcal{M}_a$
    \FOR{actor = 1,2,\dots,N}
        \FOR{each data collection step of $T$ total timesteps}
            \STATE Sample $u_b(k)$ and $u_a(k)$ from the nominal controller's policy $\pi_b$\\and attack policy $\pi_{a_\text{old}}(\cdot|\Bar{s}_t)$ respectively
            \STATE $u(k) \leftarrow u_b(k) + u_a(k)$
            \STATE Update the memory $\mathcal{M}_a$ $\leftarrow$ $\mathcal{M}_a \cup < \Bar{s}_t, u_a(k), r_t, V_{\pi_{a_\text{old}}}(\Bar{s}_t), \pi_{a_\text{old}}(a_t|\Bar{s}_t) >$
            \STATE Take combined action $u(k)$
            \IF{episode terminated}
                \STATE Change hovering point coordinates $[x_h, y_h, z_h]$ and quadrotor's\\initial position to prevent overfitting
            \ENDIF
    \ENDFOR 
    \STATE Compute advantage estimates $\hat{A}_1,\dots,\hat{A}_T$
    \ENDFOR
    \STATE Optimize surrogate $L^{\text{CLIP+VF+S}}$ wrt $\pi_a$, with $K$ epochs and minibatch\\size $M \leq NT$ from $\mathcal{M}_a$
    \STATE $\pi_{a_\text{old}} \leftarrow \pi_a$
\ENDFOR
\end{algorithmic}
\end{algorithm}

\section{Learning-based Countermeasure} \label{section:countermeasure}
From an adversary’s perspective, a quadrotor’s tracking performance can be deteriorated by injecting falsified commands generated using Algorithm \ref{algo:attacker}. In this section, we provide a solution to \textit{Problem 3}, i.e., a learning-based secure control algorithm to stabilise the quadrotor and mitigate the malicious attacks. Moreover, the proximal policy optimisation algorithm is modified to account for the attacker’s injections on top of the nominal control commands.

\subsection{Markov Decision Process of a Secured Quadrotor} \label{section:mdp-defend}
Similarly to the attacker's environment described in Section \ref{section:mdp-attack}, the Markov decision process of a quadrotor with a secure controller is a tuple where $\Tilde{\mathcal{S}}$ is the state space, $\Tilde{\mathcal{A}}$ is the action space, $\Tilde{\mathcal{P}}$ is the transition probability matrix, $\Tilde{\mathcal{R}}$ is the immediate reward function, and $\Tilde{\gamma} \in [0,1)$ is the discount factor. Carefully note the tilde instead of the bar over these elements.
This time, the action space is modified to $\Tilde{\mathcal{A}} = [v_d, \Dot{\phi}_d, \Dot{\theta}_d, \Dot{\psi}_d]$ with $v(k) \in [0, 3.5]$ and $\Dot{\phi}, \Dot{\theta}, \Dot{\psi} \in [-\frac{\pi}{3}, \frac{\pi}{3}]$. The involution of such a process is as follows:
\begin{equation}
    \Tilde{s}(k+1) \sim \Tilde{\mathcal{P}}(\Tilde{s}(k+1) | \Tilde{s}(k), u_d(k)),
\end{equation}

where $\Tilde{s} = [\Tilde{x}(k), \Tilde{y}(k), \Tilde{z}(k), \Tilde{\phi}(k), \Tilde{\theta}(k), \Tilde{\psi}(k)]$ with $\Tilde{s}(k) \in \Tilde{\mathcal{S}}$, and $u_d(k) \in \Tilde{\mathcal{A}}$.

\subsection{Learning-based Objectives for an Optimal Countermeasure}
This subsection describes the objectives in designing a secure control algorithm to mitigate false data injection attacks learned using Algorithm \ref{algo:attacker}.

The defender's objective is to use minimal control cost to recover the tracking performance disturbed by the attacker. Therefore, we define the reward $\Tilde{\mathcal{R}}(k)$ as
\begin{equation}
    \Tilde{R}(k) = - \Tilde{X}^T(k)Q_d\Tilde{X}(k) - u_d^T(k)R_du_d(k),
\end{equation}

where $Q_d \geq 0$ and $R_d > 0$ are weighting matrices.

Based on the definition of $\Tilde{R}(k)$, the defender's objective is to \textit{minimize} the action-value function $Q_{\pi_d}(\Tilde{s}(k), u_d(k))$, with $\pi_d$ being the secure policy to be learned. The definitions of $Q_{\pi_d}(\Tilde{s}(k), u_d(k))$ and $V_{\pi_d}(\Tilde{s}(k))$ are omitted due to similarity with the previous definitions from \ref{eq:Q-V-pi-a}.

Therefore, by finding a solution to the following optimal problem (\ref{eq:def-solution}),\\ \textit{Problem 3} can be solved:
\begin{equation} \label{eq:def-solution}
    \pi_d^* = arg\min_{\pi_d}Q_{\pi_d}(\Tilde{s}(k), u_d(k)),
\end{equation}

where $\pi_d^*$ is the optimal secure policy and $u_d(k)$ samples from this optimal policy.

\subsection{Learning-based Secure Control Algorithm}
The proximal policy optimization algorithm is again employed to learn the optimal countermeasure as a policy $\pi_d^*$ and value-function $V_{\pi_d^*}$. We maintain a shared neural network architecture for both the policy and value-function to learn, and train it by minimizing the loss in Eq. \ref{eq:att-ppo-loss}.
According to the objectives defined above, Algorithm \ref{algo:defender} is derived to learn the optimal secure control policy $\pi_d^*$ from which secure control commands can be sampled to recover the quadrotor's tracking performance.

\begin{algorithm}
\caption{Learning-based secure control algorithm}
\label{algo:defender}
\begin{algorithmic}[1]
\FOR{iteration = 1,2,\dots}
    \STATE Initialize replay memory buffer $\mathcal{M}_d$
    \FOR{actor = 1,2,\dots,N}
        \FOR{each data collection step of $T$ total timesteps}
            \STATE Sample $u_b(k)$, $u_a(k)$ and $u_d(k)$ from the nominal controller's policy\\$\pi_b$, attacker's policy $\pi_a$ and current policy $\pi_{d_\text{old}}(\cdot|\Tilde{s}_t)$ respectively
            \STATE $u(k) \leftarrow u_b(k) + u_a(k) + u_d(k)$
            \STATE Update the memory $\mathcal{M}_d$ $\leftarrow$ $\mathcal{M}_d \cup < \Tilde{s}_t, u_d(k), r_t, V_{\pi_{d_\text{old}}}(\Tilde{s}_t), \pi_{d_\text{old}}(a_t|\Tilde{s}_t) >$
            \STATE Take combined action $u(k)$
            \IF{episode terminated}
                \STATE Change hovering point coordinates $[x_h, y_h, z_h]$ and quadrotor's\\initial position to prevent overfitting
            \ENDIF
    \ENDFOR 
    \STATE Compute advantage estimates $\hat{A}_1,\dots,\hat{A}_T$
    \ENDFOR
    \STATE Optimize surrogate $L^{\text{CLIP+VF+S}}$ wrt $\pi_d$, with $K$ epochs and minibatch\\size $M \leq NT$ from $\mathcal{M}_d$
    \STATE $\pi_{d_\text{old}} \leftarrow \pi_d$
\ENDFOR
\end{algorithmic}
\end{algorithm}


\chapter{Experiments}
\label{cha:experiments}

\section{Training and Hyperparameter Tuning}
This section describes the training setup used throughout our experiments, introduces the hyperparameter tuning process, and discusses the training results for all three of our controllers. 

\subsection{Training Setup}
The training of our three agents, i.e. the nominal, attacker and secure controllers, was made in their distinct environments described in \ref{section:mdp-nominal}, \ref{section:mdp-attack} and \ref{section:mdp-defend}, respectively. The training process is performed in accelerated environments capped to the GPU’s capacity and is integrated into a Linux Centos virtual machine\footnote{Machine provided by the Computational Shared Facility (\nolinkurl{https://ri.itservices.manchester.ac.uk/csf3/}) of the University of Manchester.}. The machine provides a dedication of 64GB RAM, two 2.60GHz 12-core processors (Intel Xeon E5-2690 v3) and a Tesla V100 GPU. 

Our overall training framework involves five parallel environments, each assigned to an actor, for a total of $N=5$ parallel actors. In each episode, a hovering point is generated within the range $x_h \in [-1, 1], y_h \in [-1, 1], z_h \in [0, 2]$ with a uniform probability distribution, where one unit of distance in the environment is proportional to one meter in the real world. That allows the agents to learn by exploring a vast three-dimensional space and to not overfit to a single trajectory or hovering point, as described in \textbf{Remark 2}.
Furthermore, at each episode, the quadrotor is spawned at a random position taken from the same ranges as above so that its optimal path relative to the hovering point covers a greater variance. Finally, note that all three agents are trained in ideal conditions. That is, the system uses ground-truth measurements, and the reward functions are assumed to be optimal.\\

\textbf{Remark 4.} The reason why we did not do selection over a larger hyperparameter space is because of the high cost implied in the training such models.

\subsection{Hyperparameter Tuning}
To optimize the training and performance of our agents, the following hyperparameters are tuned: The policy architecture $F_\pi$, learning rate $\alpha$, batch size $M \leq NT$, number of epochs $K$, discount factor $\gamma$, and number of timesteps per iteration $T$. A grid of parameter values was created from background experience and inspiration from the successful literature \cite{jiang2021quadrotor, rubi2021deep}. The values considered for the nominal controller are the following: $T \in \{5120, 10240\}$, $F_\pi \in \{(64, 64), (128, 64), (128, 128)\}$, $\alpha \in \{1\times10^{-4}, 3\times10^{-4}\}$, $M \in \{128, 256\}$, $K=10$, and $\gamma \in \{0.85, 0.99\}$. The policy architecture of the attacker and secure controllers are reduced to $F_\pi \in \{(64, 64), (128, 64)\}$ based on the assumption that their task has improved linearity over autonomous control. As a result, we obtain 48 and 32 hyperparameter combinations, respectively. The number of iterations was decided through Early Stopping based on the average reward obtained by evaluating the policy over twenty episodes every 45 iterations. This means that training was terminated when the agent would not improve anymore or when convergence to a local maximum occurred (e.g., if the quadrotor is unable to stabilize when dealing with the nominal controller or unable to recover when dealing with the defender). \\

\textbf{On weighting matrices.} The weighting matrices $\mathcal{Q}$ and $R$ in the agents' reward functions are defined as follows. Error to reference weights $\mathcal{Q}_b, \mathcal{Q}_a, \mathcal{Q}_d = [1, 1, 1]$, meaning that error to reference in all three axes are equivalently important. Cost weights in $R_b, R_a, R_d$ are defined according to the quadrotor's energy scheme. Stability weights are $\mathcal{L}_b = [1, 1, 0]$, meaning that pitch and roll angles are equivalently important for stability while the yaw angle does not matter.

\subsubsection{Training of the nominal controller}
\begin{figure}[h!]
    \centering
    \includegraphics[width=0.77\linewidth]{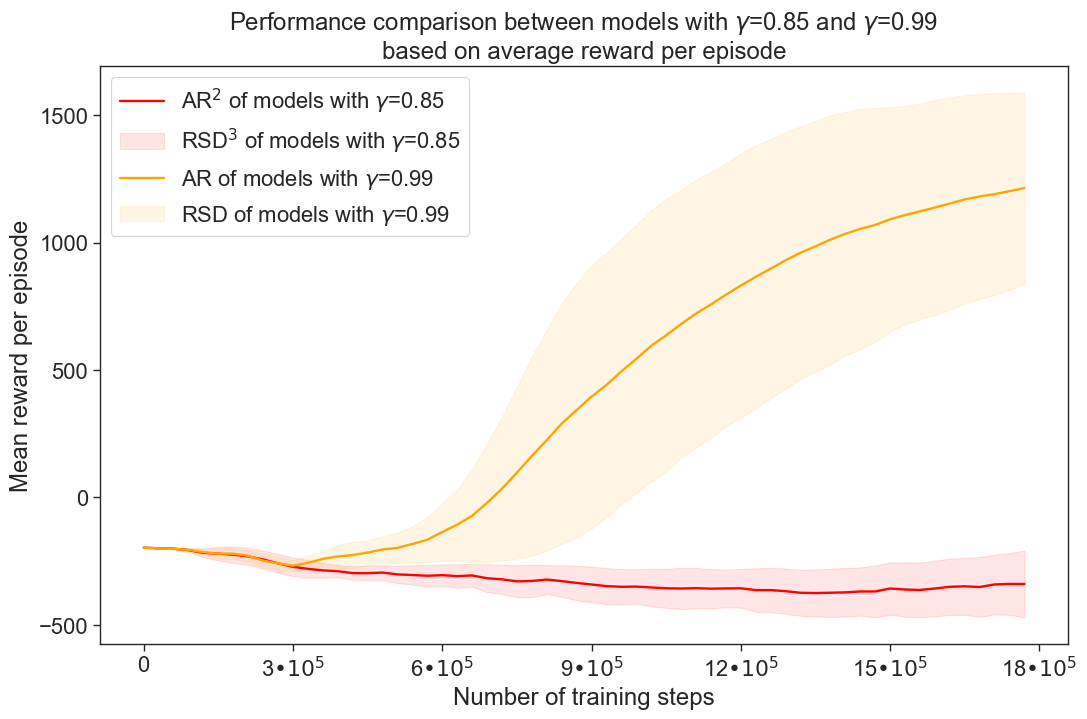}
    \caption[Comparison of different discount factor values on the nominal controller.]{Models are grouped based on their discount factor value, i.e. models with $\gamma=0.85$ and $\gamma=0.99$. Then, they are evaluated at every training iteration, where each group's AR\footnotemark[2] (Average Reward) and RSD\footnotemark[3] (Reward's Standard Deviation) are recorded.}
    \label{fig:nominal-gamma}
\end{figure}

As we can observe from Figure \ref{fig:nominal-gamma}, models with $\gamma=0.99$ are, on average, much more performant than those with $\gamma=0.85$. In fact, while the latter group receives constant rewards between -200 and -400 throughout the training, the former showcases sharp increases from -200 to over 1200 rewards on average at termination. 

These results underscore the importance of the discount factor $\gamma$ in the training of our controller and suggest that actions taken by the nominal controller throughout an episode have a long-term effect on its performance. A higher $\gamma$ value, close to 1, accounts that future rewards substantially impact the agent’s current decisions, promoting a more farsighted approach to maximizing rewards. Therefore, choosing $\gamma=0.99$ gives the controller an enhanced ability to integrate the long-term consequences of its actions into its decision-making process. Such capability is of even greater importance in our environment as it is considerably sensitive and issues non-linear dynamics. \\

\footnotetext[2]{AR means 'Average Reward'.}
\footnotetext[3]{RSD means 'Reward's Standard Deviation'.}

\begin{figure}[ht]
    \centering
    \begin{subfigure}[b]{0.49\textwidth}
        \includegraphics[width=\textwidth]{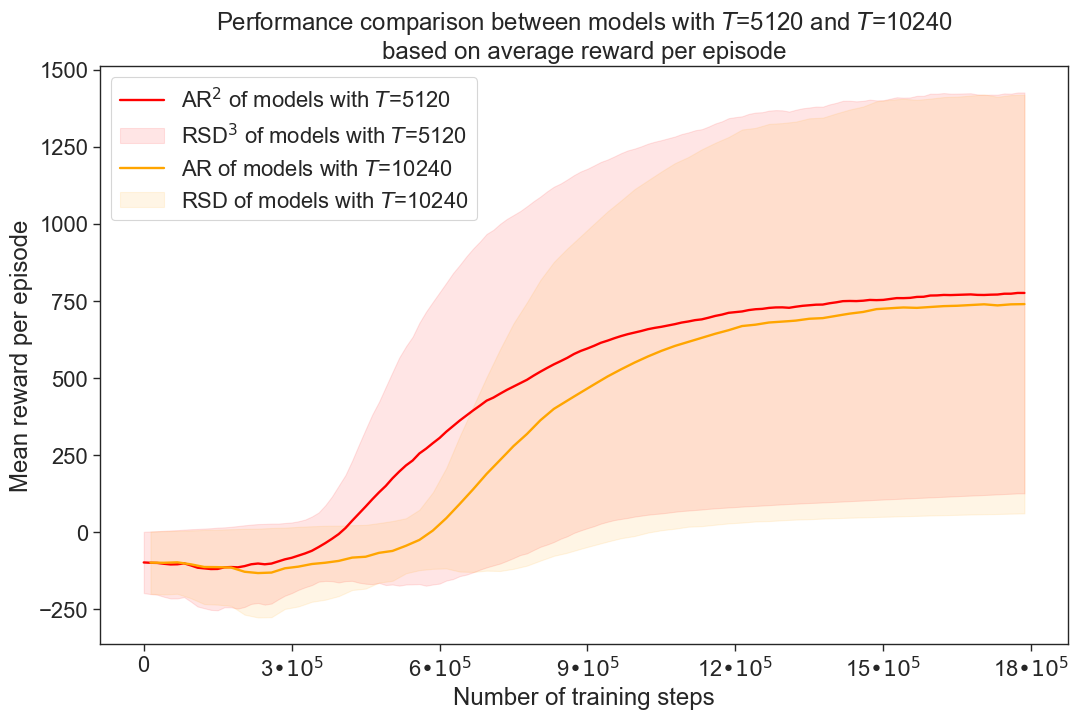}
        \caption{}
        \label{fig:nominal-T}
    \end{subfigure}
    \hfill
    \begin{subfigure}[b]{0.49\textwidth}
        \includegraphics[width=\textwidth]{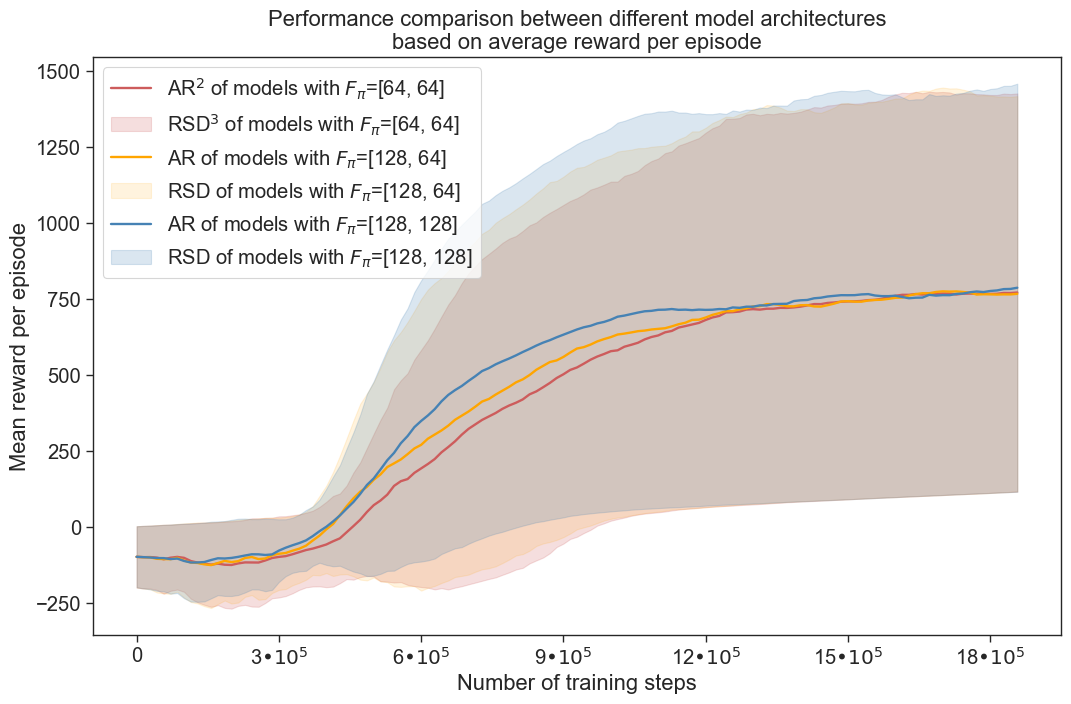}
        \caption{}
        \label{fig:nominal-arch}
    \end{subfigure}

    \begin{subfigure}[b]{0.49\textwidth}
        \includegraphics[width=\textwidth]{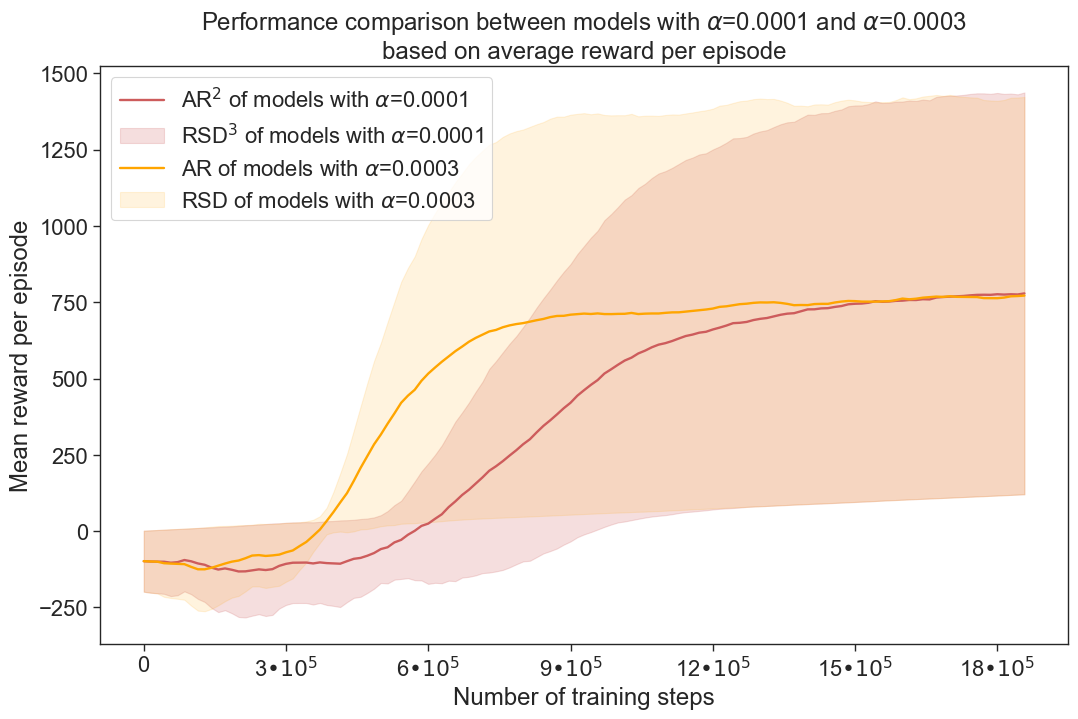}
        \caption{}
        \label{fig:nominal-lr}
    \end{subfigure}
    \hfill
    \begin{subfigure}[b]{0.49\textwidth}
        \includegraphics[width=\textwidth]{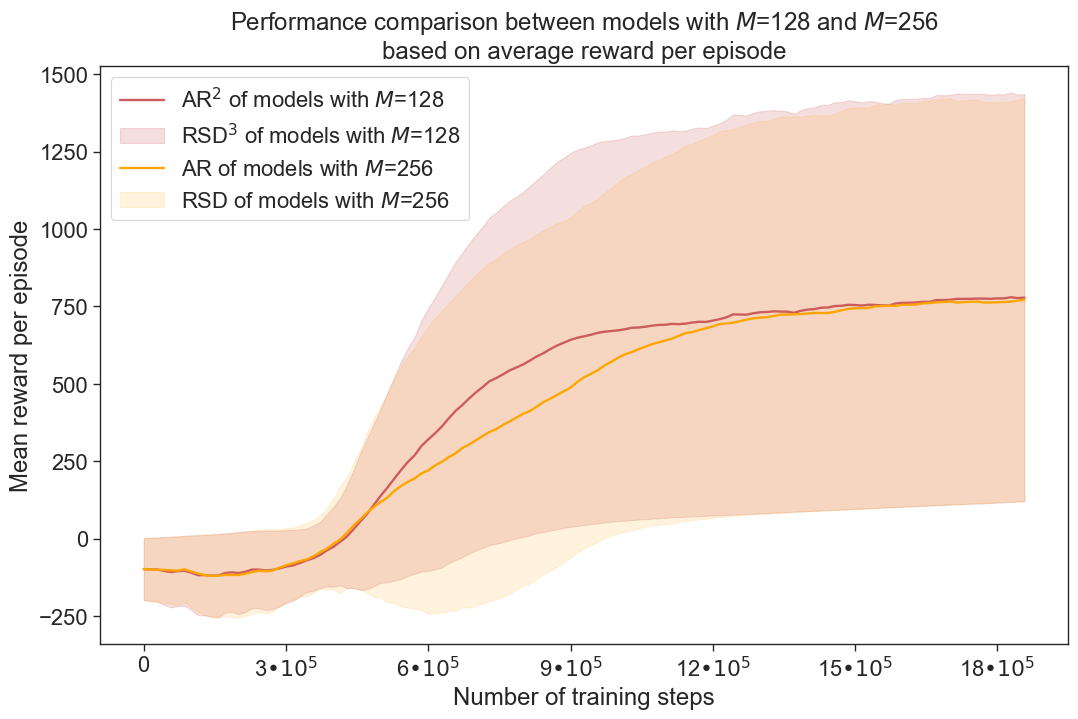}
        \caption{}
        \label{fig:nominal-bs}
    \end{subfigure}

    \caption[Comparison of various hyperparameter values on the nominal controller.]{Following the same process as in Figure \ref{fig:nominal-gamma}, we display the performance comparison between groups of nominal models with \textbf{(a)} steps per iterations $T=5120$ and $T=10240$, \textbf{(b)} model architectures $F_\pi = [64, 64]$, $F_\pi = [128, 64]$ and $F_\pi = [128, 128]$, \textbf{(c)} learning rates $\alpha=1\times10^{-4}$ and $\alpha=3\times10^{-4}$, and \textbf{(d)} batch sizes $M=128$ and $M=256$.}
    \label{fig:nominal-params}
\end{figure}

From the results cited above, we decided to discard the hyperparameter $\gamma=0.85$ and select $\gamma=0.99$ henceforth. Then, Figure \ref{fig:nominal-params} compares the effect of the other hyperparameters $T, F_\pi, \alpha$ and $M$. From Subfigure \ref{fig:nominal-T}, we observe that, although the influence of $T$ is lighter than the one of $\gamma$ in the training of our model, $T=5120$ still outperforms $T=10240$ by converging about 20\% faster and being constantly above in terms of average reward. 

The same holds for the learning rate $\alpha$ in Subfigure \ref{fig:nominal-lr}, where $\alpha=3\times10^{-4}$ converges much faster than $\alpha=1\times10^{-4}$. By increasing the learning rate, we define the step size, i.e. how much we update our model based on its loss at each iteration. The risk with a high learning rate is the exploding gradient effect \cite{philipp2017exploding}, where the learning rate is too large, causing the parameters to overshoot the minimum of the loss function and potentially diverge to infinity. However, $\alpha=3\times10^{-4}$ seems to preserve a proper balance between fast convergence and stability in learning. It should also be noted that, as described in section \ref{section:ppo}, the proximal policy optimization algorithm enhances stability in the learning by constraining the policy updates. Nonetheless, by choosing $\alpha=3\times10^{-4}$, we can reduce our training time and energy to about $9\times10^5$ training steps, compared to $16\times10^5$ with $\alpha=3\times10^{-4}$.

Regarding model architectures, Subfigure \ref{fig:nominal-arch} shows that the proposed three architectures yield similar results, with their convergence speed ordered with respect to their size. However, bigger models are more expensive to train as they require updating more weights at each iteration. Therefore, although we prioritize model performance, in case two models yield very close or equal results, we prioritize the one with lower training time and energy requirement.

\begin{table}[h!]
    \centering
    \begin{tabular}{|c|ccc|} \hline
        Policy architecture ($F_\pi$) & $[64, 64]$ & $[128, 64]$ & $[128, 128]$\\ \hline
        \makecell{Convergence speed to\\training cost ratio} & 3.3:1.00 (3.3) & 4.6:1.07 (\textbf{4.3}) & 5.5:1.12 (\textbf{4.9})\\ \hline
    \end{tabular}
    \caption[Convergence speed to training cost ratio of various policy architectures]{Convergence speed to training cost ratio of different policy architectures ($F_\pi$). Convergence speed is decided based on the total number of timesteps executed and training cost based on the total GPU usage for training.}
    \label{tab:nominal-arch-ratio}
\end{table}

Given the convergence speed to training cost ratios calculated in Table \ref{tab:nominal-arch-ratio}, we can discard the architecture $F_\pi = [64, 64]$. Although that one requires less computation, its ratio to convergence speed is under-performing compared to alternatives. On the other hand, the other two architectures can be further considered as they possess significantly greater ratios.

Finally, we decided not to make any direct conclusion on the effect of the batch sizes ($M$) as their results are instead very close and such a parameter does not seriously affect the total GPU usage.

\clearpage

\begin{figure}[ht!]
    \centering
    \includegraphics[width=0.8\linewidth]{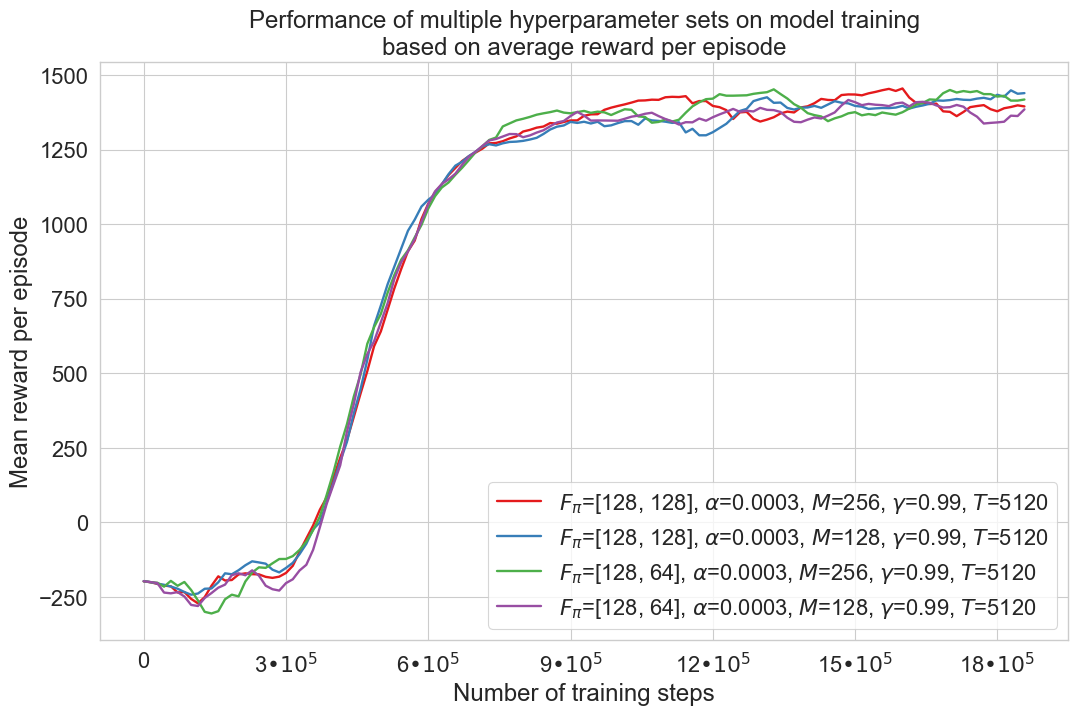}
    \caption[Training performance of the remaining nominal controllers.]{Training performance of the remaining four hyperparameter sets based on average reward per episode, for the nominal controller.}
    \label{fig:nominal-final}
\end{figure}

The training performance of the remaining four sets of hyperparameters are displayed in Figure \ref{fig:nominal-final}. We can observe that all of them yield very similar results. Therefore, given the discussion above and results from Figures \ref{fig:nominal-gamma}, \ref{fig:nominal-params} and \ref{fig:nominal-final}, the final hyperparameters chosen for our nominal controller are $T=5120$, $F_\pi=[128, 64]$, $\alpha=0.0003$, $M=256$ and $\gamma=0.99$. The reason why we decided this set over the remaining four is because, as stated previously, we prioritized the ones with cheaper training, i.e. smaller $F_\pi$. Additionally, $M=256$ had the highest reward pick and, thus, was chosen as part of our final hyperparameter set.

\clearpage

\subsubsection{Training of the attack controller}
To select the hyperparameters of our attack controller, we further constrain the space of parameters we search through based on the assumption that optimal false data injection attacks have greater linearity than autonomous control. Therefore, because a bigger policy network only benefits from the greater non-linearity it can provide, $F_\pi=[128, 128]$ may not be an appropriate option to consider. Furthermore, as the results of the nominal controller suggest, $\gamma=0.85$ does not fit our environment due to its sensitivity and complex nature. As a result, the following parameter spaces are adjusted: $F_\pi=\{(64, 64), (128, 64))\}$ and $\gamma=0.99$.

\begin{figure}[ht]
    \centering
    \begin{subfigure}[b]{0.49\textwidth}
        \includegraphics[width=\textwidth]{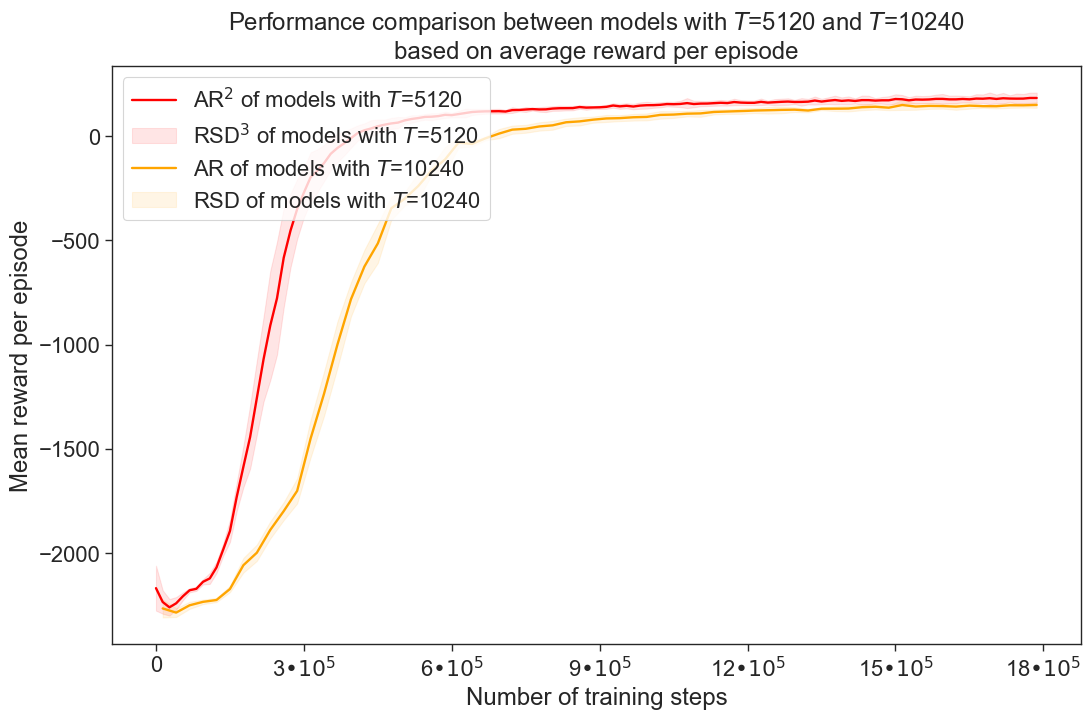}
        \caption{}
        \label{fig:attacker-T}
    \end{subfigure}
    \hfill
    \begin{subfigure}[b]{0.49\textwidth}
        \includegraphics[width=\textwidth]{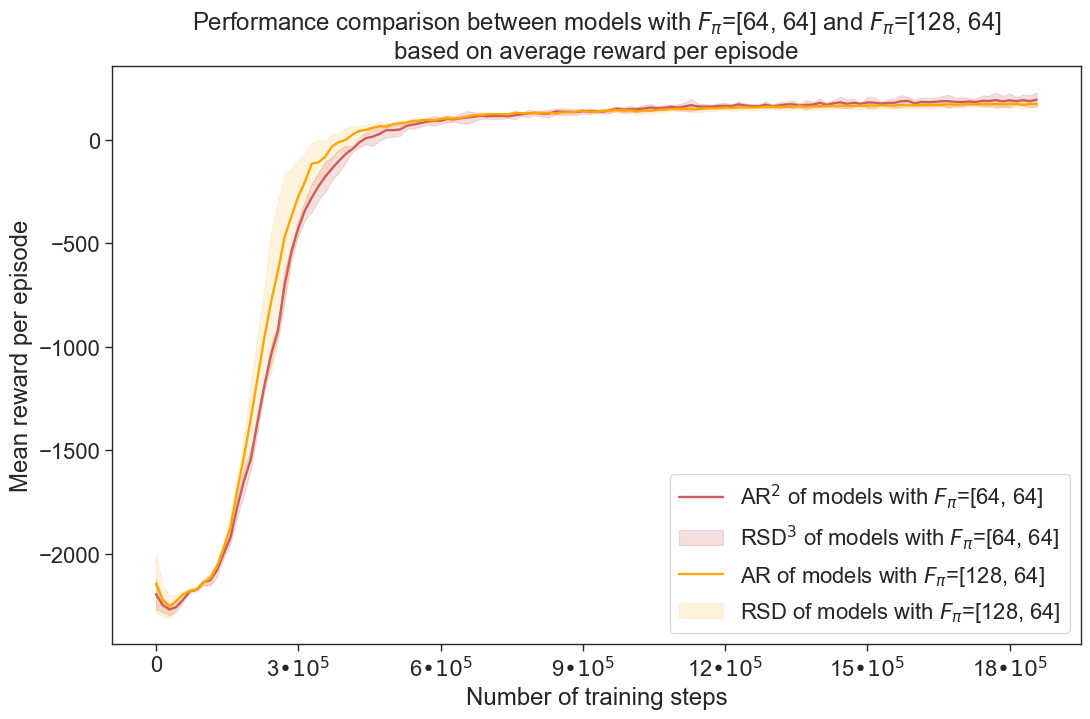}
        \caption{}
        \label{fig:attacker-arch}
    \end{subfigure}

    \begin{subfigure}[b]{0.49\textwidth}
        \includegraphics[width=\textwidth]{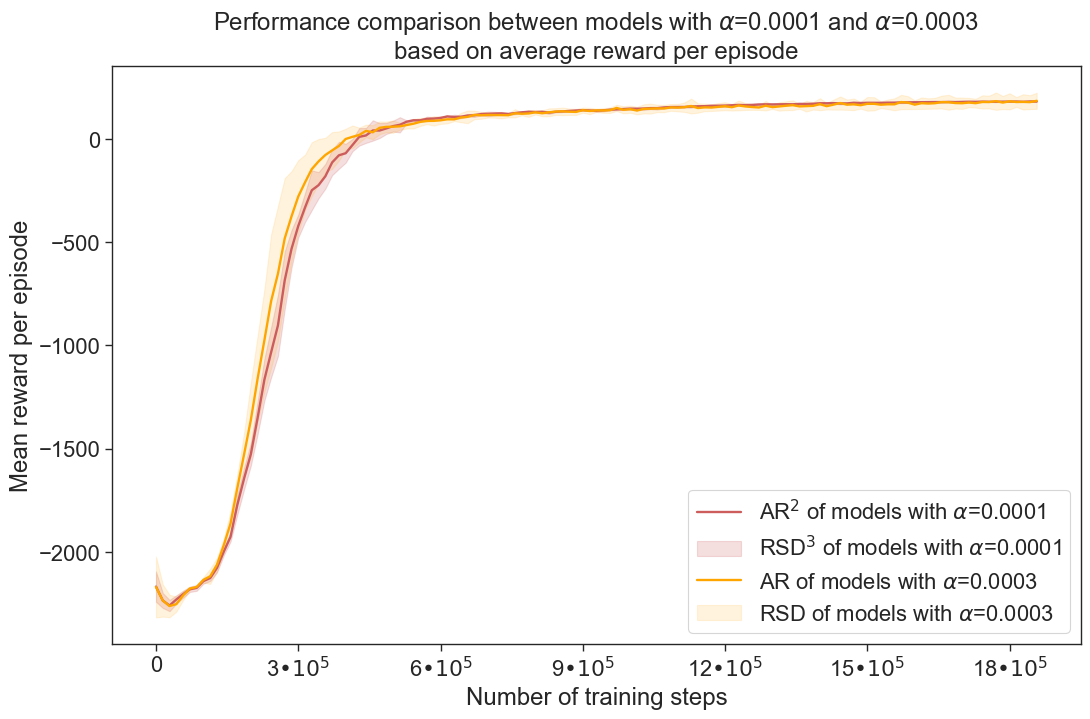}
        \caption{}
        \label{fig:attacker-lr}
    \end{subfigure}
    \hfill
    \begin{subfigure}[b]{0.49\textwidth}
        \includegraphics[width=\textwidth]{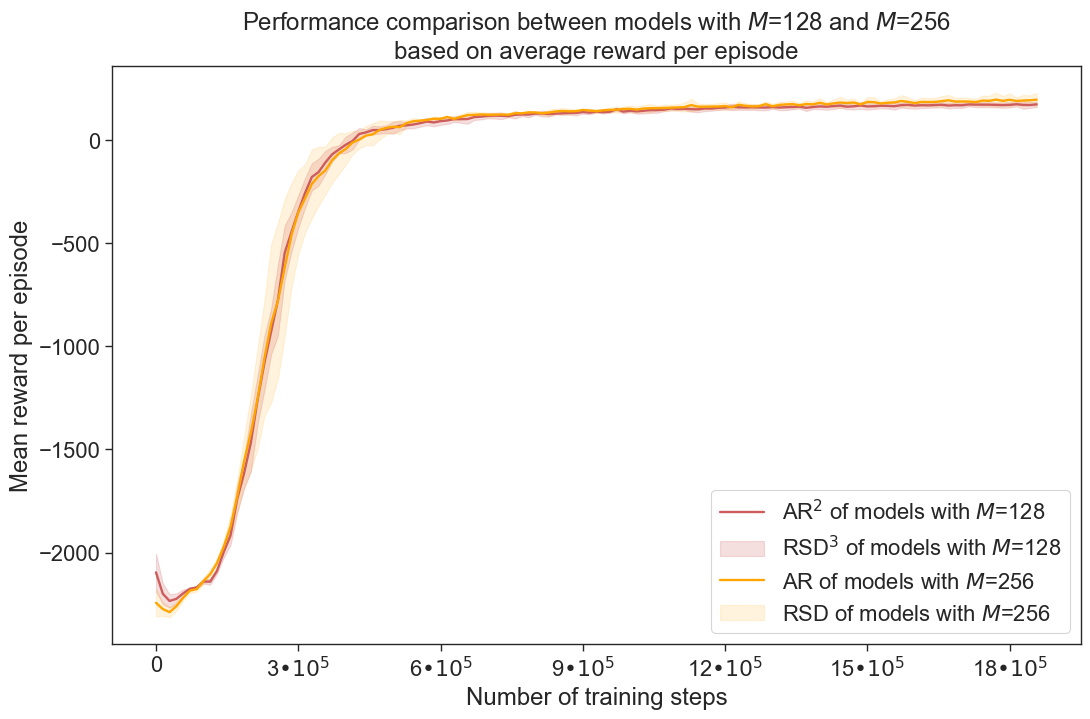}
        \caption{}
        \label{fig:attacker-bs}
    \end{subfigure}

    \caption[Comparison of various hyperparameter values on the attack controller.]{Following the same process as in Figure \ref{fig:nominal-gamma}, we display the performance comparison between groups of attacker models with \textbf{(a)} steps per iterations $T=5120$ and $T=10240$, \textbf{(b)} model architectures $F_\pi = [64, 64]$ and $F_\pi = [128, 64]$, \textbf{(c)} learning rates $\alpha=1\times10^{-4}$ and $\alpha=3\times10^{-4}$, and \textbf{(d)} batch sizes $M=128$ and $M=256$.}
    \label{fig:attacker-params}
\end{figure}

Figure \ref{fig:attacker-params} compares the effect of different hyperparameters on the training of the attack controller, in a similar fashion to what has been done previously on the nominal. However, the result differs significantly from what we obtained previously. 
First, we observe much lower standard deviations between models of the same groups. That is mainly because the attacker controller is less sensitive to the choice of hyperparameters. In addition, as suggested above, the linearity of this task has increased compared to the control of autonomous quadrotors. This observation is logical, considering that controlling a quadrotor in three-dimensional space demands significantly higher precision and offers smaller margins for error compared to one designed to crash the quadrotor.

\textbf{Remark 5.} The reader may ask why rewards are capped to low positive values. That is due to the change in the reward function used to train the malicious attacker. \\

However, it is still possible to make conclusions on the results obtained in Figure \ref{fig:attacker-params}. Subfigure \ref{fig:attacker-T} clearly demonstrates better performance when the number of timesteps per iteration $T=5120$ instead of $T=10240$. A smaller $T$ encourages the model to generalize better across a more diverse set of experiences with still the same number of total timesteps. That is because the agent is exposed to a wider variety of states and outcomes in the same training period. Therefore, as the attacker must learn a vast space of trajectories taken by the nominal controller, fewer timesteps per iteration may be better adapted to an increased exploration. 

Moreover, Subfigure \ref{fig:attacker-arch} verifies our previous assumptions on this task’s complexity. We can observe that the two architectures yield equivalent performance. Although we might notice slightly faster convergence for the bigger architecture, this does not affect the final model as they converge to the same average reward with very little overhead. Consequently, the smaller architecture $F_\pi=[64, 64]$ may provide equal performance with better training costs. 

Finally, although the learning rate $\alpha=3\times10^{-4}$ in Subfigure \ref{fig:attacker-lr} performed slightly better than $\alpha=1\times10^{-4}$, we may find it interesting to look at the result of the remaining four hyperparameter sets, which include variations in $M \in \{128, 256\}$ and $\alpha \in \{\alpha=1\times10^{-4}, \alpha=3\times10^{-4}\}$.

\begin{figure}[h!]
    \centering
    \includegraphics[width=0.8\linewidth]{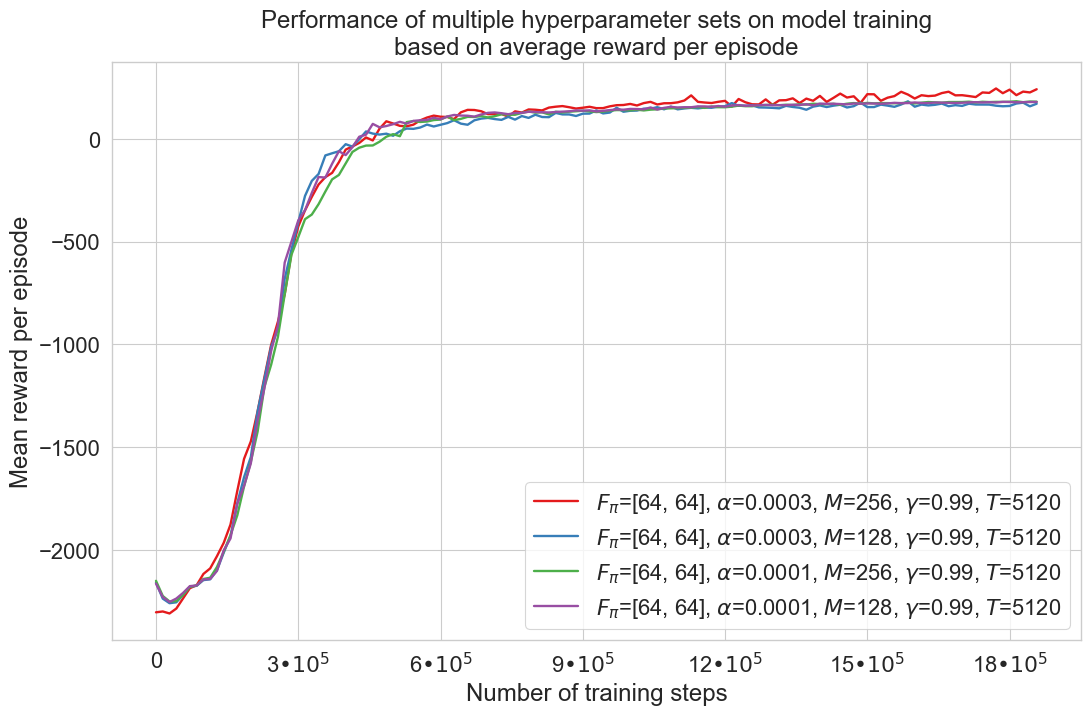}
    \caption[Training performance of the remaining attack controllers.]{Training performance of the remaining four hyperparameter sets based on average reward per episode, for the attack controller.}
    \label{fig:attacker-final}
\end{figure}

Figure \ref{fig:attacker-final} displays the results of the remaining four hyperparameter sets. These are once again very close to each other. However, they denote an interesting dilemma. The model trained on $\alpha=3\times10^{-4}$ and $M=256$ yields better results than the other three models by a slight but notable margin. However, this model is quite unstable and fluctuating compared to the one trained on $\alpha=1\times10^{-4}$ and $M=128$, which instead showcased lower performance. Although the choice here will not drastically affect our results, this phenomenon is quite common in reinforcement learning and often leads to a difficult dilemma. In our study, we decided to go with the stable behaviour, but its alternative could have very well been chosen instead.

As a result, the final hyperparameters selected for our attack controller are the following: $T=5120$, $F_\pi=[64, 64]$, $\alpha=0.0001$, $M=128$ and $\gamma=0.99$.

\subsubsection{Training of the secure controller}
\begin{figure}[ht]
    \centering
    \begin{subfigure}[b]{0.49\textwidth}
        \includegraphics[width=\textwidth]{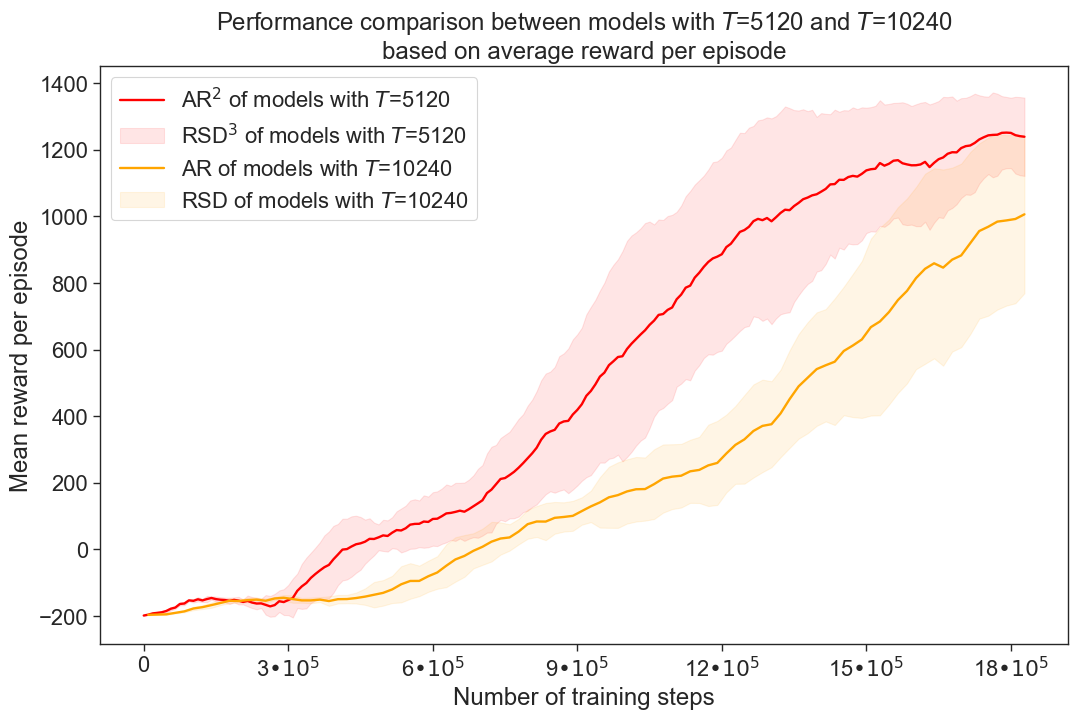}
        \caption{}
        \label{fig:defender-T}
    \end{subfigure}
    \hfill
    \begin{subfigure}[b]{0.49\textwidth}
        \includegraphics[width=\textwidth]{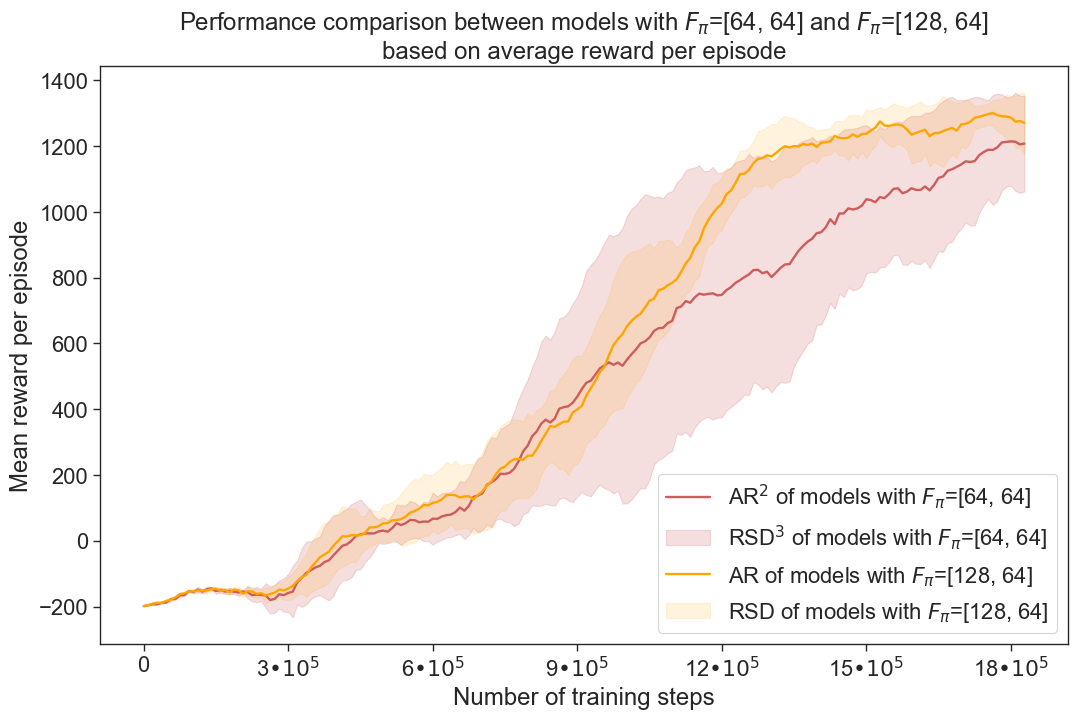}
        \caption{}
        \label{fig:defender-arch}
    \end{subfigure}

    \begin{subfigure}[b]{0.49\textwidth}
        \includegraphics[width=\textwidth]{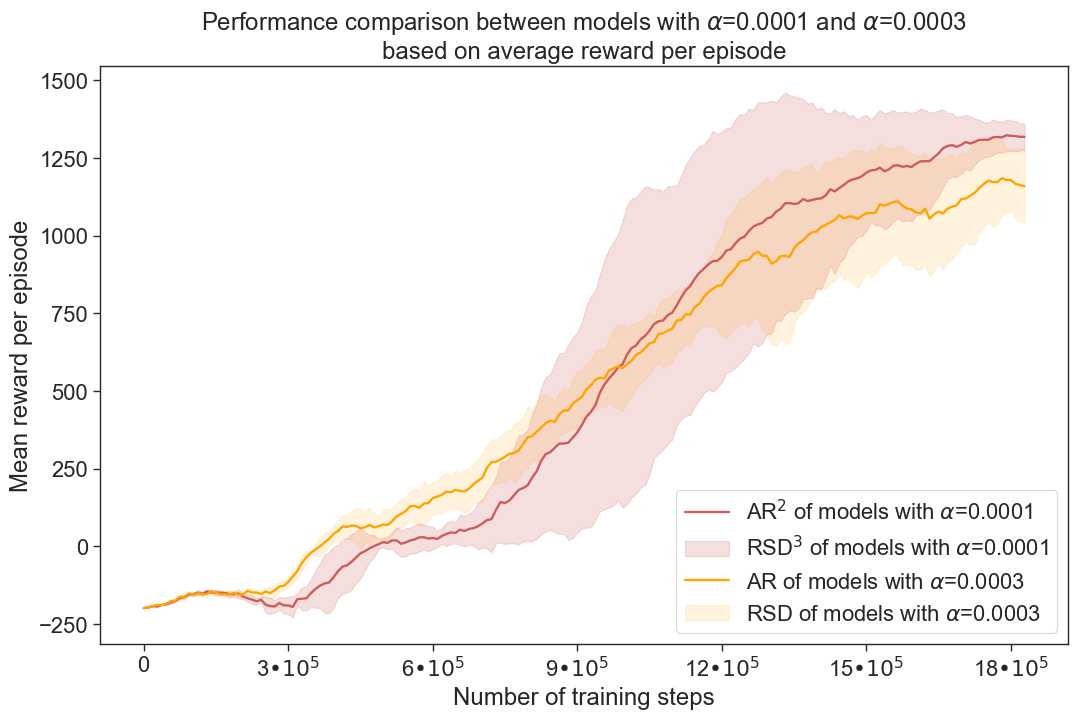}
        \caption{}
        \label{fig:defender-lr}
    \end{subfigure}
    \hfill
    \begin{subfigure}[b]{0.49\textwidth}
        \includegraphics[width=\textwidth]{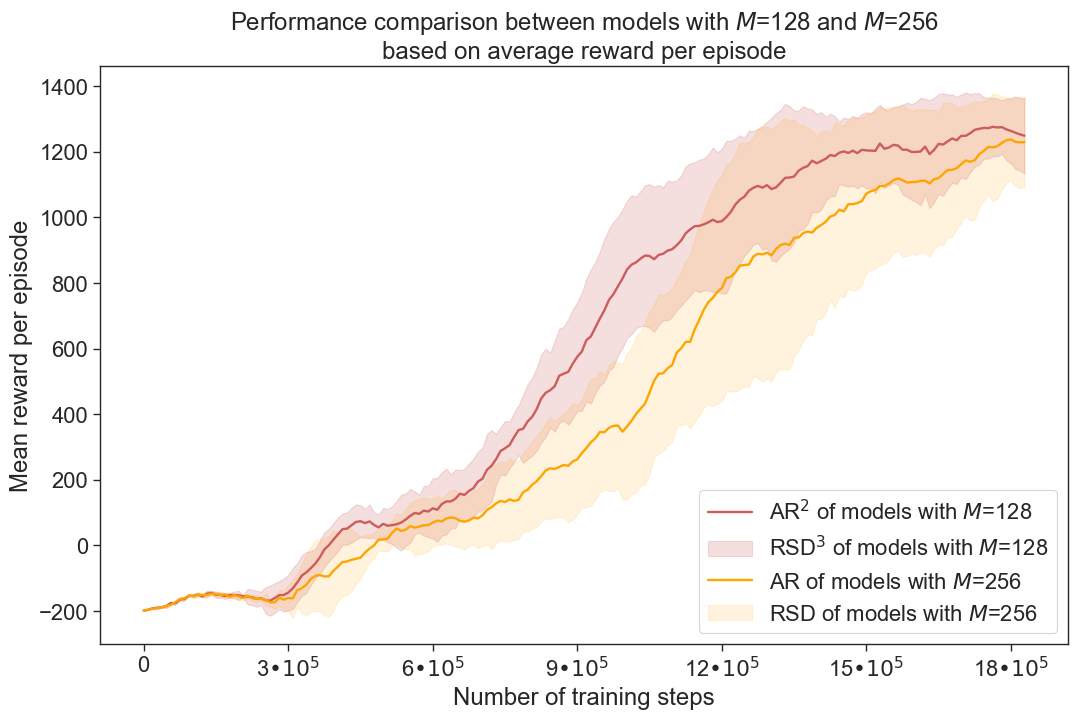}
        \caption{}
        \label{fig:defender-bs}
    \end{subfigure}

    \caption[Comparison of various hyperparameter values on the secure controller.]{Following the same process as in Figure \ref{fig:nominal-gamma}, we display the performance comparison between groups of defender models with \textbf{(a)} steps per iterations $T=5120$ and $T=10240$, \textbf{(b)} model architectures $F_\pi = [64, 64]$ and $F_\pi = [128, 64]$, \textbf{(c)} learning rates $\alpha=1\times10^{-4}$ and $\alpha=3\times10^{-4}$, and \textbf{(d)} batch sizes $M=128$ and $M=256$.}
    \label{fig:defender-params}
\end{figure}

For our secure controller, we search through the same hyperparameter space as the one used by the attack controller. Figure \ref{fig:defender-params} compares the effect between these parameters based on the same model groups.

As we can observe in Subfigures \ref{fig:defender-T} and \ref{fig:defender-lr}, both the number of timesteps per iteration ($T$) and the learning rate ($\alpha$) exhibited a strong influence on model performance. In fact, these discrepancies follow the ones obtained on the attack controller but with amplified effects on the average reward.

\begin{figure}[h!]
    \centering
    \includegraphics[width=0.83\linewidth]{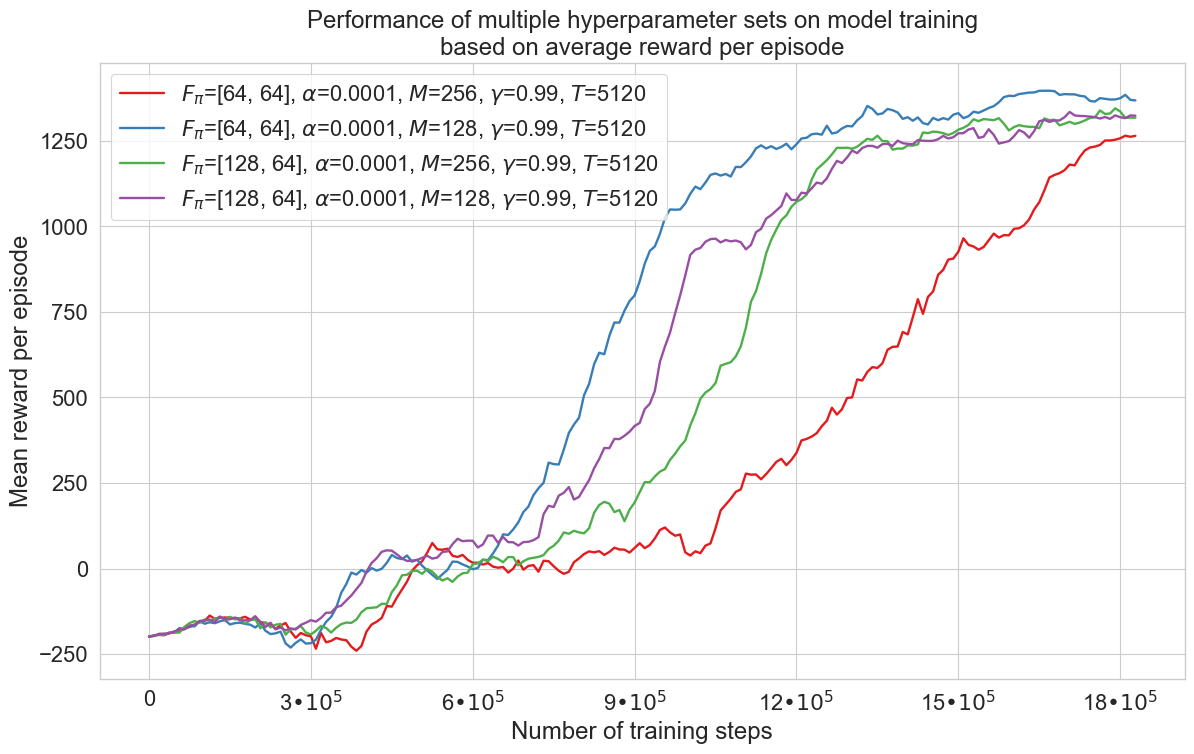}
    \caption[Training performance of the remaining secure controllers.]{Training performance of the remaining four hyperparameter sets based on average reward per episode, for the secure controller.}
    \label{fig:defender-final}
\end{figure}

Moreover, Subfigures \ref{fig:defender-bs} and \ref{fig:defender-arch} show that the batch size ($M$) and model architectures ($F_\pi$) further impact training performance. However, given the results from the four hyperparameter sets on $M$ and $F_\pi$ displayed in Figure \ref{fig:defender-final}, $M=128$ may indeed be better than $M=256$. Surprisingly, however, $F_\pi=[64, 64]$ reveals better individual performance compared to the overall analysis from \ref{fig:defender-arch}. 

In conclusion, the chosen hyperparameters for our secure controller are $T=5120$, $F_\pi=[64, 64]$, $\alpha=0.0001$, $M=128$ and $\gamma=0.99$. In fact, this set is exactly the same as the one selected for the attack controller.

\subsubsection{Conclusion on hyperparameter selection}
The final hyperparameters chosen for our three agents are summarized in Table \ref{tab:final-hyperparams}. \\

We notice that the exact same hyperparameters came up on top for the attack and secure controllers. Furthermore, we note that the common choices between the nominal and the other controllers generally refer to hyperparameters that are affected by the environment, i.e. timesteps per iterations $T$, batch size $M$ and discount factor $\gamma$. On the other hand, the ones varying are solely based on the task and its reward, i.e. the policy architecture $F_\pi$ and learning rate $\alpha$. \\

The entire training process took about 28 hours on the hardware configuration provided above. As an indication, this represents an average of 15 minutes per model.

\begin{table}[ht!]
    \centering
    \begin{tabular}{lccccc} \hline
        Controller & \makecell{Timesteps per\\iteration ($T$)} & \makecell{Architecture\\of policy ($F_\pi$)} & \makecell{Learning\\rate ($\alpha$)} & \makecell{Batch\\size ($M$)} & \makecell{Discount\\factor ($\gamma$)}\\ \hline \hline
        Nominal & 5120 & [128, 64] & $3\times10^{-4}$ & 256 & 0.99\\
        Attack & 5120 & [64, 64] & $1\times10^{-4}$ & 128 & 0.99\\
        Defence & 5120 & [64, 64] & $1\times10^{-4}$ & 128 & 0.99\\ \hline
    \end{tabular}
    \caption{Final hyperparameters selected for each of the three controllers.}
    \label{tab:final-hyperparams}
\end{table}

\section{Evaluation in Simulation} \label{section:evaluation}
In this section, we provide the simulation results on the autonomous control system (i.e. the nominal controller) and validate the proposed optimal false data injection attack scheme and secure countermeasure. In our experiment, the quadrotor is initialised at coordinates $(0.0, 0.0, 0.5)$ and tasked with reaching six different hover points. The six hover points are set up such that the experiment covers a wide trajectory space. Throughout this section, we define hover point coordinates as $P_r = (x_r, y_r, z_r)$.

\subsection{Tracking Control of a Quadrotor using the Nominal Controller designed in Algorithm \ref{algo:nominal}}
\label{section:eval-sim-nominal}

\begin{figure}[ht]
    \centering
    \begin{subfigure}[b]{0.32\textwidth}
        \includegraphics[width=\textwidth]{img/nominal/eval/sim_0.85_0.9_1.7_.png}
        \caption{$P_r = (0.85, 0.90, 1.7)$}
        \label{fig:nominal_sim-0.85_0.9_1.7}
    \end{subfigure}
    \hfill
    \begin{subfigure}[b]{0.33\textwidth}
        \includegraphics[width=\textwidth]{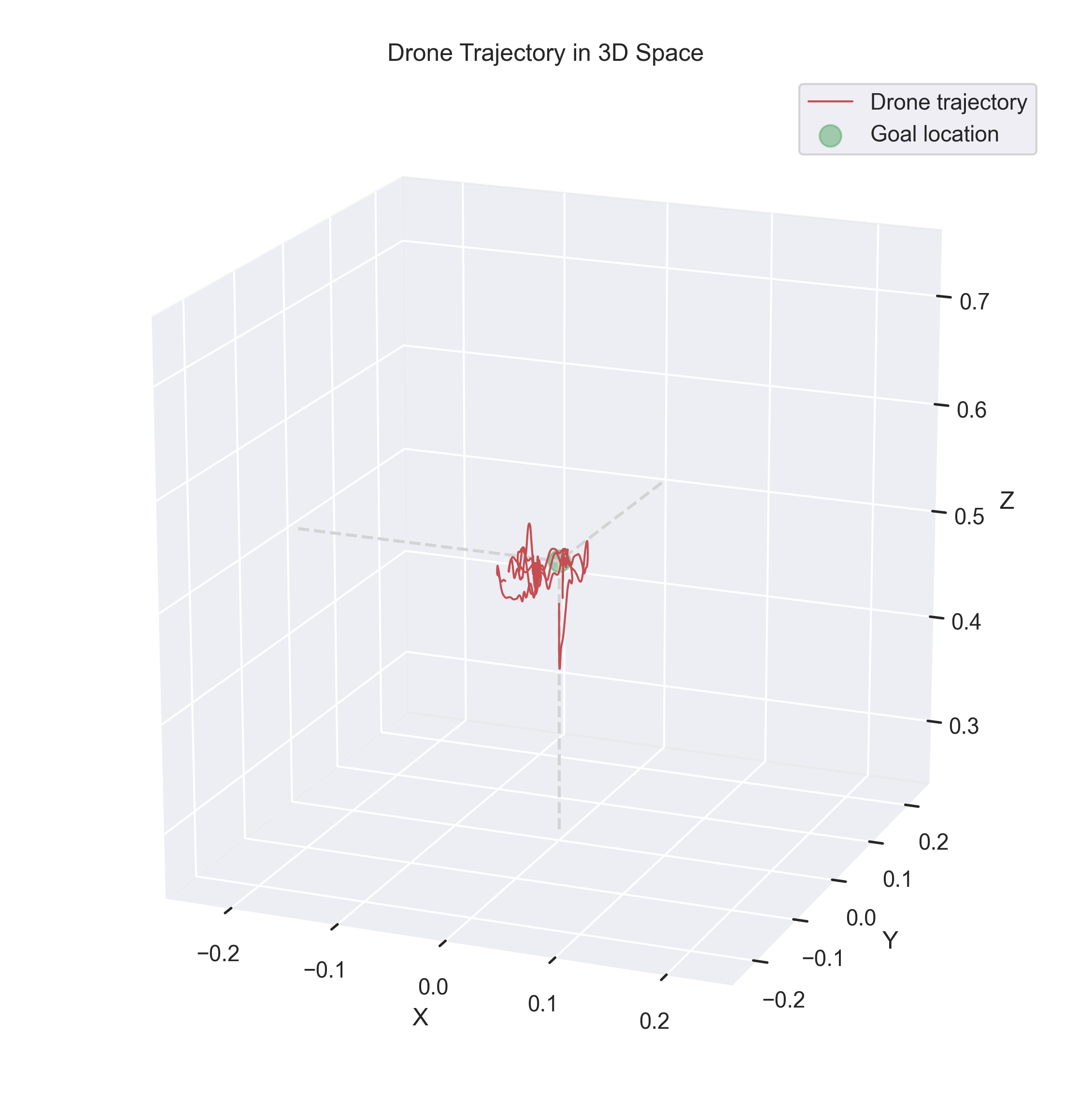}
        \caption{$P_r = (0.0, 0.0, 0.5)$}
        \label{fig:nominal_sim-0.0_0.0_0.5}
    \end{subfigure}
    \hfill
    \begin{subfigure}[b]{0.32\textwidth}
        \includegraphics[width=\textwidth]{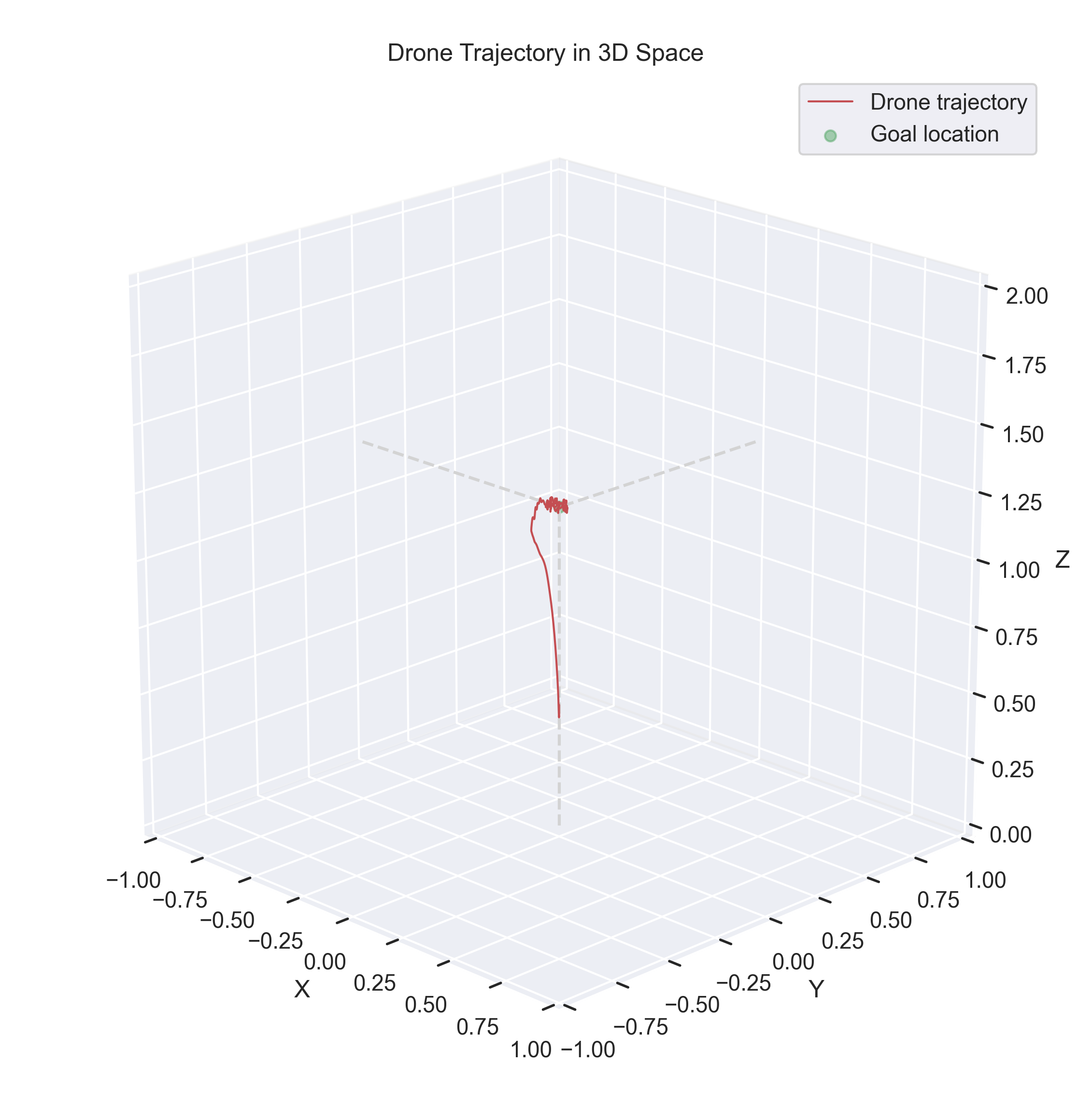}
        \caption{$P_r = (0.0, 0.0, 1.2)$}
        \label{fig:nominal_sim-0.0_0.0_1.2}
    \end{subfigure}

    \begin{subfigure}[b]{0.32\textwidth}
        \includegraphics[width=\textwidth]{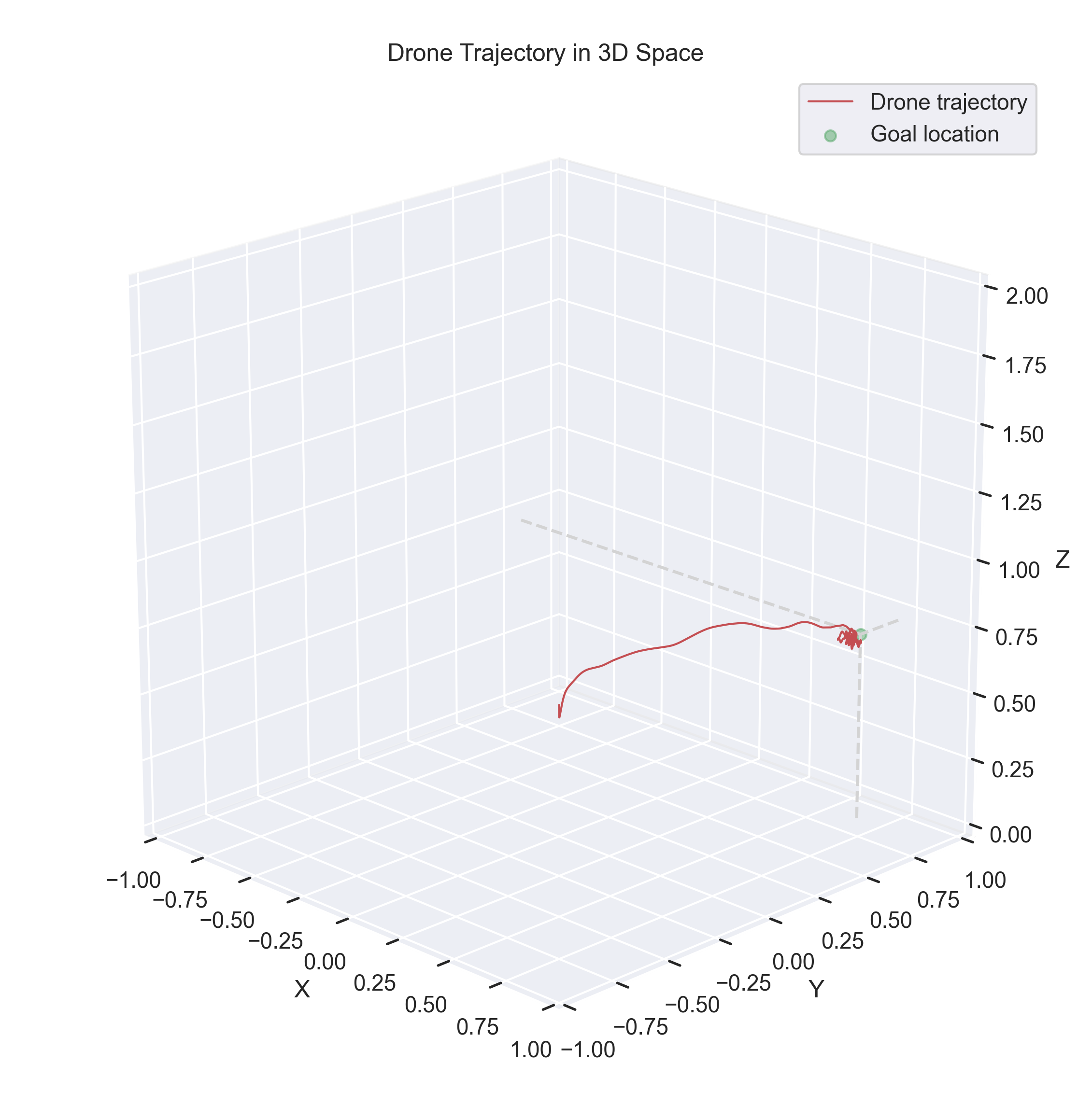}
        \caption{$P_r = (0.7, 0.85, 0.7)$}
        \label{fig:nominal_sim-0.7_0.85_0.7}
    \end{subfigure}
    \hfill
    \begin{subfigure}[b]{0.32\textwidth}
        \includegraphics[width=\textwidth]{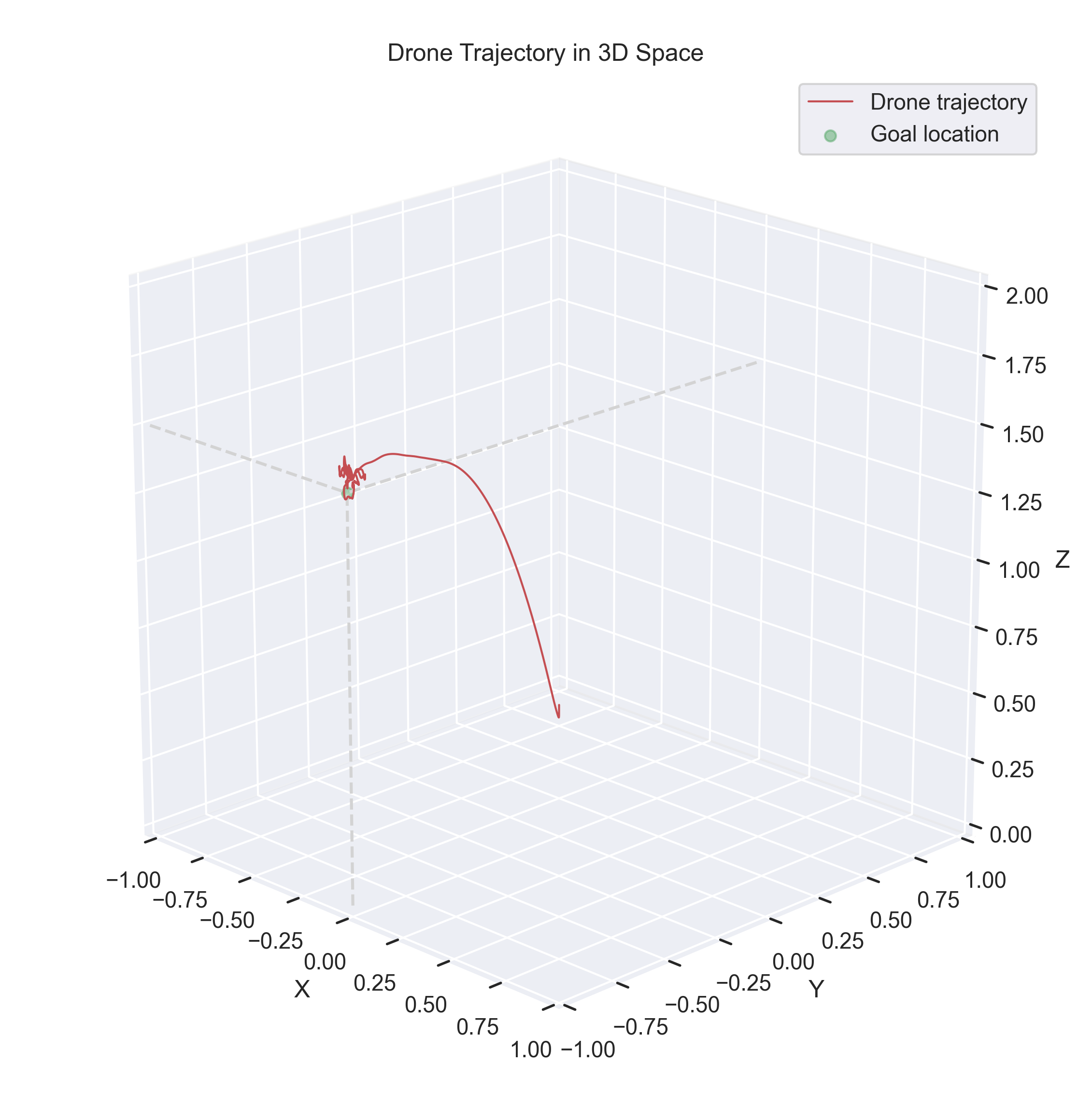}
        \caption{$P_r = (0.0, -1.0, 1.5)$}
        \label{fig:nominal_sim-0.0_-1.0_1.5}
    \end{subfigure}
    \hfill
    \begin{subfigure}[b]{0.32\textwidth}
        \includegraphics[width=\textwidth]{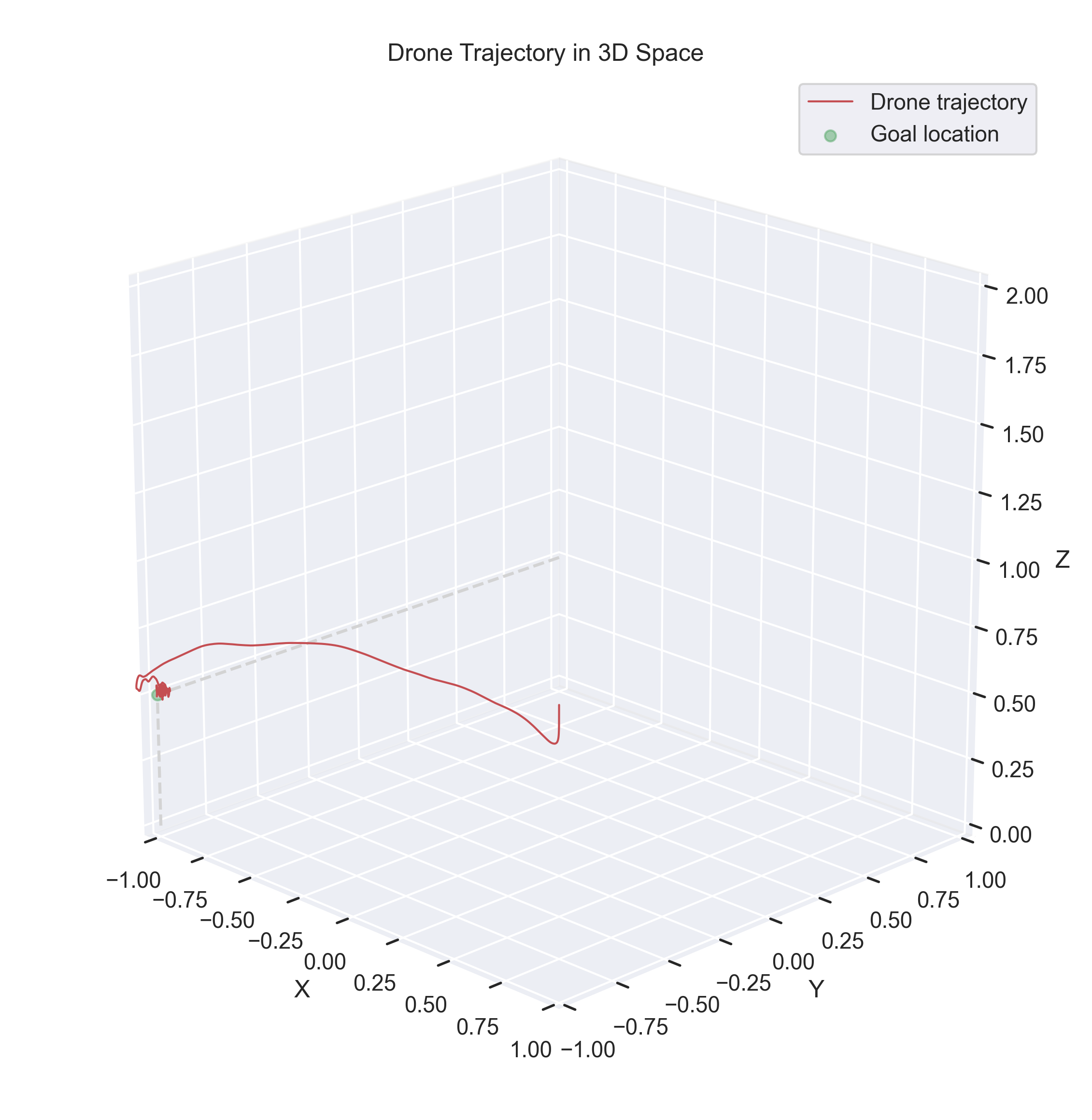}
        \caption{$P_r = (-1.0, -1.0, 0.5)$}
        \label{fig:nominal_sim--1.0_-1.0_0.5}
    \end{subfigure}

    \caption[Trajectories of the quadrotor under the nominal controller.]{Trajectories of the quadrotor under the nominal controller, towards each of the six hovering points. Each experiment has a maximum duration of ten seconds, i.e. the expected duration if the quadrotor does not crash.}
    \label{fig:nominal-sim}
\end{figure}

Figure \ref{fig:nominal-sim} shows the tracking performance of the nominal controller designed in Algorithm \ref{algo:nominal}. As we can observe from these simulations, the learning-based controller can successfully command the quadrotor to reach all six hover points by employing close-to-optimal trajectories. Additionally, the controller could stabilise the quadrotor upon arrival at the hovering location. Although some small fluctuations occur during stabilisation, this is an expected behaviour when using reinforcement-learning-based controllers. In fact, these results are aligned with the state-of-the-art in \cite{jiang2021quadrotor}.

Subfigures \ref{fig:nominal_sim-0.85_0.9_1.7} and \ref{fig:nominal_sim-0.0_-1.0_1.5} show the ability of the nominal controller to fly the quadrotor over \textbf{wide-amplitude} areas and reach \textbf{high-altitude} locations through short optimal paths. Moreover, the trajectories suggest that the controller makes good use of dynamics. In fact, such trajectories resemble some that would be taken by professional drone pilots. Finally, because of the broad trajectories, we could expect difficulties when stabilising upon reaching the target position. However, the controller was able to stabilise the quadrotor without the need for any adaptation time.

Subfigures \ref{fig:nominal_sim-0.7_0.85_0.7} and \ref{fig:nominal_sim--1.0_-1.0_0.5} exhibit the quadrotor's ability to reach \textbf{low-altitude} locations through \textbf{wide-amplitude} trajectories. As with the above examples, the quadrotor could take short optimal paths towards its target location and stabilise upon arrival. It is worth noticing that the nominal controller is doing a great job maintaining a stable altitude throughout its flights. \\

Finally, Subfigures \ref{fig:nominal_sim-0.0_0.0_0.5} and \ref{fig:nominal_sim-0.0_0.0_1.2} aim to further illustrate the quadrotor's capability to \textbf{lift} and \textbf{hover}. Please note that figure \ref{fig:nominal_sim-0.0_0.0_0.5} is zoomed in for the reader's convenience. Subfigure \ref{fig:nominal_sim-0.0_0.0_0.5} tasks the quadrotor with instantly hovering upon the episode's start. This represents a challenging task as the quadrotor needs to instantly catch the dynamics and adapt its thrust to its initial altitude without much time for observation. However, we can observe that it could answer this challenge by immediately stabilising and hovering. Furthermore, Subfigure \ref{fig:nominal_sim-0.0_0.0_1.2} displays the quadrotor trajectory when tasked to lift and hover. While some non-optimal controllers may take an arbitrary path towards this location, our controller was able to lift reasonably straight to quickly reach the hover point and stabilise.\\

\begin{figure}[ht]
    \centering
    \begin{subfigure}[b]{0.32\textwidth}
        \includegraphics[width=\textwidth]{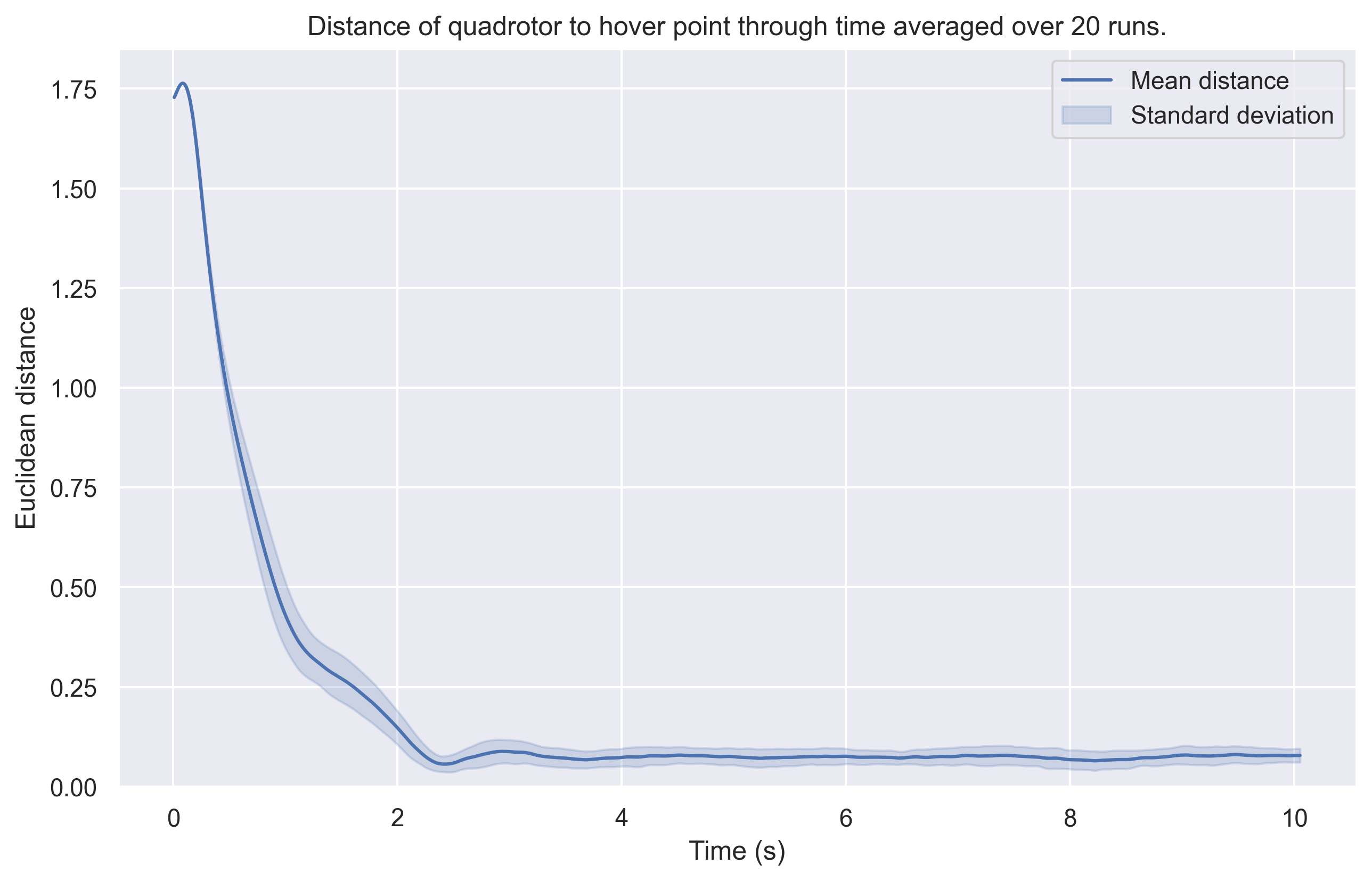}
        \caption{$P_r = (0.85, 0.90, 1.7)$}
        \label{fig:nominal_dist-0.85_0.9_1.7}
    \end{subfigure}
    \hfill
    \begin{subfigure}[b]{0.32\textwidth}
        \includegraphics[width=\textwidth]{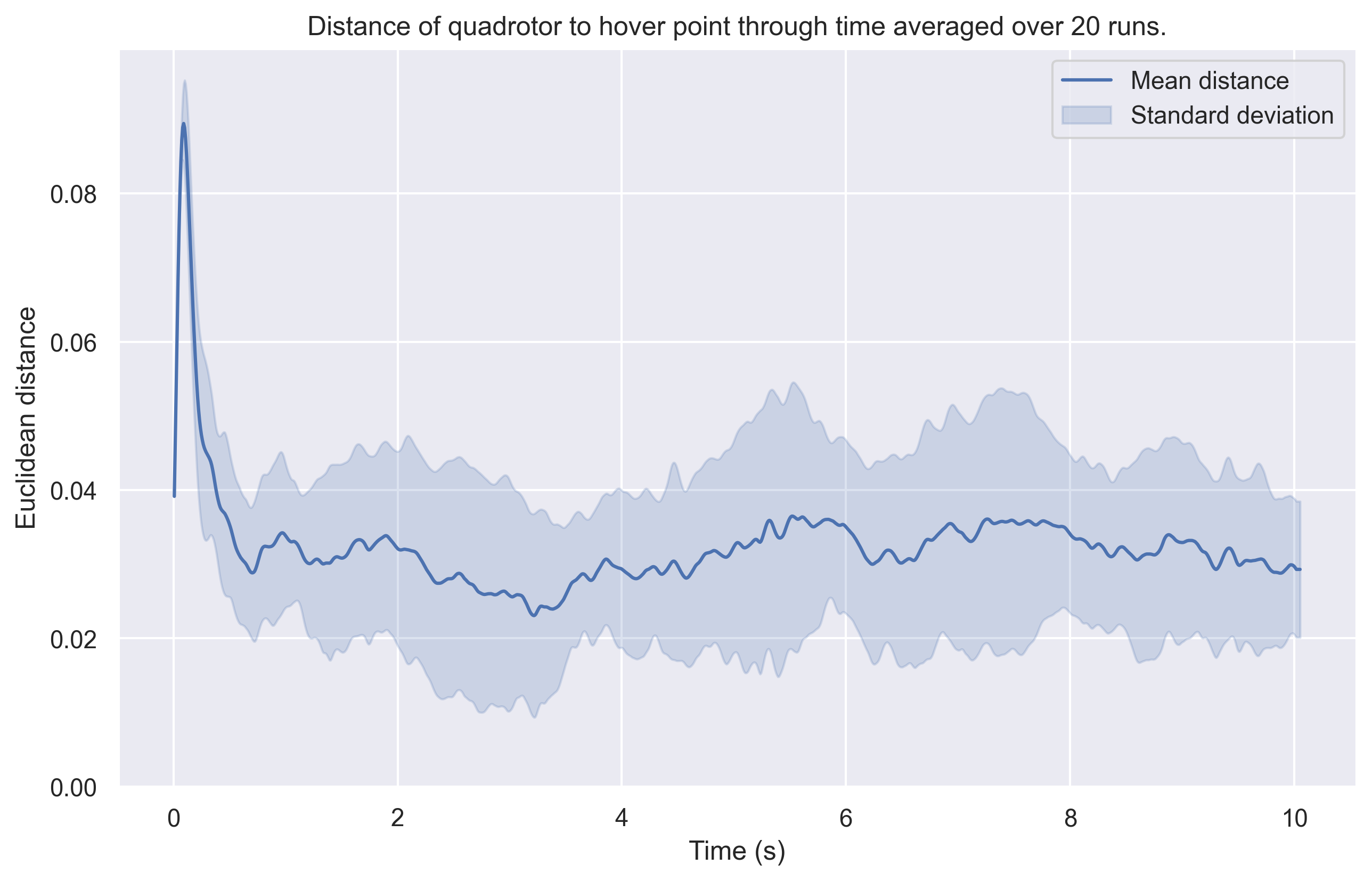}
        \caption{$P_r = (0.0, 0.0, 0.5)$}
        \label{fig:nominal_dist-0.0_0.0_0.5}
    \end{subfigure}
    \hfill
    \begin{subfigure}[b]{0.32\textwidth}
        \includegraphics[width=\textwidth]{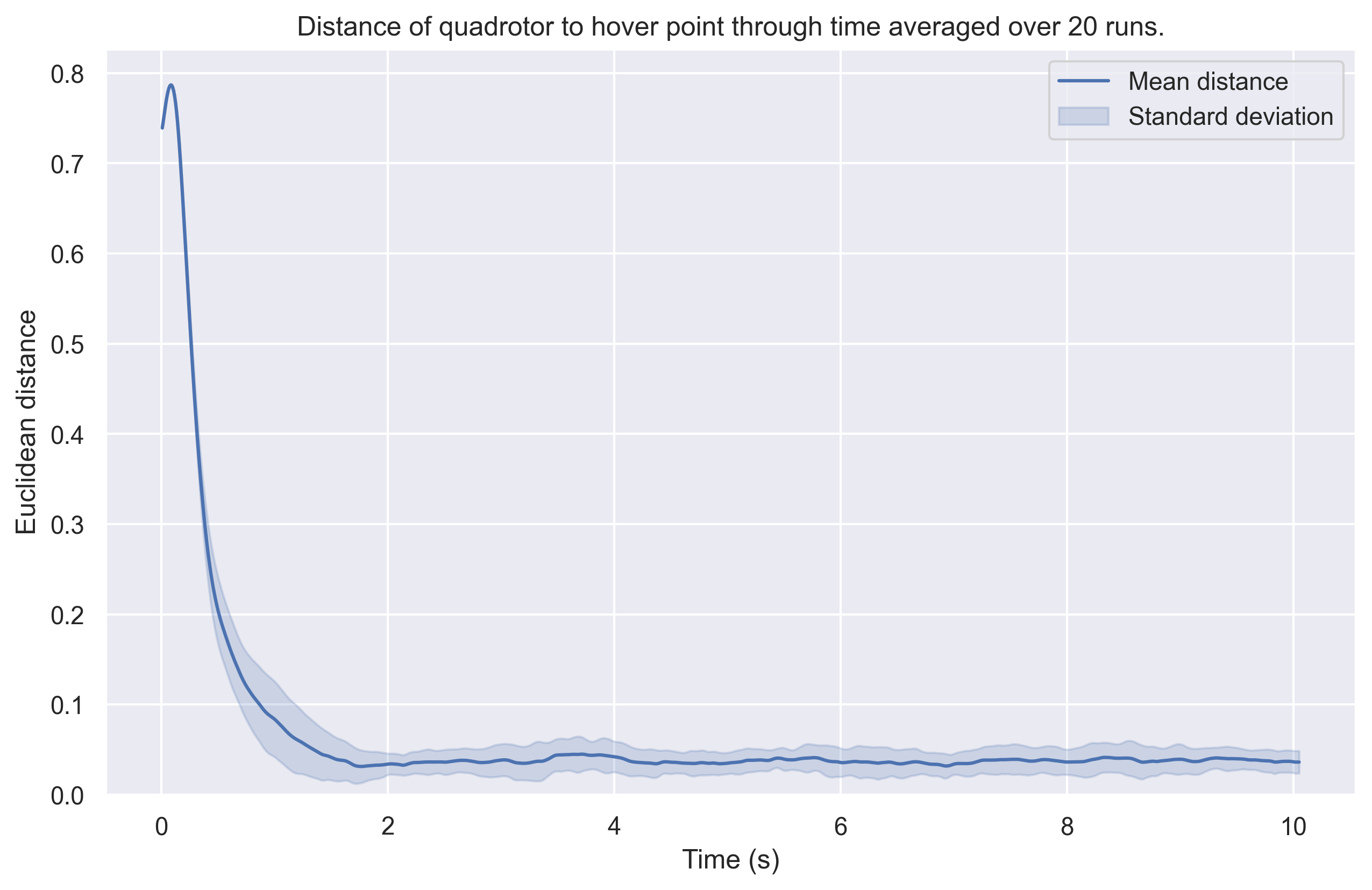}
        \caption{$P_r = (0.0, 0.0, 1.2)$}
        \label{fig:nominal_dist-0.0_0.0_1.2}
    \end{subfigure}

    \begin{subfigure}[b]{0.32\textwidth}
        \includegraphics[width=\textwidth]{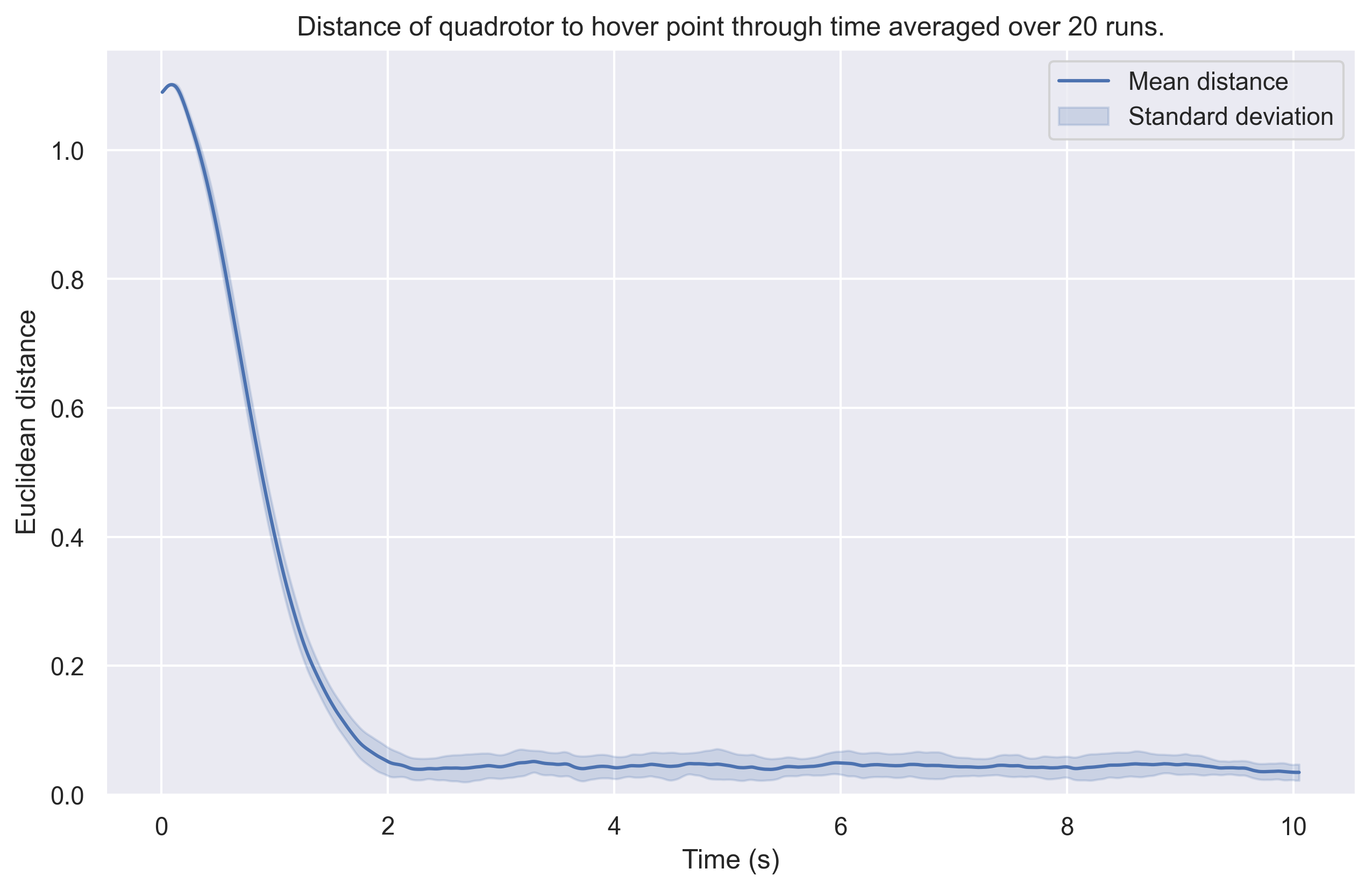}
        \caption{$P_r = (0.7, 0.85, 0.7)$}
        \label{fig:nominal_dist-0.7_0.85_0.7}
    \end{subfigure}
    \hfill
    \begin{subfigure}[b]{0.32\textwidth}
        \includegraphics[width=\textwidth]{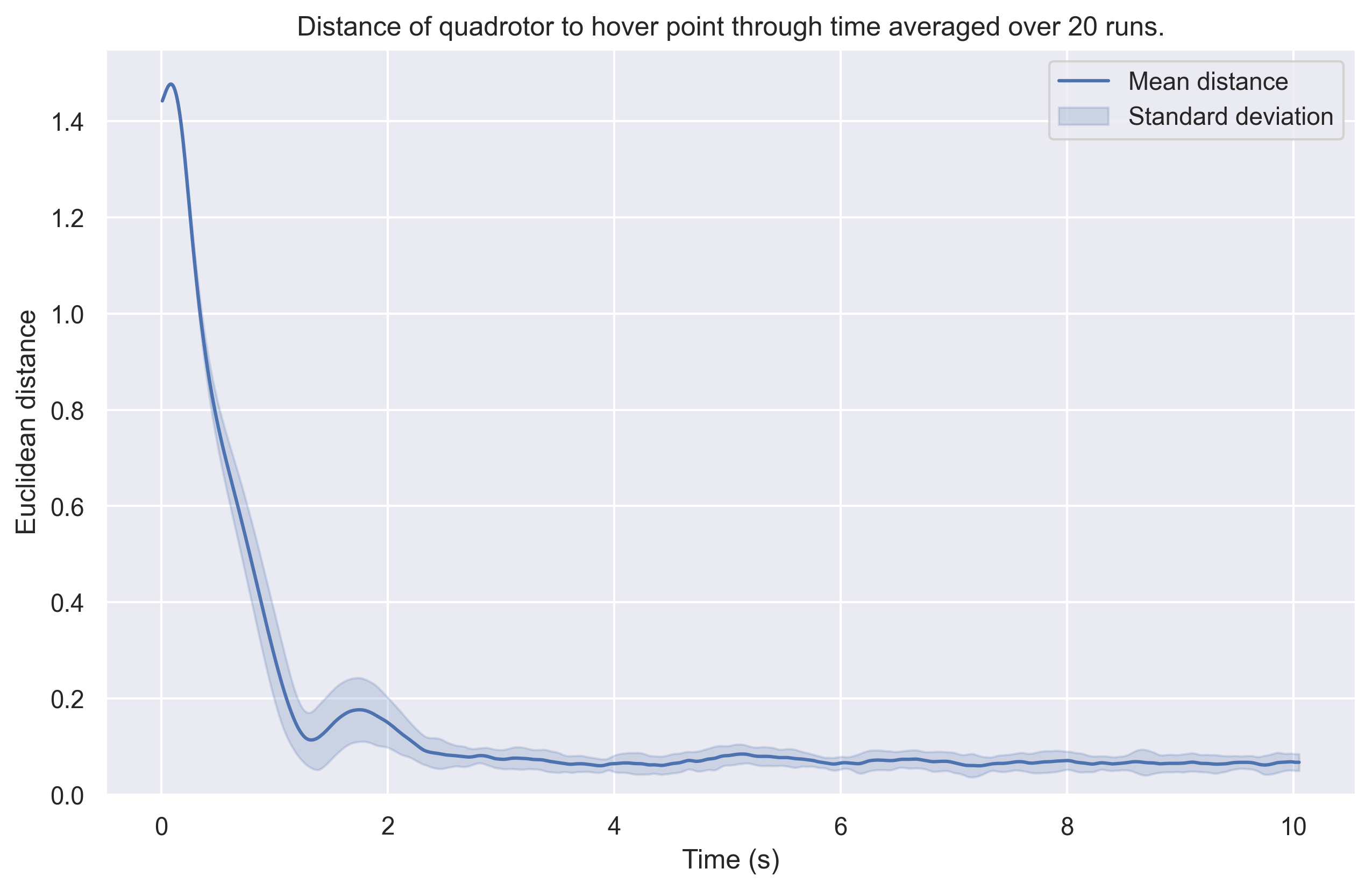}
        \caption{$P_r = (0.0, -1.0, 1.5)$}
        \label{fig:nominal_dist-0.0_-1.0_1.5}
    \end{subfigure}
    \hfill
    \begin{subfigure}[b]{0.32\textwidth}
        \includegraphics[width=\textwidth]{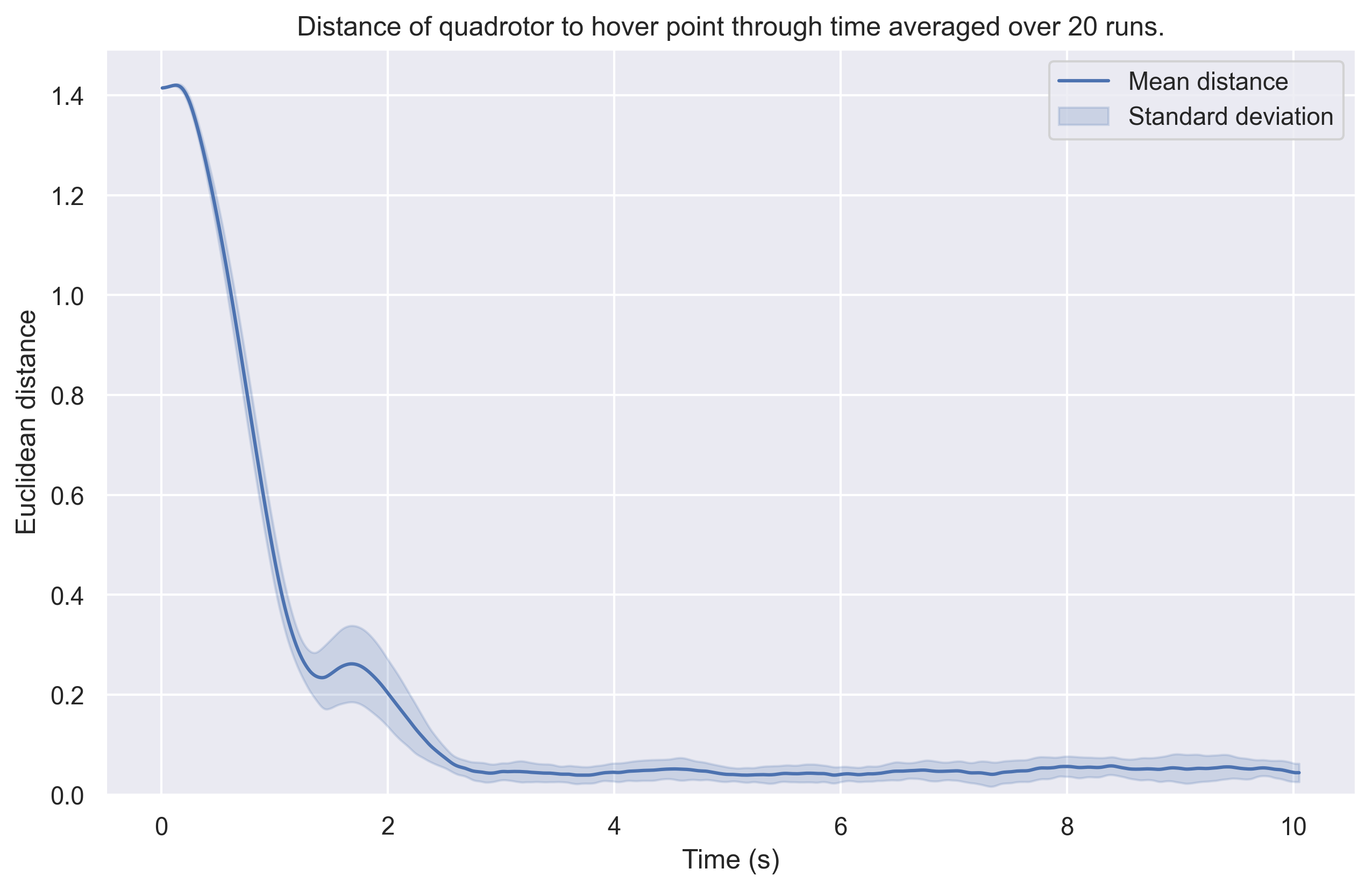}
        \caption{$P_r = (-1.0, -1.0, 0.5)$}
        \label{fig:nominal_dist--1.0_-1.0_0.5}
    \end{subfigure}

    \caption[Trajectory errors of the quadrotor under the nominal controller.]{Euclidean distance of the quadrotor to each of the hovering points over time, starting from its initial position. Each experiment is run twenty times and Euclidean distances are averaged to provide unbiased results. The maximum duration of an experiment is ten seconds, i.e. the expected duration if the quadrotor does not crash.}
    \label{fig:nominal-dist}
\end{figure}

On the other hand, simulation results in Figure \ref{fig:nominal-dist} provide a metric-based analysis for each of the six experiments, illustrating how the quadrotor evolves towards its target position. This is done by plotting the average distance of the quadrotor to the hover point over time, averaging over twenty episodes to guarantee unbiased results.
Subfigures \ref{fig:nominal_dist-0.85_0.9_1.7}, \ref{fig:nominal_dist-0.0_0.0_1.2}, \ref{fig:nominal_dist-0.7_0.85_0.7} and \ref{fig:nominal_dist--1.0_-1.0_0.5} confirm that the quadrotor undertakes smooth and monotonic paths. Specifically, \ref{fig:nominal_dist-0.0_0.0_1.2} shows a straight-up lifting towards the hovering point, and \ref{fig:nominal_dist-0.0_0.0_0.5} demonstrates the stability of the quadrotor when hovering, with oscillations of only $\pm0.015$ units. In addition, the latter proves how short the adaptation time is between the episode's start and the quadrotor's successful stabilisation (less than 0.15s). 
Finally, the little bumps in \ref{fig:nominal_dist-0.0_-1.0_1.5} and \ref{fig:nominal_dist--1.0_-1.0_0.5} can be explained as the quadrotor counters the dragging force resulting from its high thrust to stabilise upon reaching the target position. \\

To conclude on the nominal controller's evaluation, the above experiments showed that the learning-based controller designed in Algorithm \ref{algo:nominal} could successfully control the drone in complete autonomy, within simulation, and under ideal conditions, i.e. without false data injection attacks.

\subsection{Tracking Control of a Quadrotor under False Data Injection Attacks designed by Algorithm \ref{algo:attacker}}
\label{section:eval-sim-attack}

In this subsection, we ran the same experiments as above, but with the quadrotor under false data injection attacks generated by the attack controller designed in Algorithm \ref{algo:attacker}. Additionally, we compared the performance of \textit{optimal} false data injection attacks and those of \textit{random} data injections. \\

\begin{figure}[ht]
    \centering
    \begin{subfigure}[b]{0.32\textwidth}
        \includegraphics[width=\textwidth]{img/attacker/eval/_0.85_0.9_1.7_sim.png}
        \caption{$P_r = (0.85, 0.90, 1.7)$}
        \label{fig:attacker_sim-0.85_0.9_1.7}
    \end{subfigure}
    \hfill
    \begin{subfigure}[b]{0.32\textwidth}
        \includegraphics[width=\textwidth]{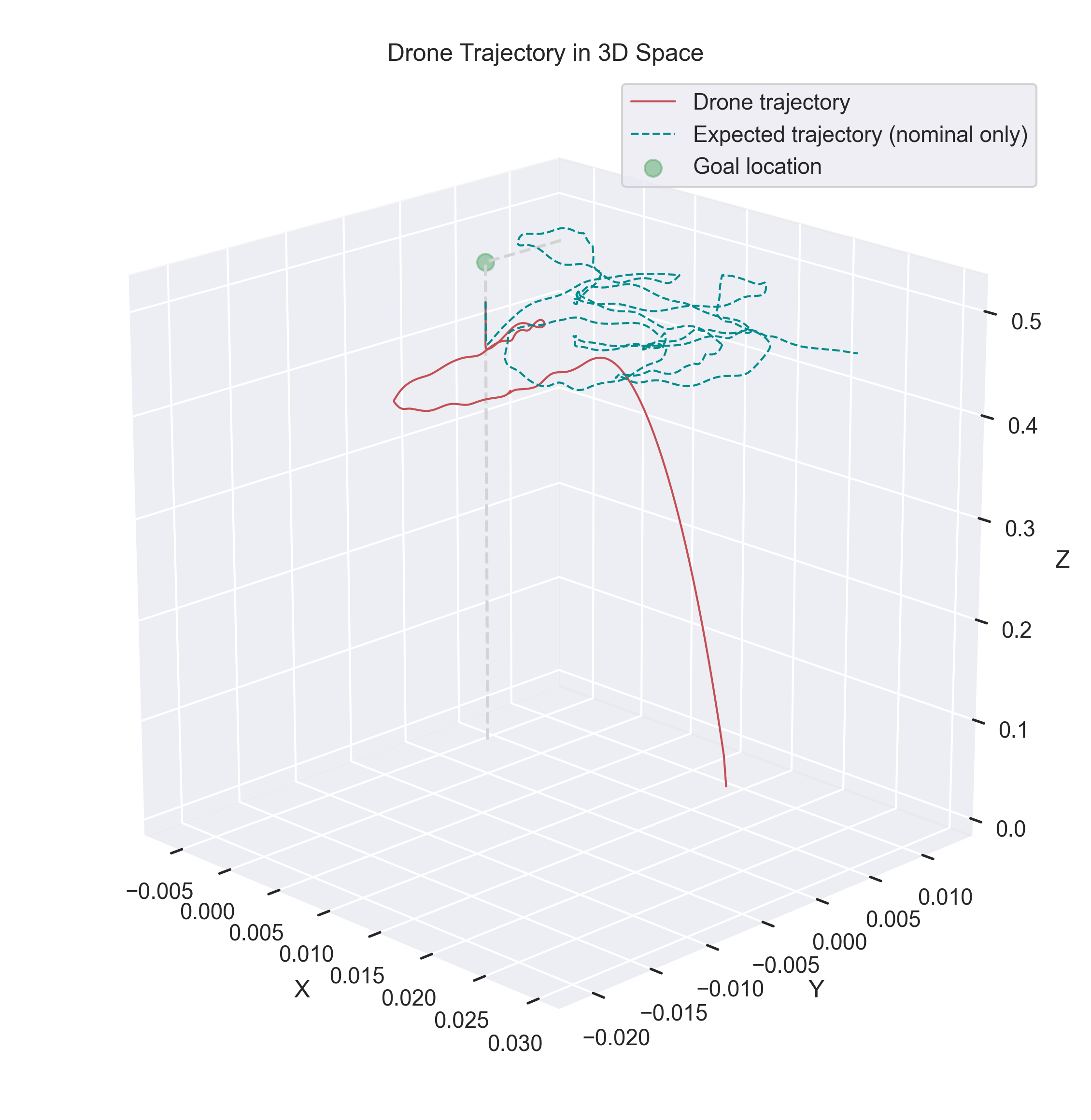}
        \caption{$P_r = (0.0, 0.0, 0.5)$}
        \label{fig:attacker_sim-0.0_0.0_0.5}
    \end{subfigure}
    \hfill
    \begin{subfigure}[b]{0.32\textwidth}
        \includegraphics[width=\textwidth]{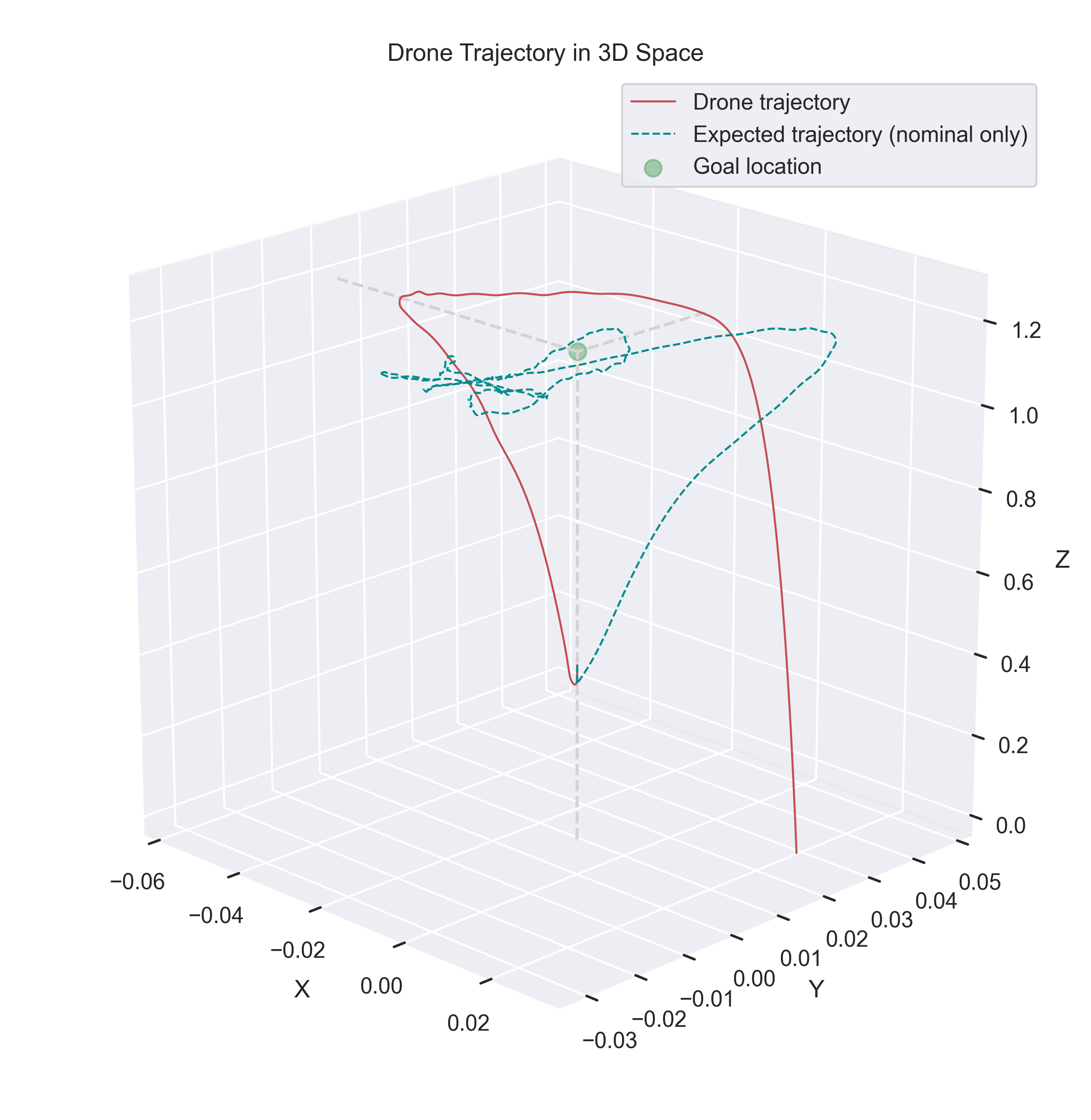}
        \caption{$P_r = (0.0, 0.0, 1.2)$}
        \label{fig:attacker_sim-0.0_0.0_1.2}
    \end{subfigure}

    \begin{subfigure}[b]{0.32\textwidth}
        \includegraphics[width=\textwidth]{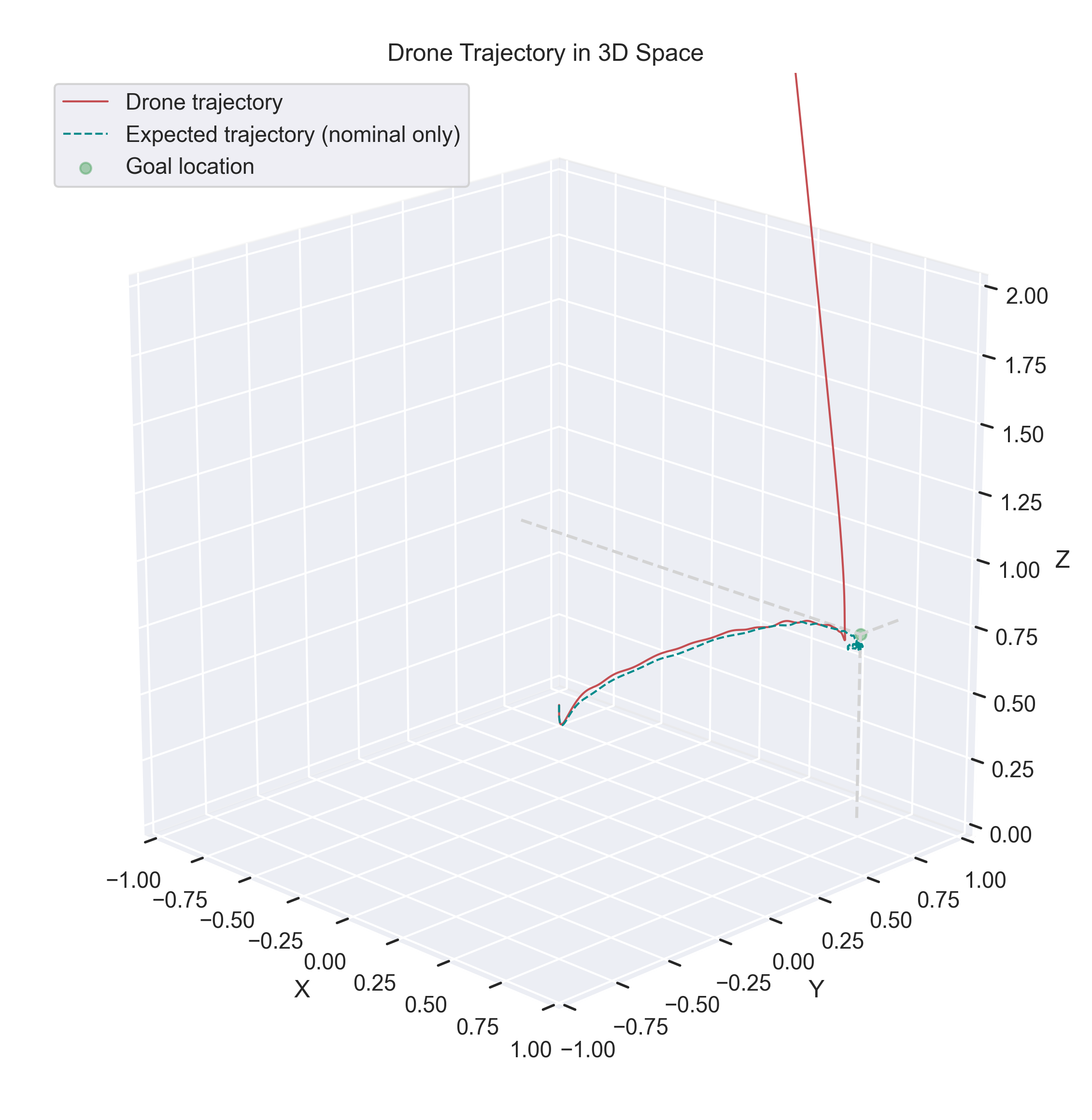}
        \caption{$P_r = (0.7, 0.85, 0.7)$}
        \label{fig:attacker_sim-0.7_0.85_0.7}
    \end{subfigure}
    \hfill
    \begin{subfigure}[b]{0.32\textwidth}
        \includegraphics[width=\textwidth]{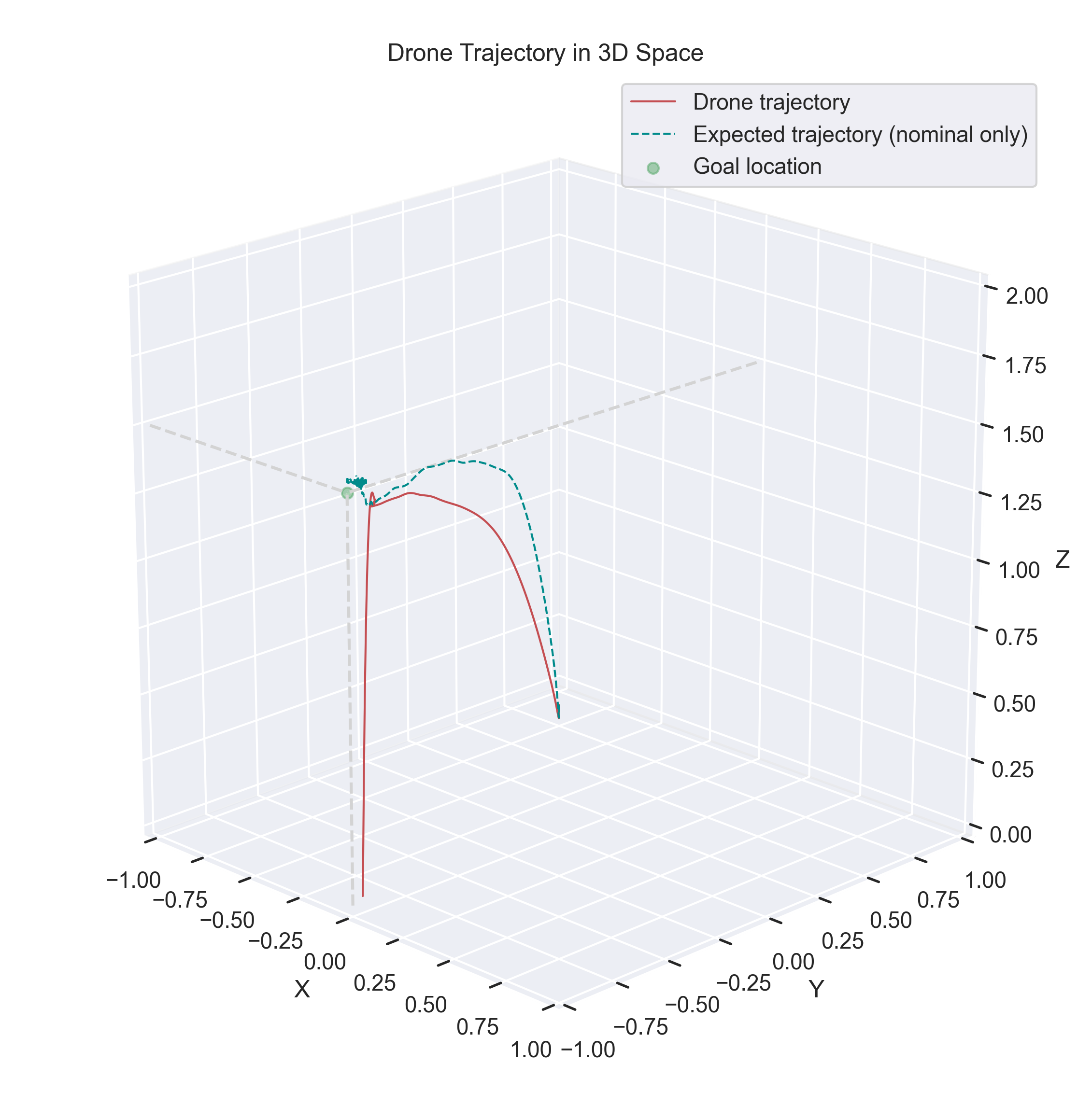}
        \caption{$P_r = (0.0, -1.0, 1.5)$}
        \label{fig:attacker_sim-0.0_-1.0_1.5}
    \end{subfigure}
    \hfill
    \begin{subfigure}[b]{0.32\textwidth}
        \includegraphics[width=\textwidth]{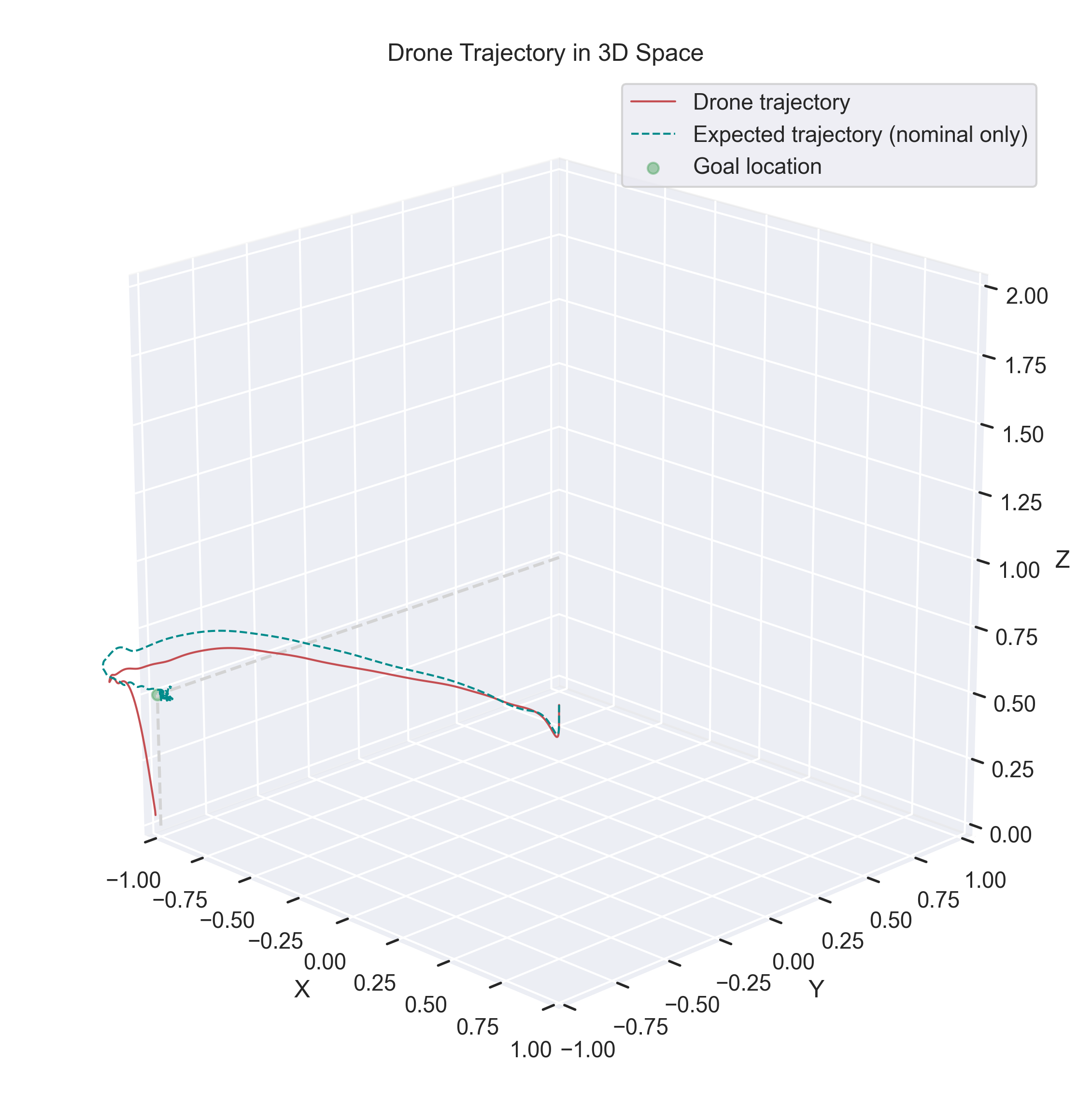}
        \caption{$P_r = (-1.0, -1.0, 0.5)$}
        \label{fig:attacker_sim--1.0_-1.0_0.5}
    \end{subfigure}

    \caption[Trajectories of the quadrotor under optimal attacks.]{Trajectories of the quadrotor under optimal false data injections, towards each of the six hovering points. Each experiment has a maximum duration of ten seconds, i.e. the expected duration if the quadrotor does not crash.}
    \label{fig:attacker-sim}
\end{figure}

Note that the aim is to simulate a false data injection attack in the middle of a quadrotor flight. Therefore, since malicious adversaries often execute attacks when the quadrotor already flies, attacks are launched at $t=2s$ of each episode. The simulation results providing tracking performance of the quadrotor under the learned false data injection attacks are displayed in Figure \ref{fig:attacker-sim}. Subfigures \ref{fig:attacker_sim-0.0_0.0_0.5} and \ref{fig:attacker_sim-0.0_0.0_1.2} are zoomed in for convenience.

As these figures show, although the quadrotor can sometimes reach the hover point when this is done in under two seconds, i.e. the time before the attack starts, it will always end up crashing. Specifically, we can denote two types of failures that the attack controller generated. The first is motor failure. The process is relatively simple: the attacker cuts off the power sent to the individual motors, resulting in an instant crash. The second is maximal boosting, i.e. when all four motors have their power increased to maximal power. Note that while both lead to instant failures, the first one is preferred as it also fulfils the condition of minimal energy usage, as specified in the reward function. \\

\begin{figure}[ht]
    \centering
    \begin{subfigure}[b]{0.32\textwidth}
        \includegraphics[width=\textwidth]{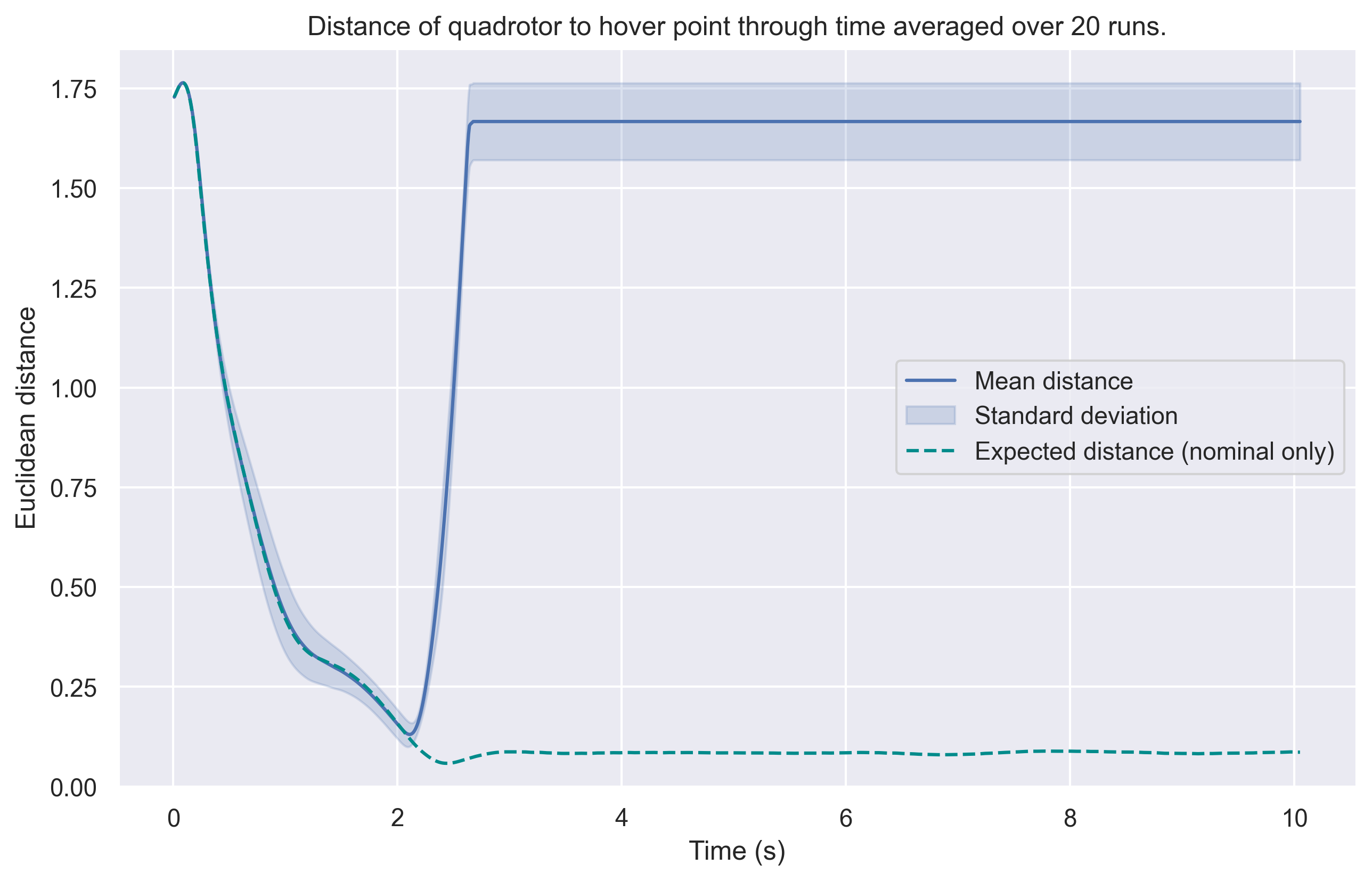}
        \caption{$P_r = (0.85, 0.90, 1.7)$}
        \label{fig:attacker_dist-0.85_0.9_1.7}
    \end{subfigure}
    \hfill
    \begin{subfigure}[b]{0.32\textwidth}
        \includegraphics[width=\textwidth]{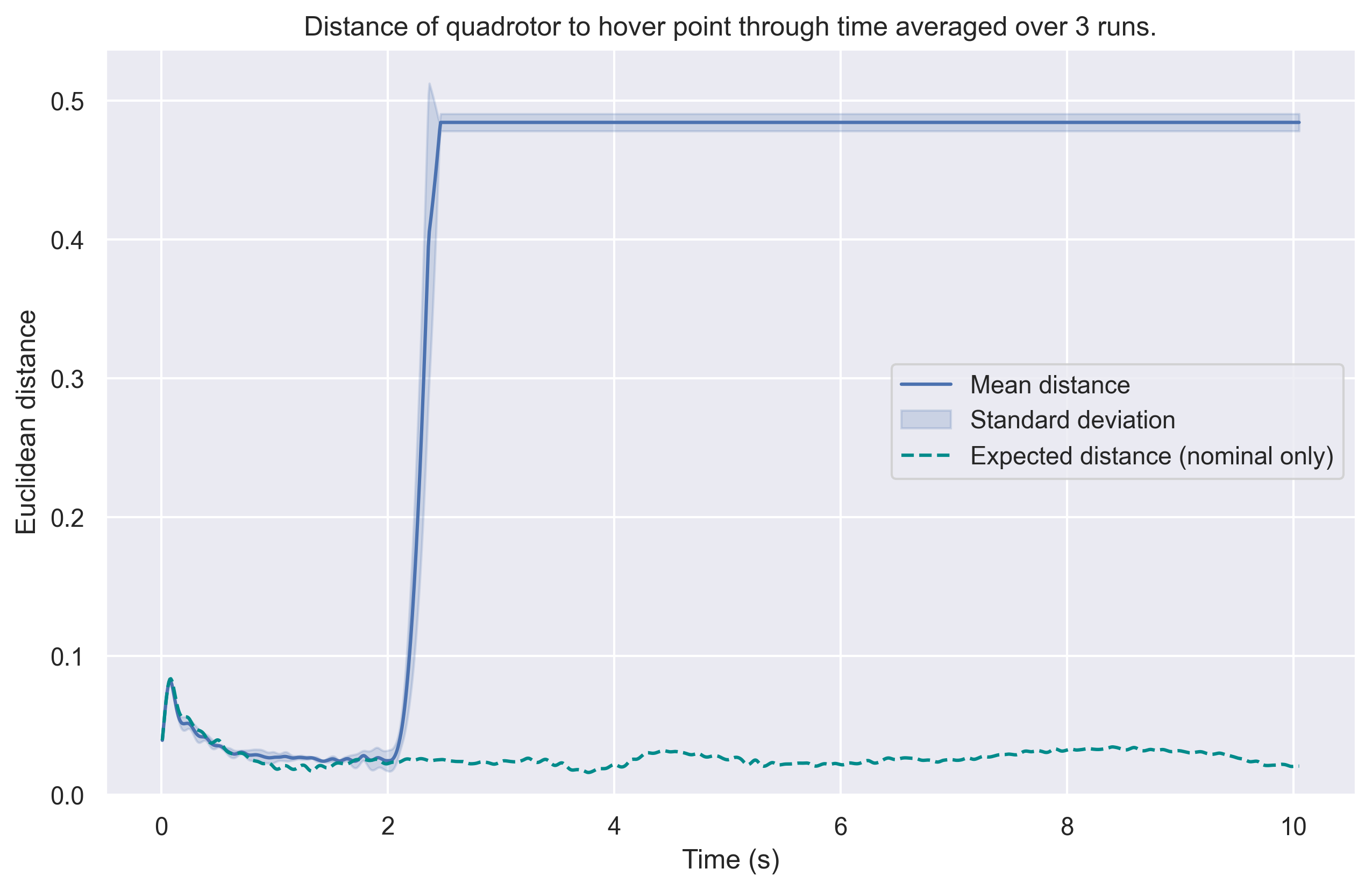}
        \caption{$P_r = (0.0, 0.0, 0.5)$}
        \label{fig:attacker_dist-0.0_0.0_0.5}
    \end{subfigure}
    \hfill
    \begin{subfigure}[b]{0.32\textwidth}
        \includegraphics[width=\textwidth]{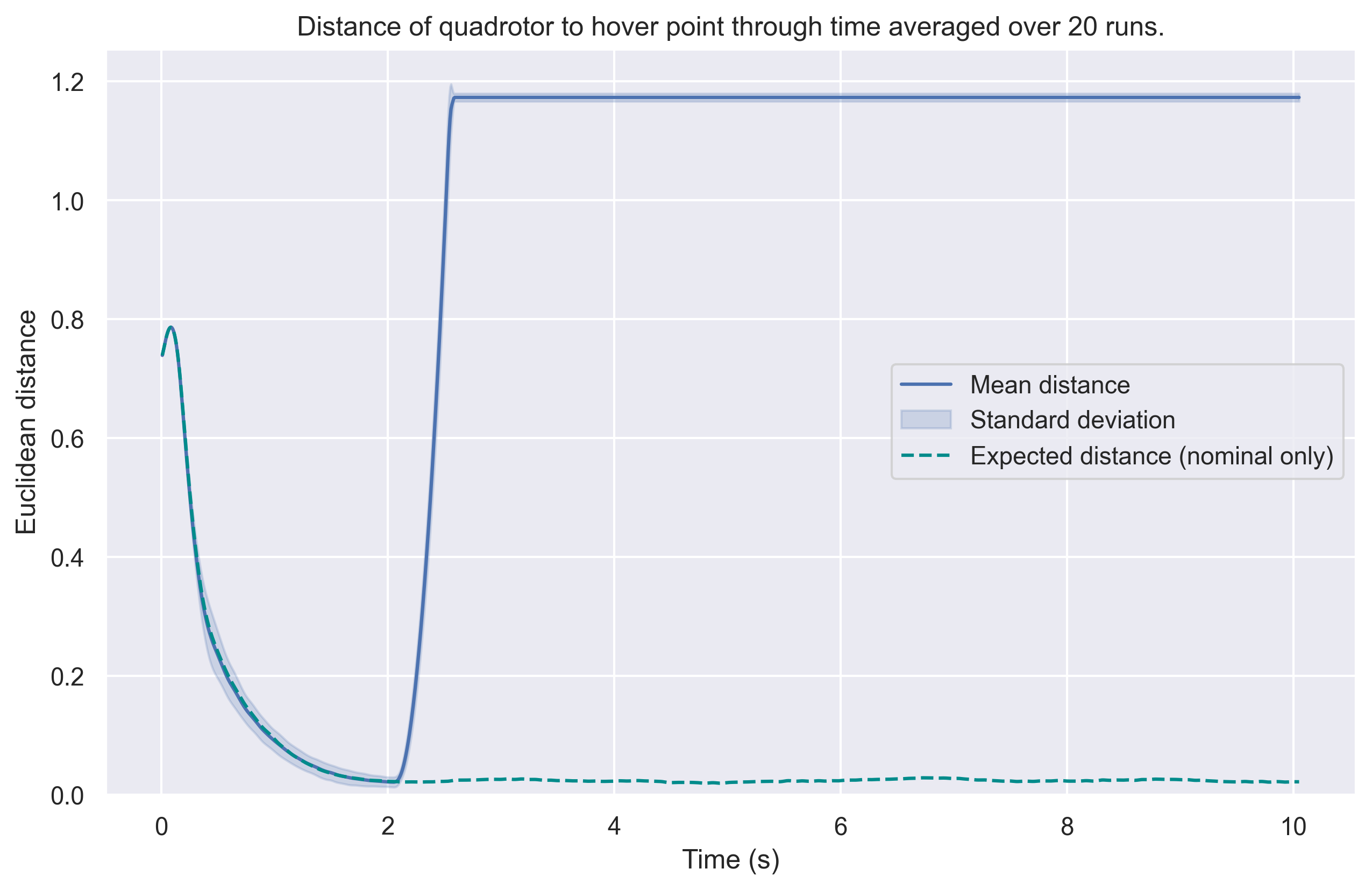}
        \caption{$P_r = (0.0, 0.0, 1.2)$}
        \label{fig:attacker_dist-0.0_0.0_1.2}
    \end{subfigure}

    \begin{subfigure}[b]{0.32\textwidth}
        \includegraphics[width=\textwidth]{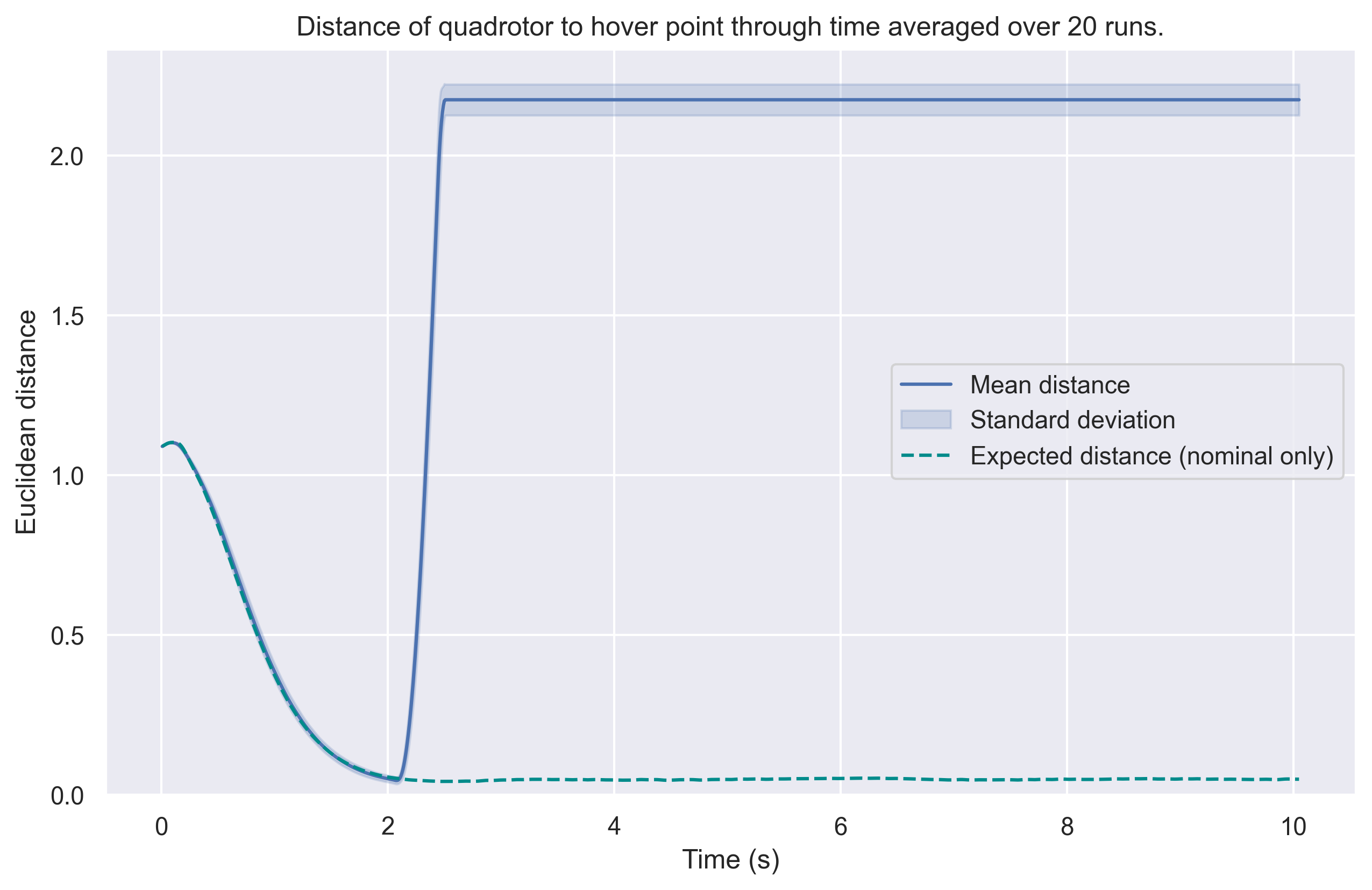}
        \caption{$P_r = (0.7, 0.85, 0.7)$}
        \label{fig:attacker_dist-0.7_0.85_0.7}
    \end{subfigure}
    \hfill
    \begin{subfigure}[b]{0.32\textwidth}
        \includegraphics[width=\textwidth]{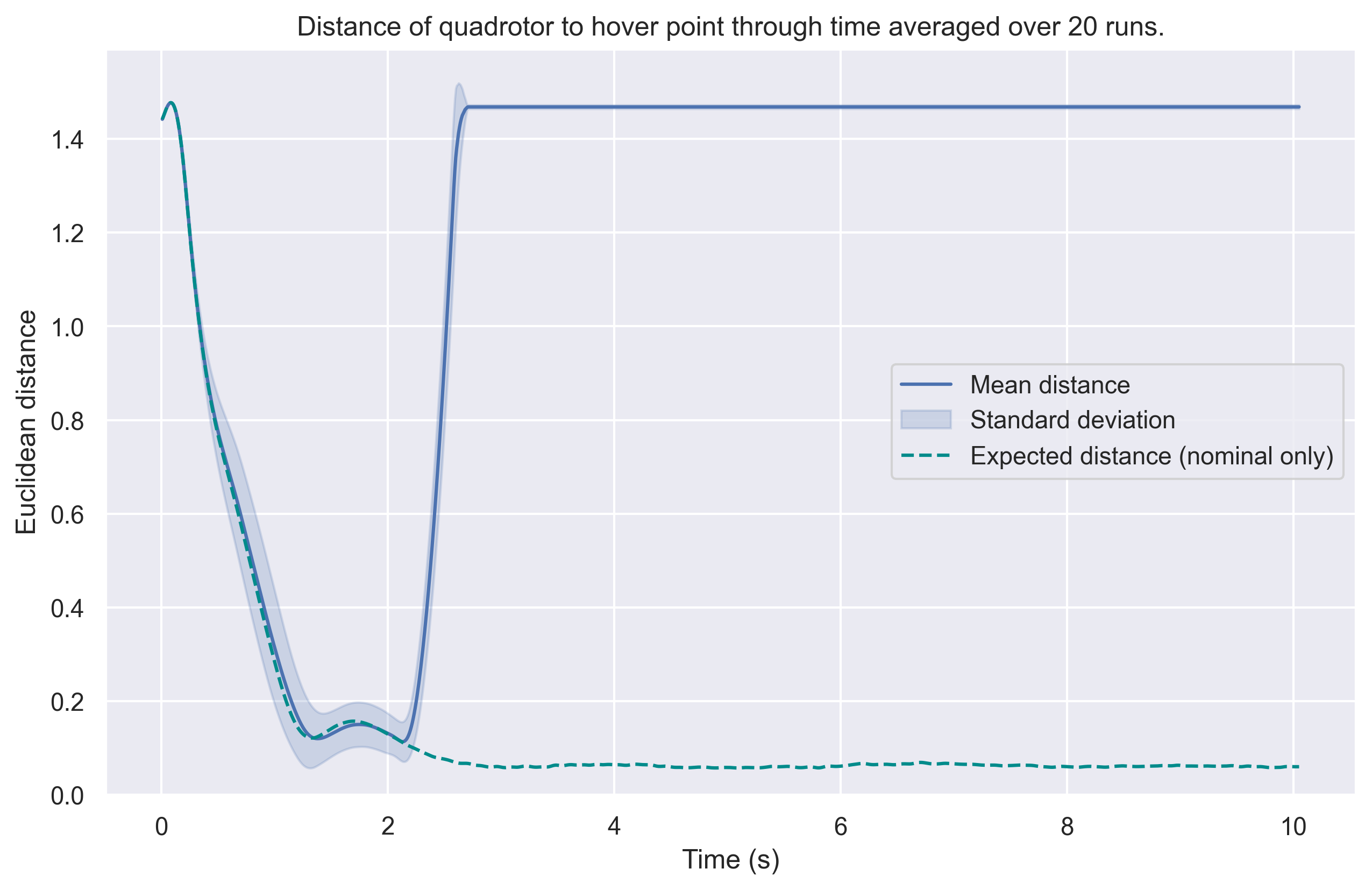}
        \caption{$P_r = (0.0, -1.0, 1.5)$}
        \label{fig:attacker_dist-0.0_-1.0_1.5}
    \end{subfigure}
    \hfill
    \begin{subfigure}[b]{0.32\textwidth}
        \includegraphics[width=\textwidth]{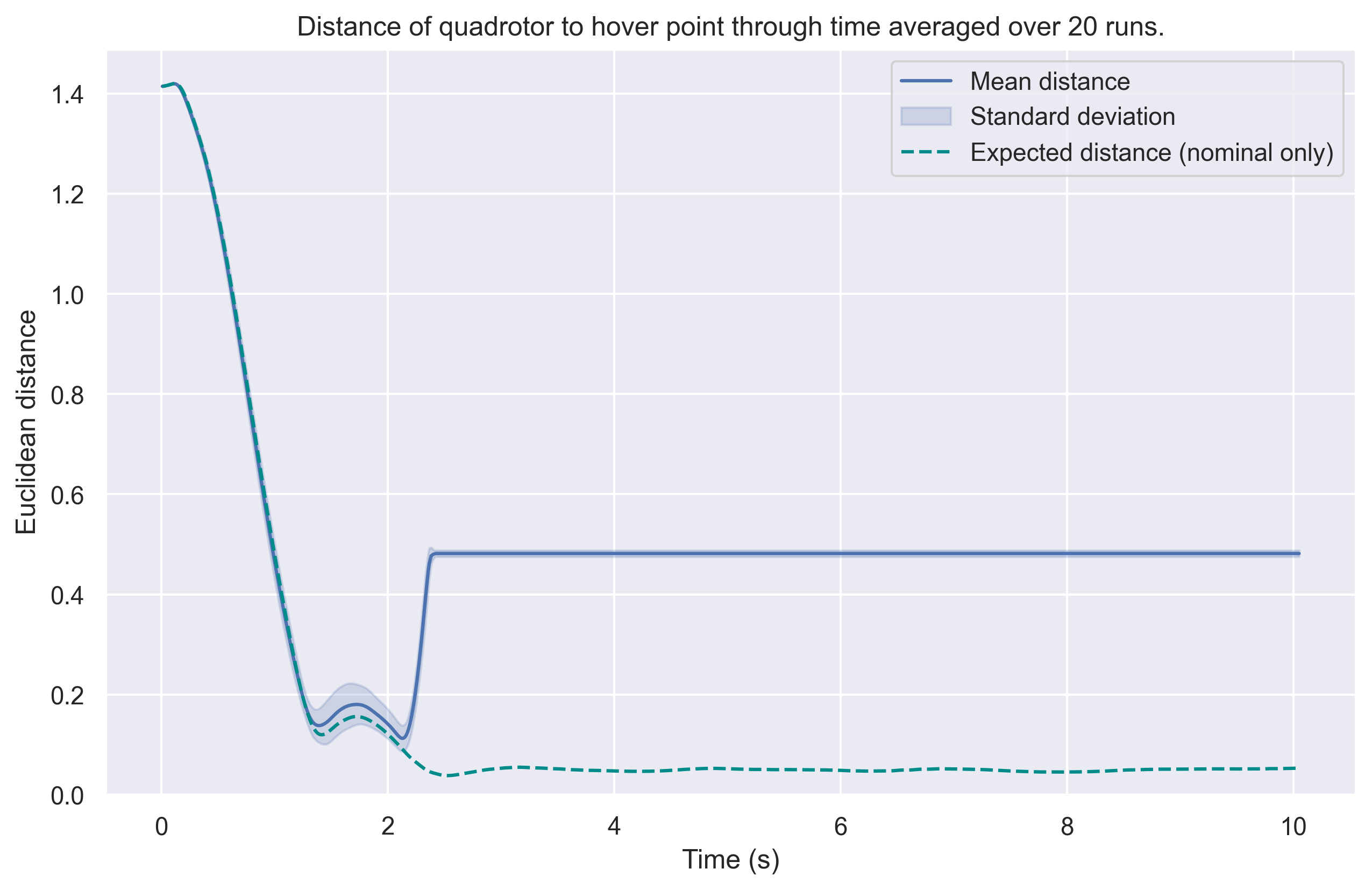}
        \caption{$P_r = (-1.0, -1.0, 0.5)$}
        \label{fig:attacker_dist--1.0_-1.0_0.5}
    \end{subfigure}

    \caption[Trajectory errors of the quadrotor under optimal attacks.]{Euclidean distance of the quadrotor to each of the hovering points over time, starting from its initial position. Each experiment is run twenty times and Euclidean distances are averaged to provide unbiased results. The maximum duration of an experiment is ten seconds, i.e. the expected duration if the quadrotor does not crash.}
    \label{fig:attacker-dist}
\end{figure}

For better visualisation of the quadrotor's position over time relative to the hover point, we plot metric-based analysis for each of the six experiments in Figure \ref{fig:attacker-dist}.

\begin{figure}[ht]
    \centering
    \begin{subfigure}[b]{0.32\textwidth}
        \includegraphics[width=\textwidth]{img/attacker/random/high-altitude.png}
        \caption{$P_r = (0.85, 0.90, 1.7)$}
        \label{fig:attacker_random-0.85_0.9_1.7}
    \end{subfigure}
    \hfill
    \begin{subfigure}[b]{0.32\textwidth}
        \includegraphics[width=\textwidth]{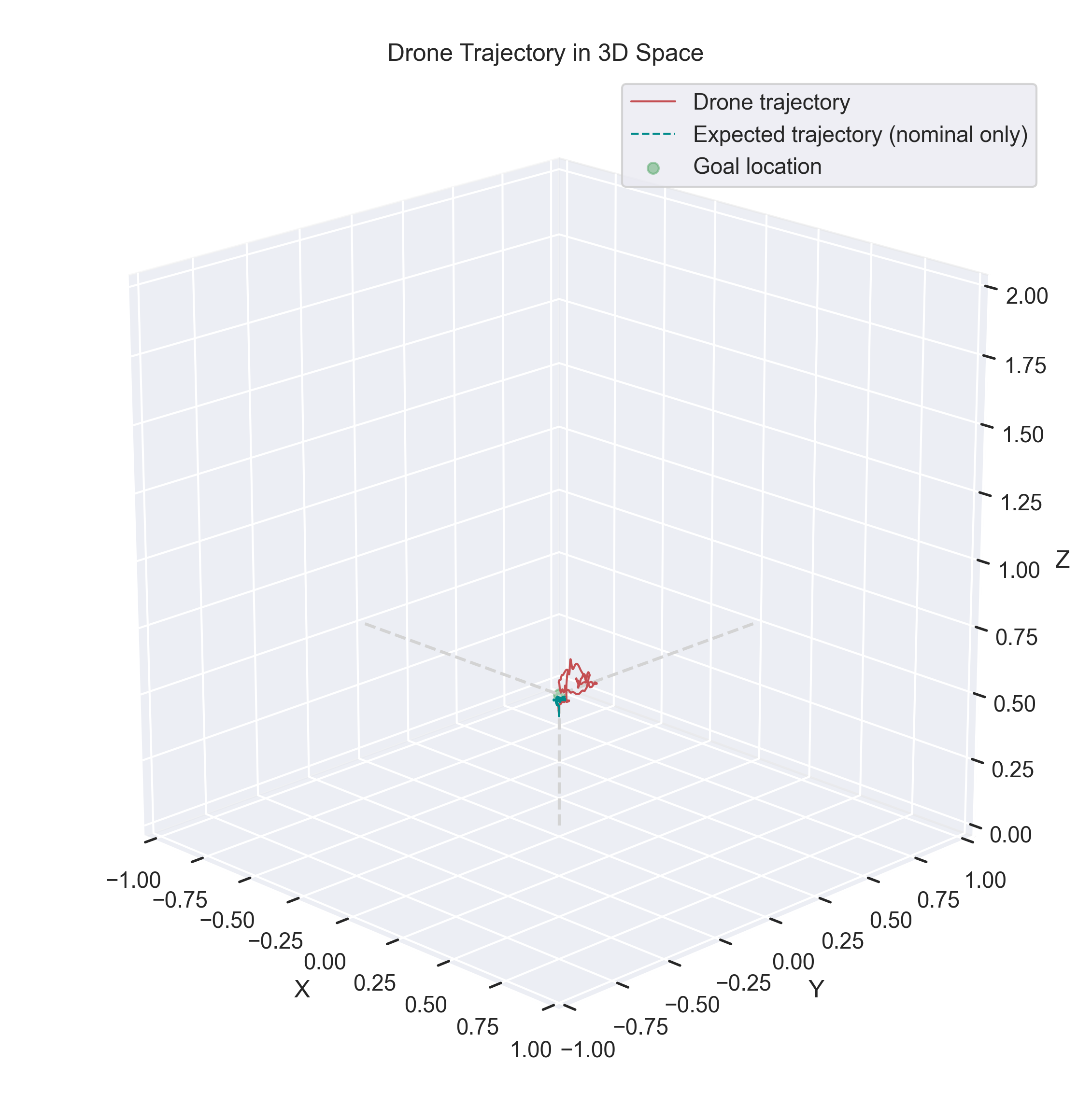}
        \caption{$P_r = (0.0, 0.0, 0.5)$}
        \label{fig:attacker_random-0.0_0.0_0.5}
    \end{subfigure}
    \hfill
    \begin{subfigure}[b]{0.32\textwidth}
        \includegraphics[width=\textwidth]{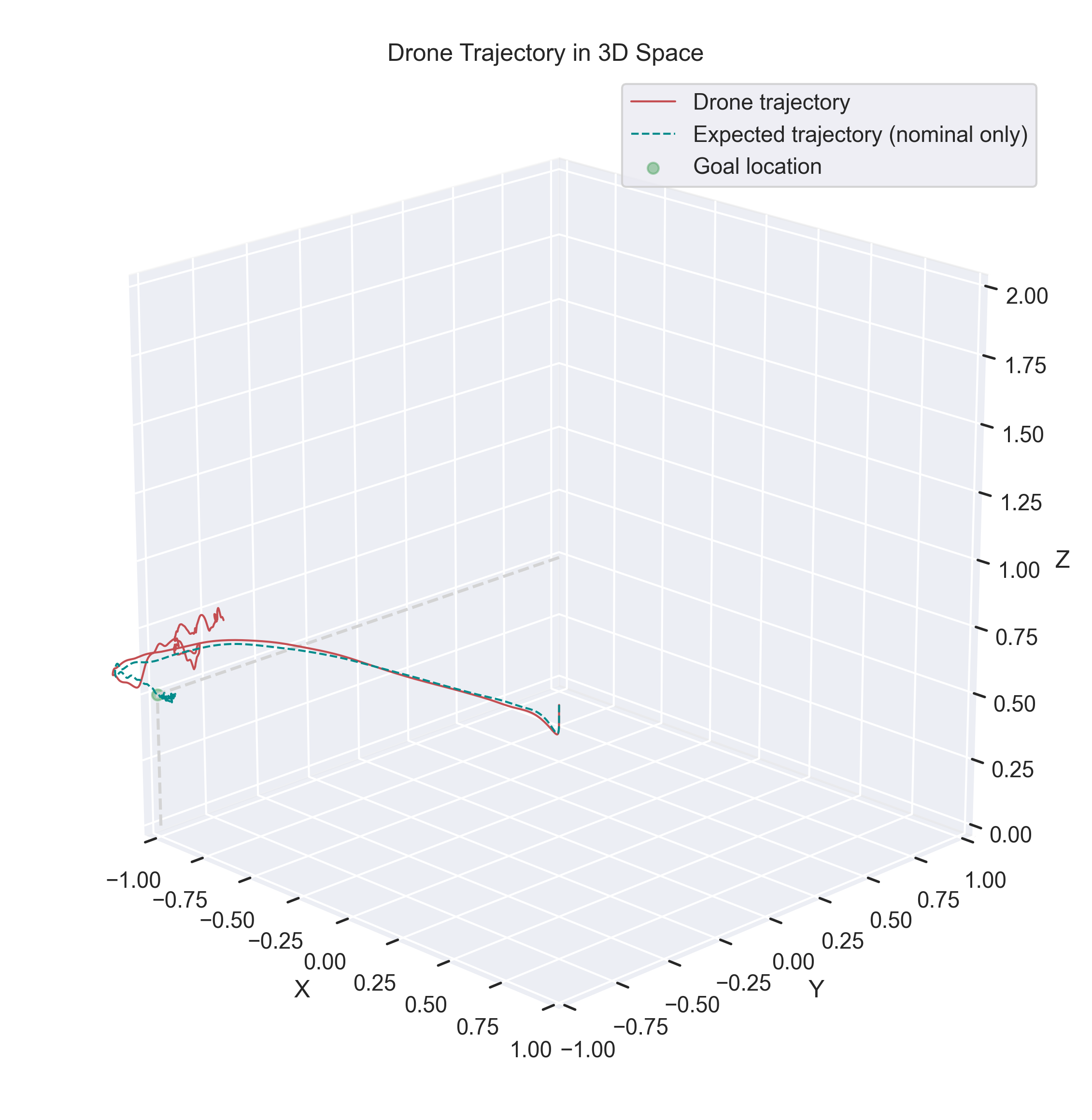}
        \caption{$P_r = (-1.0, -1.0, 0.5)$}
        \label{fig:attacker_random--1.0_-1.0_0.5}
    \end{subfigure}

    \caption[Trajectories of the quadrotor under optimal vs. random attacks.]{Trajectories of the quadrotor under \textit{random} injection attacks, towards the hovering points (a), (b) and (f).}
    \label{fig:attacker-random}
\end{figure}

\clearpage
As a comparison, Figure \ref{fig:attacker-random} displayed the trajectories of the quadrotor under random false data injections, \textbf{without} any countermeasure applied.

Although random attacks are often considered in the literature as able to deteriorate tracking performance, the optimal attacks designed in this study outperformed the former by a significant margin. This further indicates that the attacker's performance depends on both the amount of information available and how it is used. \\

The given results clearly show that the desired trajectories cannot be followed anymore under optimal attacks, with all six experiments resulting in a quick failure of the quadrotor.
Although a robust nominal controller is used, the quadrotor's tracking performance deteriorates as soon as optimal false data injections occur. When such disruption occurs, the quadrotor may crash into obstacles or other robots, causing expensive damage. Indeed, if robots in an industrial production line are attacked, defective products may be produced. More drastically, the quadrotor could even crash into humans and cause severe injuries. \\

Our simulation results proved the effectiveness of the proposed optimal attack scheme and highlighted the necessity and significance of securing quadrotors.

\clearpage
\subsection{Tracking Control of a Quadrotor under the Secure Controller learned using Algorithm \ref{algo:defender}}
\label{section:eval-def}

In this subsection, we ran the same experiments as above. However, we added the secure controller to show that the tracking performance of an attacked quadrotor could be recovered by using the countermeasure designed in Algorithm \ref{algo:defender}. \\

\begin{figure}[ht]
    \centering
    \begin{subfigure}[b]{0.32\textwidth}
        \includegraphics[width=\textwidth]{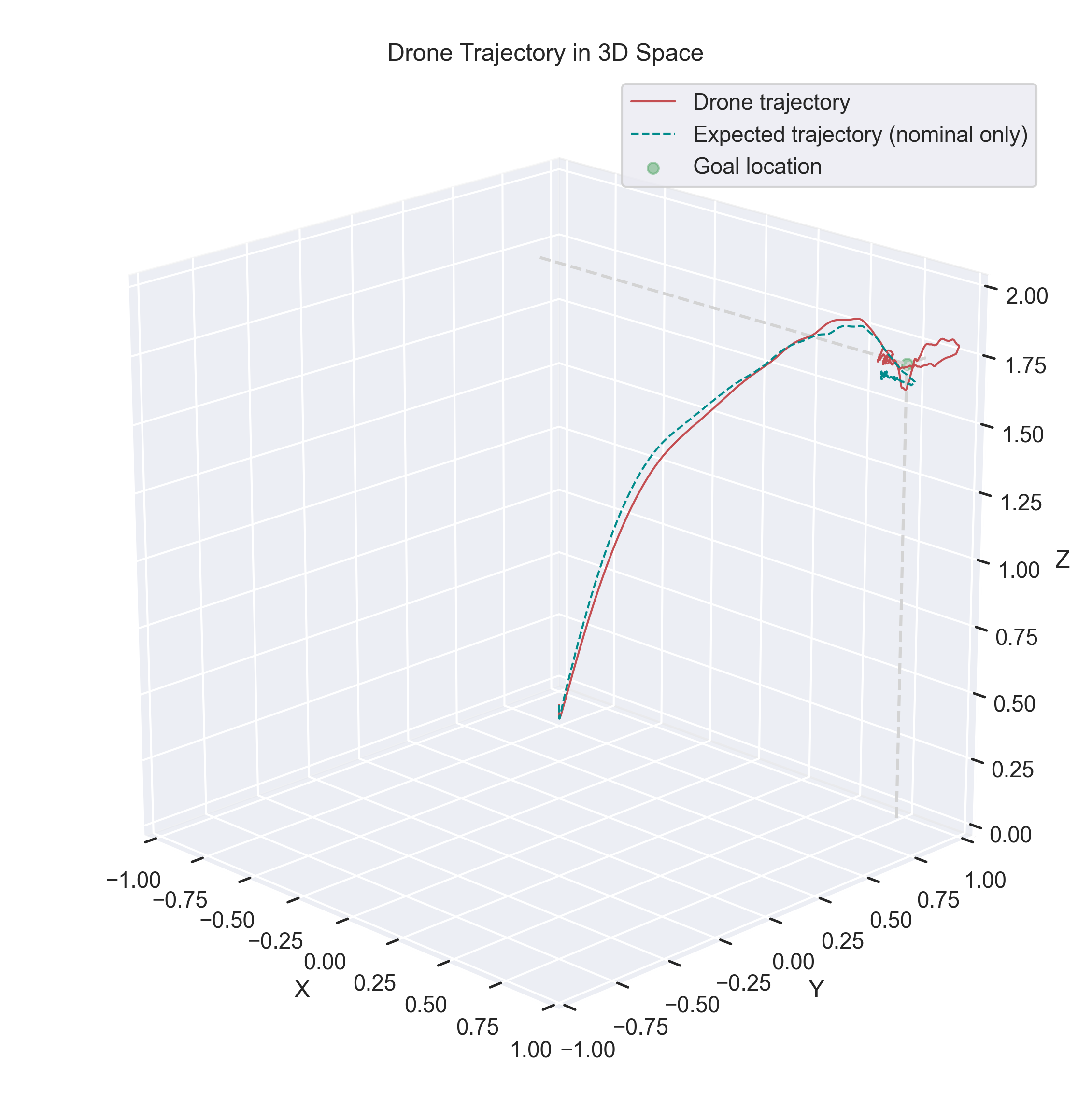}
        \caption{$P_r = (0.85, 0.90, 1.7)$}
        \label{fig:defender_sim-0.85_0.9_1.7}
    \end{subfigure}
    \hfill
    \begin{subfigure}[b]{0.32\textwidth}
        \includegraphics[width=\textwidth]{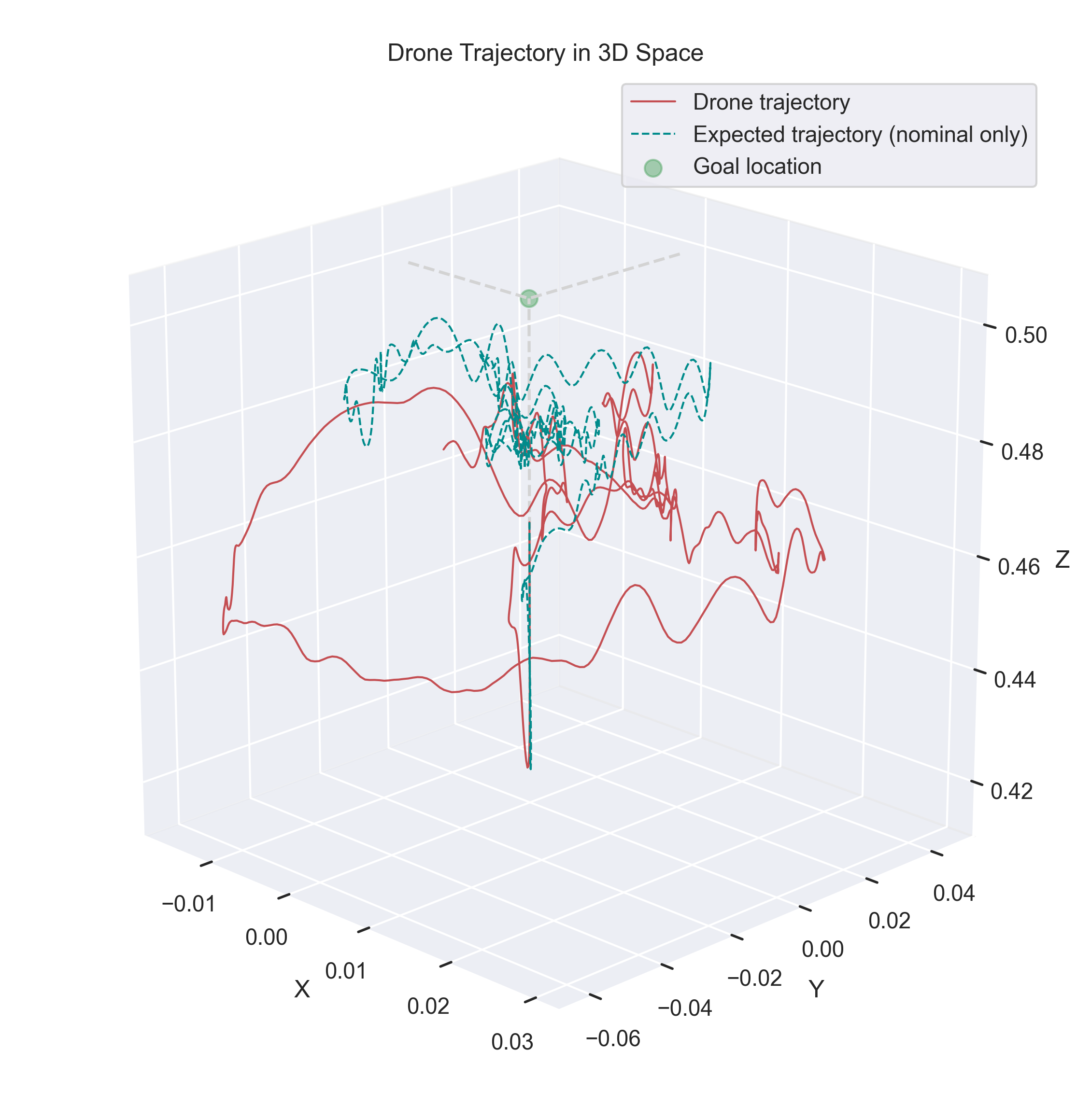}
        \caption{$P_r = (0.0, 0.0, 0.5)$}
        \label{fig:defender_sim-0.0_0.0_0.5}
    \end{subfigure}
    \hfill
    \begin{subfigure}[b]{0.32\textwidth}
        \includegraphics[width=\textwidth]{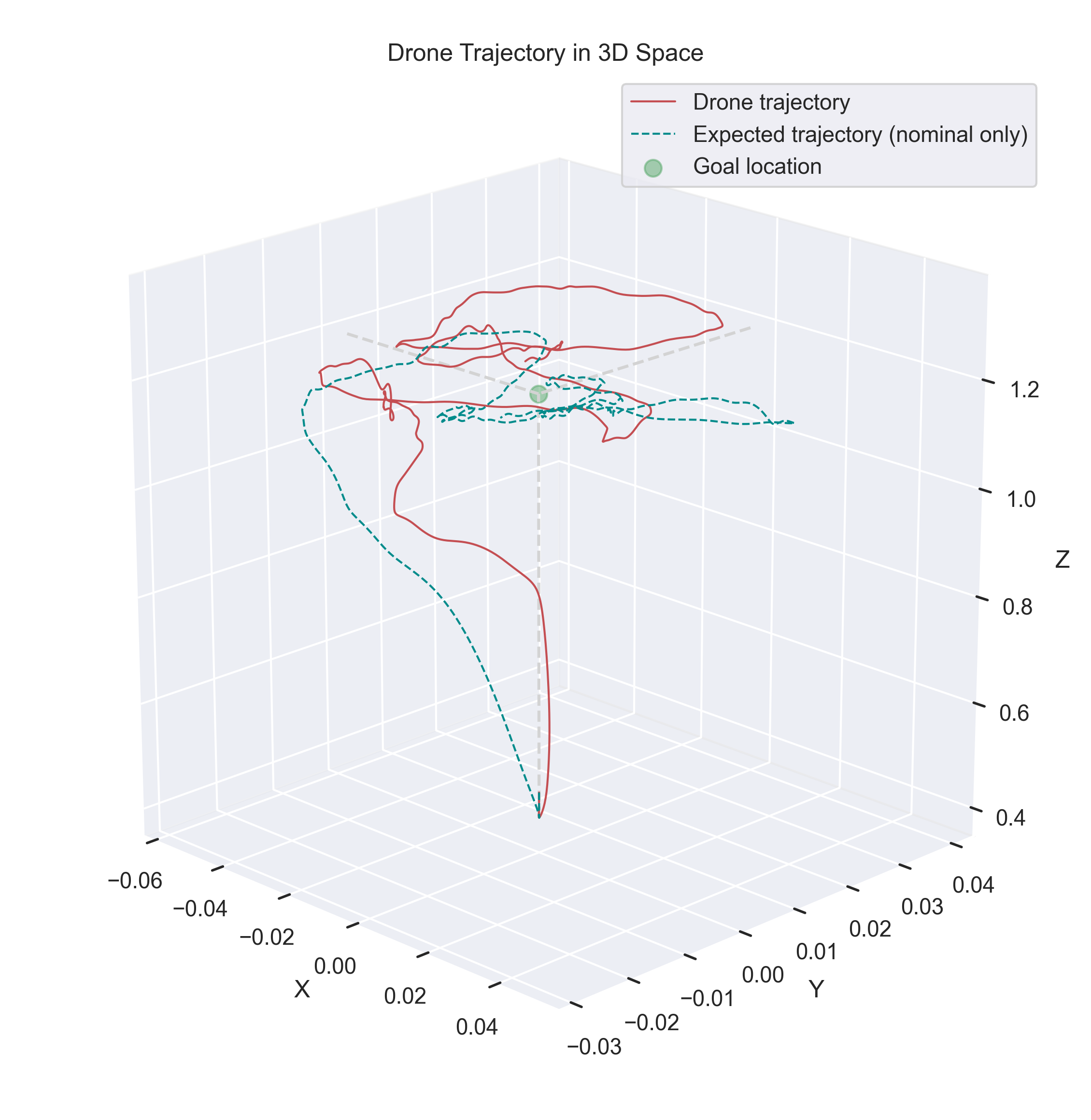}
        \caption{$P_r = (0.0, 0.0, 1.2)$}
        \label{fig:defender_sim-0.0_0.0_1.2}
    \end{subfigure}

    \begin{subfigure}[b]{0.32\textwidth}
        \includegraphics[width=\textwidth]{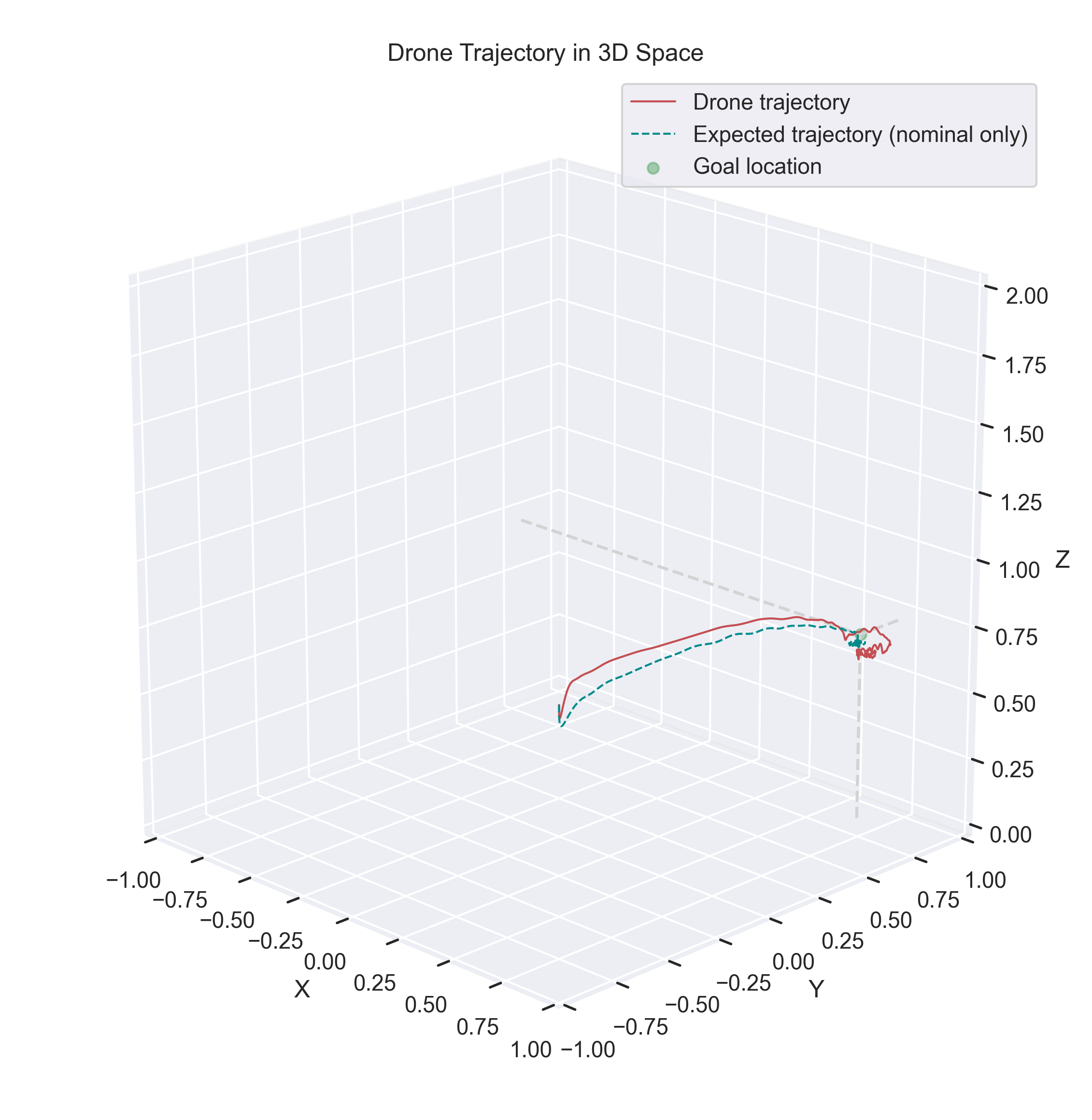}
        \caption{$P_r = (0.7, 0.85, 0.7)$}
        \label{fig:defender_sim-0.7_0.85_0.7}
    \end{subfigure}
    \hfill
    \begin{subfigure}[b]{0.32\textwidth}
        \includegraphics[width=\textwidth]{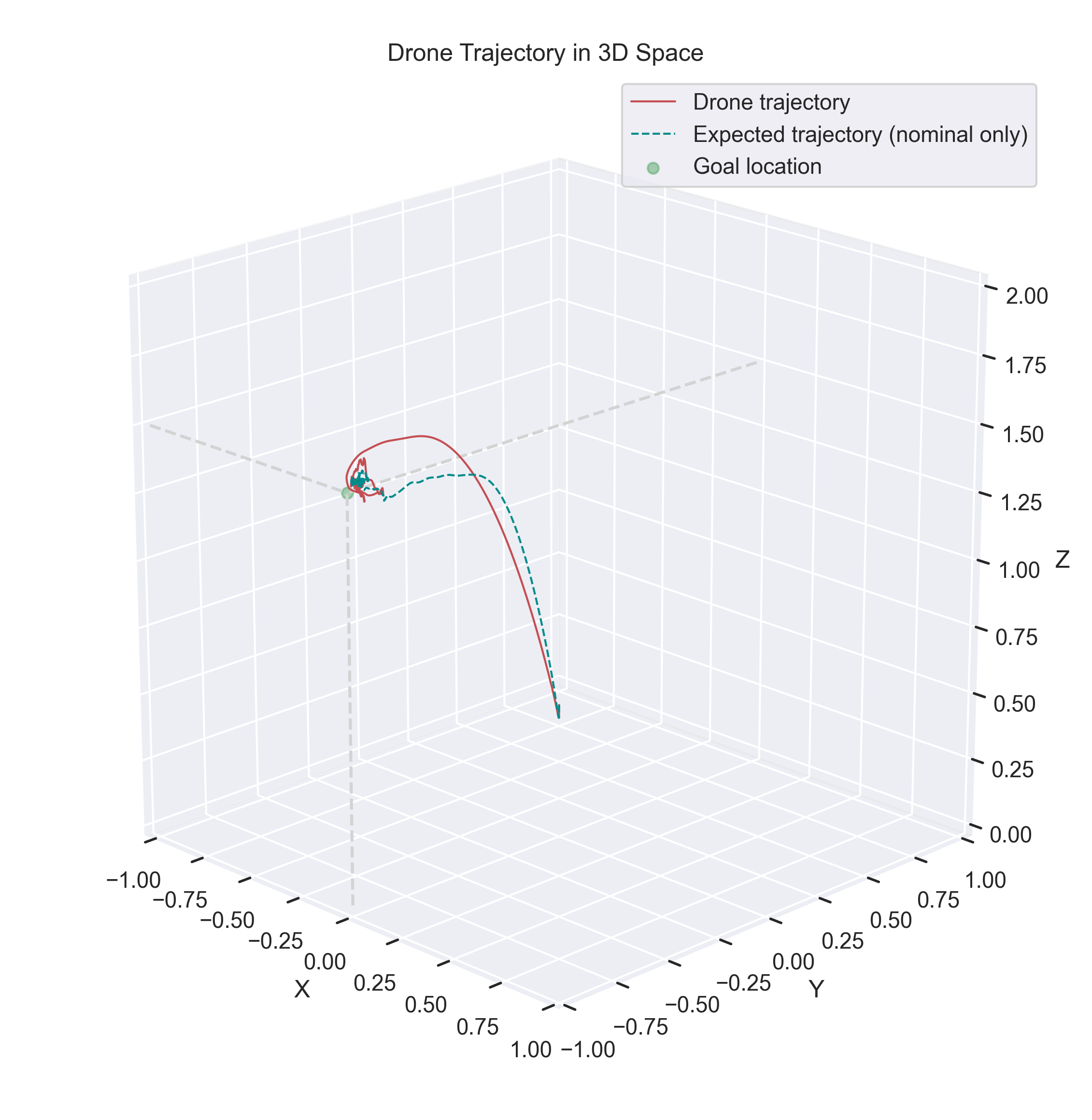}
        \caption{$P_r = (0.0, -1.0, 1.5)$}
        \label{fig:defender_sim-0.0_-1.0_1.5}
    \end{subfigure}
    \hfill
    \begin{subfigure}[b]{0.32\textwidth}
        \includegraphics[width=\textwidth]{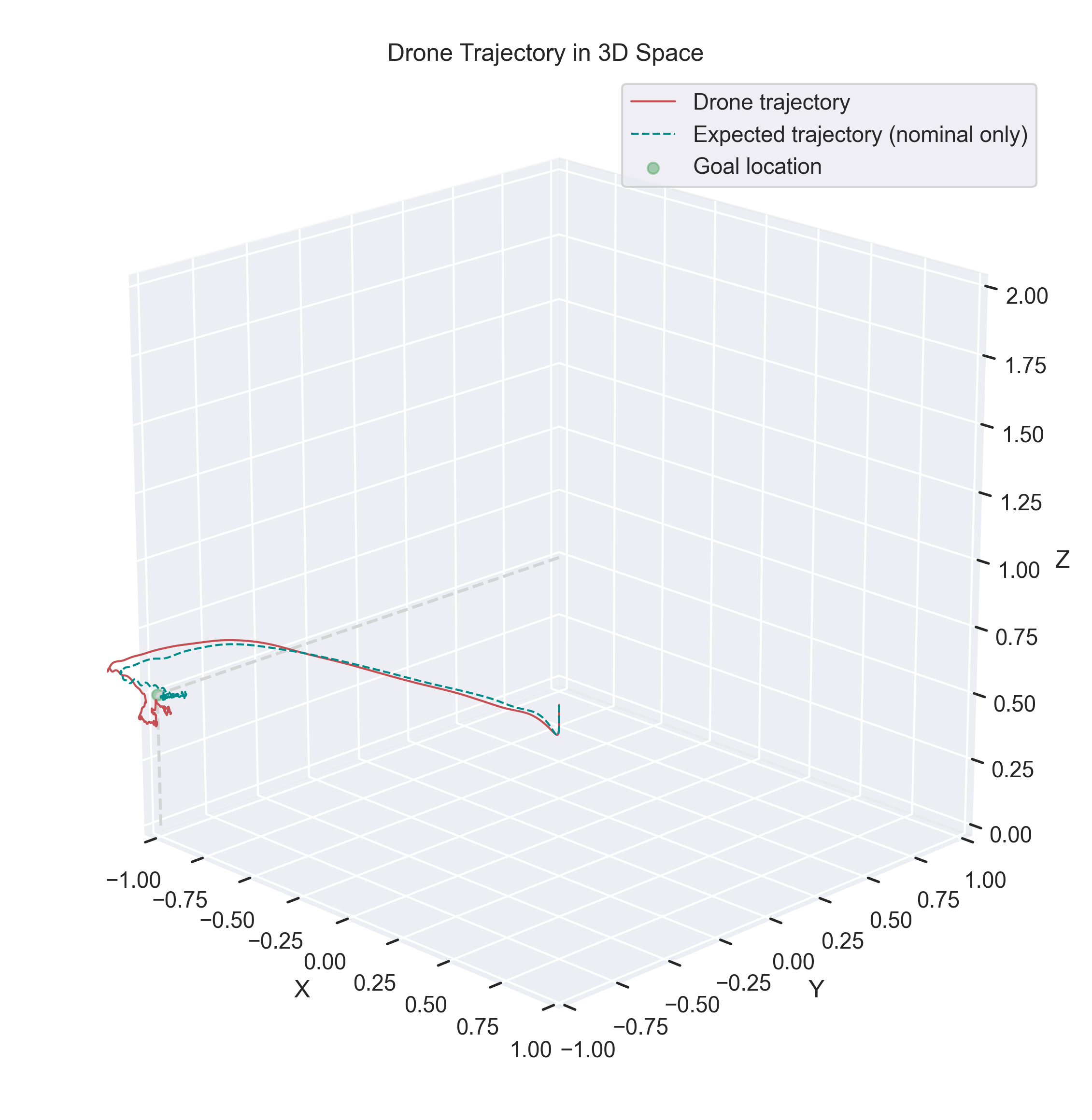}
        \caption{$P_r = (-1.0, -1.0, 0.5)$}
        \label{fig:defender_sim--1.0_-1.0_0.5}
    \end{subfigure}

    \caption[Trajectories of the quadrotor under secure countermeasures with optimal attacks.]{Trajectories of the quadrotor under optimal attacks and secure countermeasures, towards each of the six hovering points. Each experiment has a maximum duration of ten seconds, i.e. the expected duration if the quadrotor does not crash.}
    \label{fig:defender-sim}
\end{figure}

The simulation results providing tracking performance of the quadrotor under secure countermeasures are displayed in Figure \ref{fig:defender-sim}. Subfigures \ref{fig:defender_sim-0.0_0.0_0.5} and \ref{fig:defender_sim-0.0_0.0_1.2} are zoomed in for convenience. 

These results demonstrate that the quadrotor's tracking performance under the secure controller is recovered in all six experiments. Specifically, the controller is able to provide countermeasures for the two types of attack proposed by the malicious attacker, i.e., motor failure and maximal boosting.

Subfigures \ref{fig:defender_sim-0.85_0.9_1.7} and \ref{fig:defender_sim-0.0_-1.0_1.5} illustrate the capabilities of the defender to recover the trajectory and stabilise even in wide-amplitude and high-altitude settings. 
In addition, Subfigures \ref{fig:defender_sim-0.7_0.85_0.7} and \ref{fig:defender_sim--1.0_-1.0_0.5} exhibit the ability of the quadrotor to preserve tracking performance in low-altitude cases. 
Finally, subfigures \ref{fig:defender_sim-0.0_0.0_0.5} and \ref{fig:defender_sim-0.0_0.0_1.2} showcase the secure controller in lifting and hovering settings. We can observe that the secure controller can indeed hover with minimal displacement caused by the false data injections.

These results demonstrate that the proposed secure control algorithm could mitigate optimal attacks and recover tracking performance disrupted by optimal false data injections. \\

\begin{figure}[ht]
    \centering
    \begin{subfigure}[b]{0.32\textwidth}
        \includegraphics[width=\textwidth]{img/attacker/random/high-altitude.png}
        \caption{$P_r = (0.85, 0.90, 1.7)$}
        \label{fig:no_defender_random-0.85_0.9_1.7}
    \end{subfigure}
    \hfill
    \begin{subfigure}[b]{0.32\textwidth}
        \includegraphics[width=\textwidth]{img/attacker/random/hover.png}
        \caption{$P_r = (0.0, 0.0, 0.5)$}
        \label{fig:no_defender_random-0.0_0.0_0.5}
    \end{subfigure}
    \hfill
    \begin{subfigure}[b]{0.32\textwidth}
        \includegraphics[width=\textwidth]{img/attacker/random/low-altitude.png}
        \caption{$P_r = (-1.0, -1.0, 0.5)$}
        \label{fig:no_defender_random--1.0_-1.0_0.5}
    \end{subfigure}

    \begin{subfigure}[b]{0.32\textwidth}
        \includegraphics[width=\textwidth]{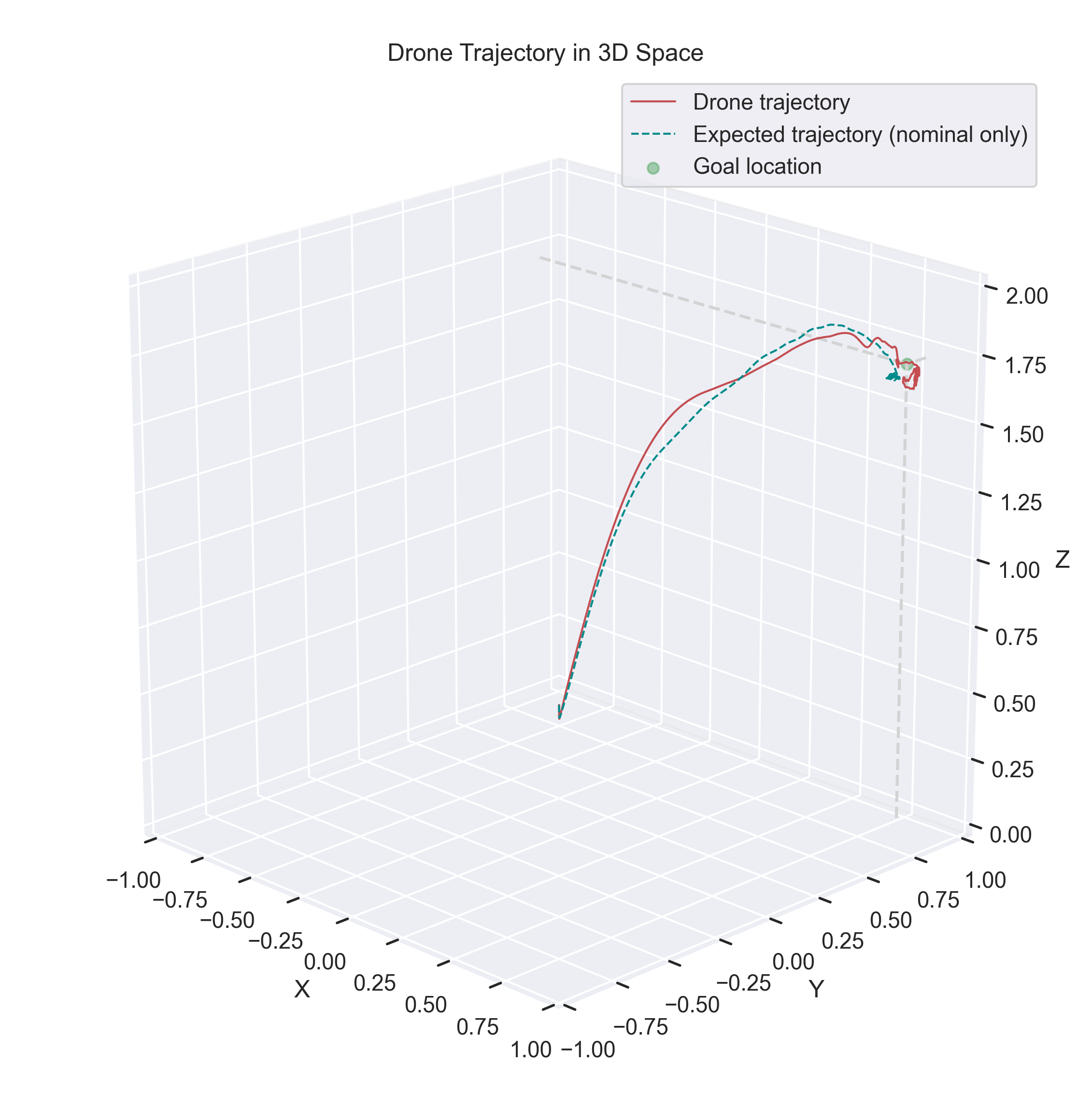}
        \caption{$P_r = (0.85, 0.90, 1.7)$}
        \label{fig:defender_random-0.85_0.9_1.7}
    \end{subfigure}
    \hfill
    \begin{subfigure}[b]{0.32\textwidth}
        \includegraphics[width=\textwidth]{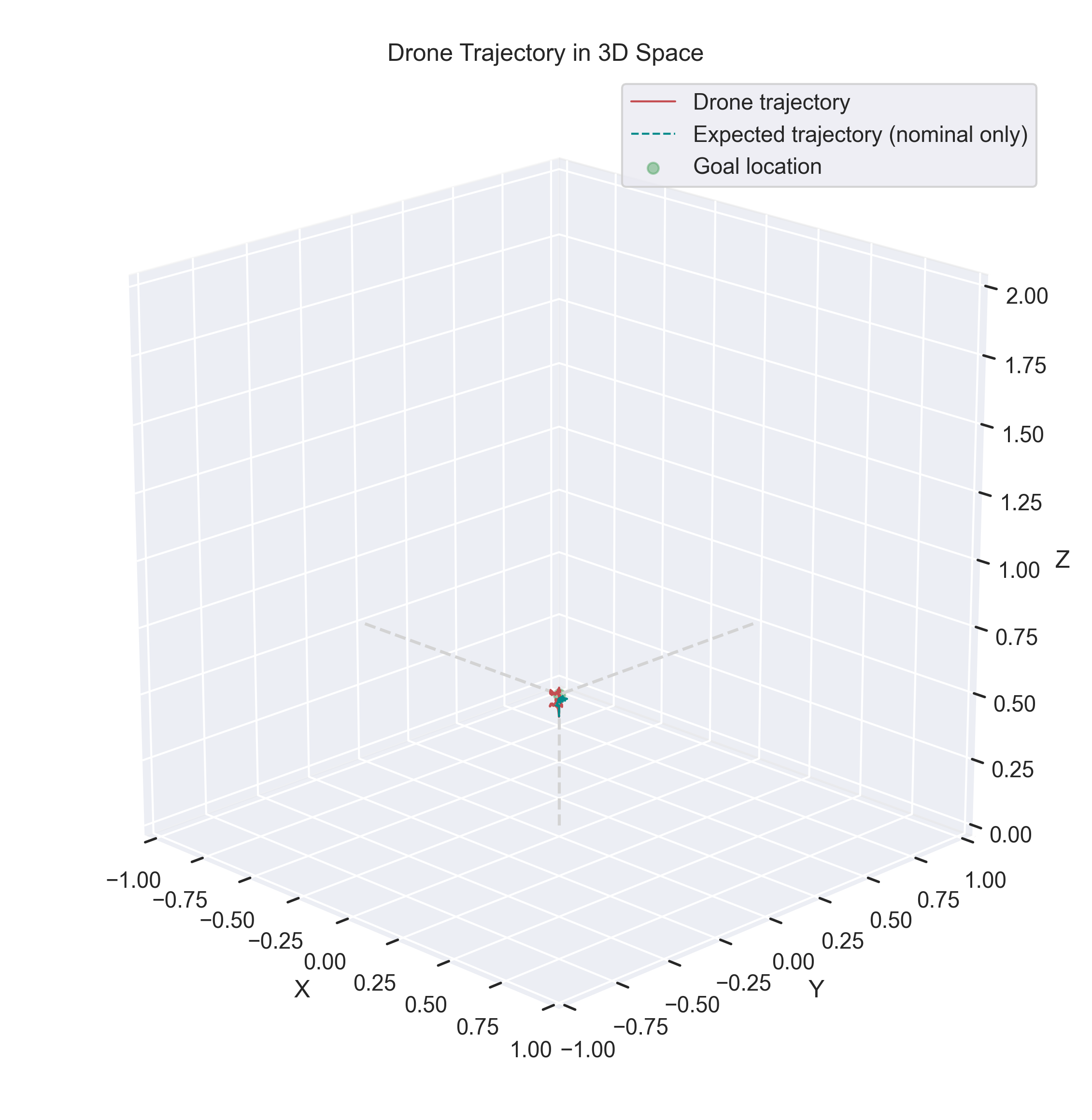}
        \caption{$P_r = (0.0, 0.0, 0.5)$}
        \label{fig:defender_random-0.0_0.0_0.5}
    \end{subfigure}
    \hfill
    \begin{subfigure}[b]{0.32\textwidth}
        \includegraphics[width=\textwidth]{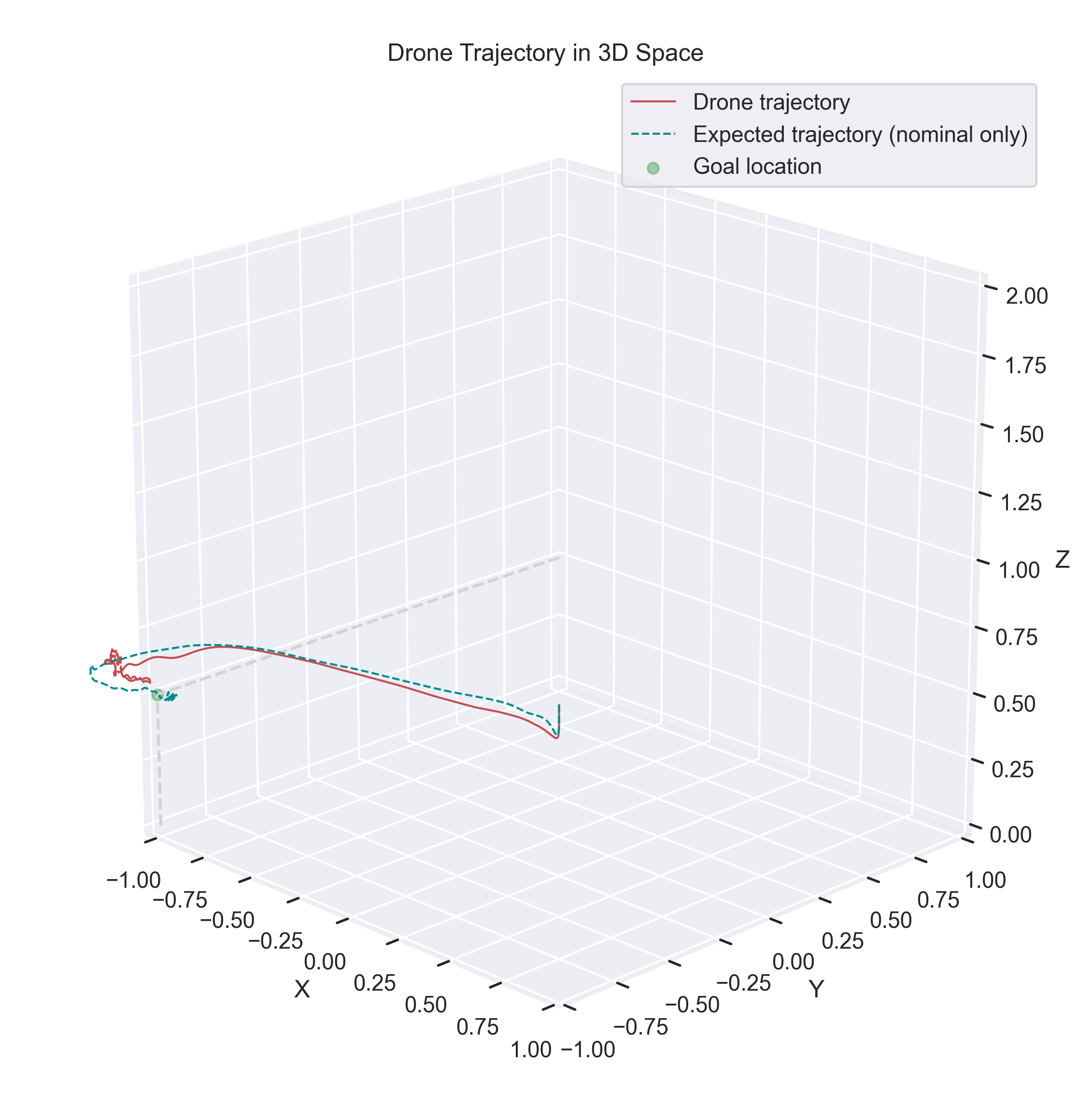}
        \caption{$P_r = (-1.0, -1.0, 0.5)$}
        \label{fig:defender_random--1.0_-1.0_0.5}
    \end{subfigure}

    \caption[Trajectories of the quadrotor under secure countermeasures with random attacks.]{Trajectories of the quadrotor under random attacks without (a,b,c) and with (d,e,f) countermeasures.}
    \label{fig:defender-random}
\end{figure}

Although, as exemplified in Figure \ref{fig:attacker-random}, optimal data injection attacks outperform random attacks, we evaluated our secure controller in mitigating random attacks to further show the effectiveness of the proposed secure control algorithm. Figure \ref{fig:defender-random} verifies the tracking performance under random attacks with (\ref{fig:defender_random-0.85_0.9_1.7}, \ref{fig:defender_random-0.0_0.0_0.5} \ref{fig:defender_random--1.0_-1.0_0.5}) and without (\ref{fig:no_defender_random-0.85_0.9_1.7}, \ref{fig:no_defender_random-0.0_0.0_0.5} \ref{fig:no_defender_random--1.0_-1.0_0.5}) countermeasures. The results demonstrate that the proposed secure control algorithm can mitigate other kinds of attacks just as much. \\

It is important to note that a small tracking error still exists under the secure controller. If the fine-tuning method is used to train deep neural networks continuously, such errors may decrease. However, because malicious adversaries can alter attacks, it is impossible to guarantee that the defender will always be able to mitigate these.

\section{Deployment in Real World} \label{section:deployment}
In this section, we deployed our three learning-based controllers on physical quadrotors, i.e., on real hardware. The aim is to provide real-life settings and observe the performance of our controllers under non-ideal conditions.

\subsection{Agilicious Agent} \label{section:deployement-agilicious}
As demonstrated in \cite{plainenglish_rl_transformer}, learning-based controllers are less predictable than mathematical models. In addition, as described in section \ref{section:agilicious}, Agilicious is an extremely agile and powerful quadrotor with a top speed of 131km/h. Consequently, deploying learning-based controllers on such a quadrotor in real-life conditions requires adequate settings to avoid damaging equipment and harming surrounding people. Therefore, we decided to minimise the potential risks by taking the following approach. \\

\begin{figure}[ht!]
    \centering
    \begin{subfigure}[b]{0.48\textwidth}
        \includegraphics[width=\textwidth]{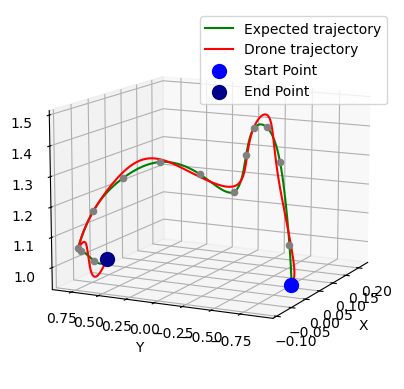}
        \caption{Without attack}
        \label{fig:deploy-agil-nom}
    \end{subfigure}
    \hfill
    \begin{subfigure}[b]{0.48\textwidth}
        \includegraphics[width=\textwidth]{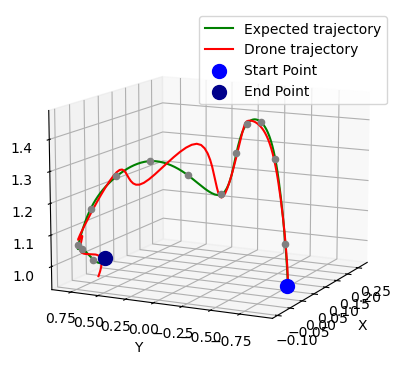}
        \caption{With optimal attack and countermeasure}
        \label{fig:deploy-agil-def}
    \end{subfigure}
    
    \caption[Trajectories of Agilicious quadrotor in planning-based deployment.]{Trajectories of the Agilicious quadrotor with learning-based controllers for planning and MPC for control, without (a) and with (b) optimal attack and secure countermeasure.}
    \label{fig:deploy-agil-sim}
\end{figure}

We deployed our learning-based controllers within the planning trajectory module and used a mathematical model (MPC) for control. As suggested in \cite{plainenglish_rl_transformer}, using the learning-based controller within planning provides a good indication of model performance while limiting deployment risks. By doing so, we can preview the quadrotor's trajectory before launching and, thus, fly Agilicious in our laboratory with enhanced predictability and guaranteed safety. The results of this experiment are displayed in Figure \ref{fig:deploy-agil-sim}. \\

\begin{figure}[ht!]
    \centering
    \begin{subfigure}[b]{0.48\textwidth}
        \includegraphics[width=\textwidth]{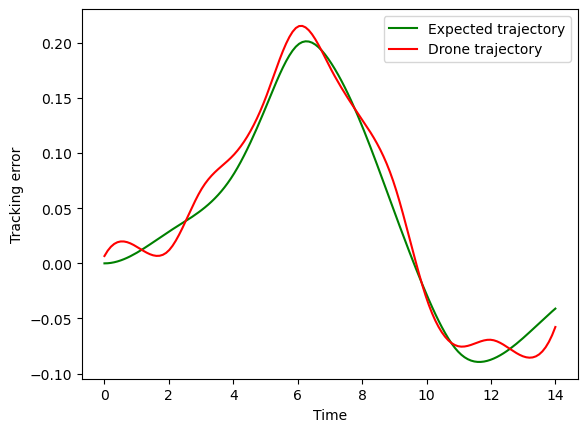}
        \caption{Without attack}
        \label{fig:deploy-agil-nom-dist}
    \end{subfigure}
    \hfill
    \begin{subfigure}[b]{0.48\textwidth}
        \includegraphics[width=\textwidth]{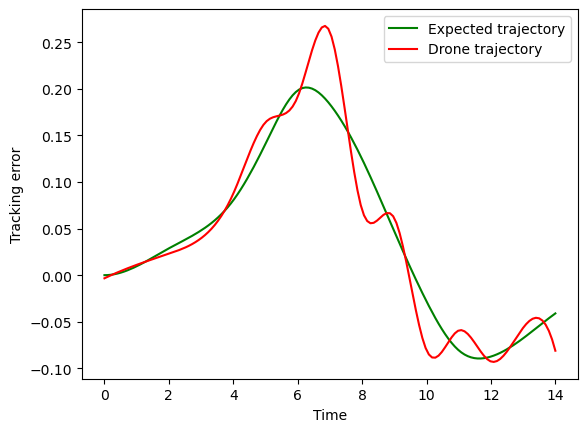}
        \caption{With optimal attack and countermeasure}
        \label{fig:deploy-agil-def-dist}
    \end{subfigure}
    
    \caption[Trajectory errors of Agilicious quadrotor in planning-based deployment.]{Tracking error of the Agilicious quadrotor with learning-based controllers for planning and MPC for control, without (a) and with (b) optimal attack and secure countermeasure. Each experiment is run twenty times and tracking errors are averaged to provide unbiased results. The maximum duration of an experiment is fourteen seconds, i.e. the expected duration if the quadrotor does not crash.}
    \label{fig:deploy-agil-dist}
\end{figure}

As we can observe from Figures \ref{fig:deploy-agil-nom} and \ref{fig:deploy-agil-nom-dist}, the quadrotor under nominal planning control and no attack can successfully fly in real-world settings. 
On the other hand, Figures \ref{fig:deploy-agil-def} and \ref{fig:deploy-agil-def-dist} demonstrate the behaviour of the same quadrotor but under optimal false data injection and secure countermeasure. These results show the ability of our secure countermeasure to counter the optimal attack and recover tracking performance under the same settings. More specifically, the first three seconds are relatively accurate to the expected trajectory. That is because the attack is only launched after three seconds of flying. Subsequently, the drone starts to deviate from the expected trajectory due to the perturbation introduced by the falsified data. Finally, we can see that the quadrotor recovers gradually over time to reach the target location with satisfactory tracking performance, similar to those in simulation (Section \ref{section:eval-def}). \\

It is important to note that, although this experiment tries to further approximate the conditions of flying Agilicious in the real world under learning-based autonomy, we cannot guarantee that such results are entirely accurate to real-world outcomes. In fact, this problem is part of a hot challenge called Simulation-to-Real-World transfer, where researchers explain that simulation conditions can only approximate those from the real world with small but unavoidable error margins.

\subsection{Crazyflie Agent} \label{section:deployement-crazyflie}
Moreover, as described in section \ref{section:crazyflies}, Cazyflies are less hazardous for deployment in non-adapted environments. Therefore, we deployed our learning-based controllers on a Crazyflie quadrotor in real life to compare with previous planning-limited controls. \\

To this end, our learning-based controller is directly integrated within the control module of the quadrotor. That is, no mathematical model is used anymore, and the learning-based controller is no longer limited to planning. This allows to accurately reproduce real-life settings and accounts for non-ideal conditions such as sensor measurement incertitude, motion capture margins, hardware measurement errors, etc.

\begin{figure}[ht!]
    \centering
    \begin{subfigure}[b]{0.48\textwidth}
        \includegraphics[width=\textwidth]{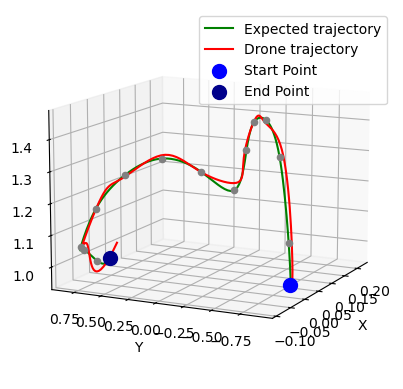}
        \caption{Without attack}
        \label{fig:deploy-cf-nom}
    \end{subfigure}
    \hfill
    \begin{subfigure}[b]{0.48\textwidth}
        \includegraphics[width=\textwidth]{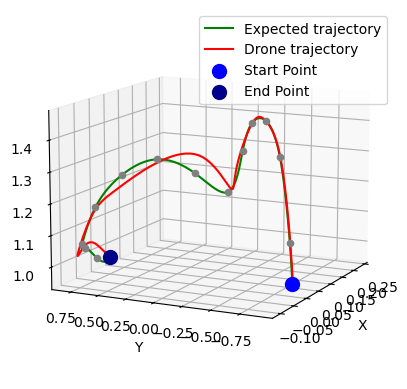}
        \caption{With optimal attack and countermeasure}
        \label{fig:deploy-cf-def}
    \end{subfigure}
    
    \caption[Trajectories of Crazyflie quadrotor in control-based deployment.]{Trajectories of the Crazyflie quadrotor with learning-based controls, without (a) and with (b) optimal attack and secure countermeasure.}
    \label{fig:deploy-cf-sim}
\end{figure}

\begin{figure}[ht!]
    \centering
    \begin{subfigure}[b]{0.48\textwidth}
        \includegraphics[width=\textwidth]{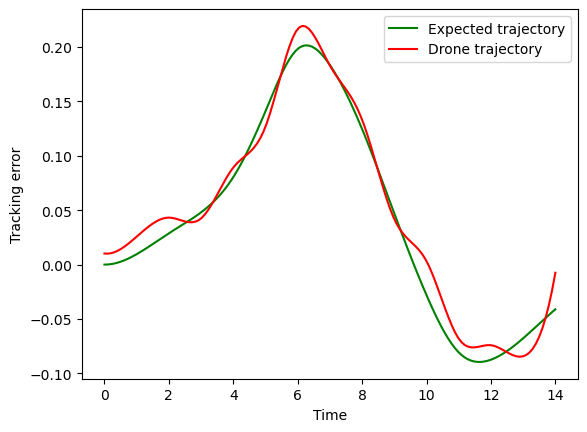}
        \caption{Without attack}
        \label{fig:deploy-cf-nom-dist}
    \end{subfigure}
    \hfill
    \begin{subfigure}[b]{0.48\textwidth}
        \includegraphics[width=\textwidth]{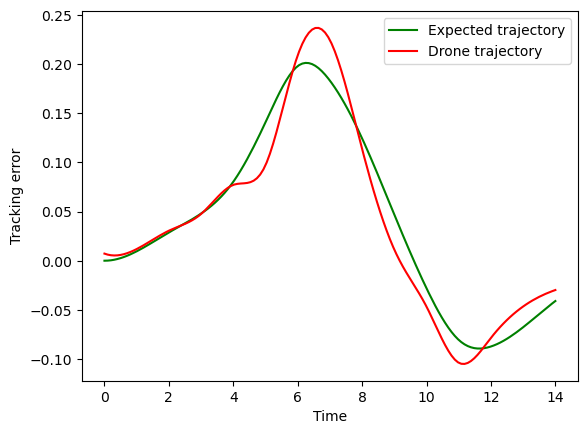}
        \caption{With optimal attack and countermeasure}
        \label{fig:deploy-cf-def-dist}
    \end{subfigure}
    
    \caption[Trajectory errors of Crazyflie quadrotor in control-based deployment.]{Tracking error of the Crazyflie quadrotor with learning-based controls, without (a) and with (b) optimal attack and secure countermeasure. Each experiment is run twenty times and tracking errors are averaged to provide unbiased results. The maximum duration of an experiment is fourteen seconds, i.e. the expected duration if the quadrotor does not crash.}
    \label{fig:deploy-cf-dist}
\end{figure}

The experimental setup is given in Figure \ref{fig:mocap_system} and studies the same scenario as in the Agilicious deployment described above. Furthermore, the experiments were conducted in a closed environment to reduce gusting and light reflection for the motion capture system. \\

The results of Crazyflie's deployment are displayed in Figures \ref{fig:deploy-cf-sim} and \ref{fig:deploy-cf-dist} as snapshots of the quadrotor's trajectory in real life and tracking error wrt time. 
As we can observe, the results are similar to what was obtained on Agilicious in section \ref{section:deployement-agilicious}. Indeed, although this experiment evaluates the trained controller in autonomous and real-world settings, the results obtained confirm the capabilities of our controllers to transfer into real-world conditions without limitations on the control module. 

\clearpage
\section{Results}

\subsection{Outcomes in Simulation}
The experimental results obtained in subsection \ref{section:eval-sim-nominal} demonstrated the ability of the nominal controller designed in Algorithm \ref{algo:nominal} to optimally fly a quadrotor towards a desired location and hover with minimal fluctuations upon arrival. More specifically, we proved the nominal controller to be efficient in flying over wide amplitudes, high altitudes (a,e) and low altitudes (d,f), as well as lifting (c) and hovering (b). In addition, the learning-based controller could make good use of its dynamics, with trajectories resembling some that professional drone pilots would take.

Furthermore, subsection \ref{section:eval-sim-attack} revealed the effect of injecting false data into the quadrotor's actuators. Specifically, Figure \ref{fig:attacker-random} showed that the optimal false data injections designed in Algorithm \ref{algo:attacker} outperform random attacks at tracking performance deterioration. Figure \ref{fig:attacker-sim} proved the effectiveness of our malicious attack algorithm to disrupt the quadrotor's trajectories and crash the system in all provided experiments, with two major types of attacks: motor failure and maximal boosting.

Finally, subsection \ref{section:eval-def} demonstrated the secure countermeasure's effectiveness in recovering the quadrotor's tracking performance to complete its trajectory. All provided experiments, which previously failed under attacks, were then able to complete successfully thanks to the secure countermeasure designed in Algorithm \ref{algo:defender}. Furthermore, our secure controller also proved to be efficient under random attacks in all experiments.

\subsection{Outcomes in Real World}
Experimental results from subsection \ref{section:deployment} illustrated all three controllers in action within the planning module of the Agilicious quadrotor. Although we were unable to use them within the control module for safety reasons, pairing them with an MPC suggested that all three controllers were indeed effective in close-to-real-world conditions (Figures \ref{fig:deploy-agil-sim} and \ref{fig:deploy-agil-dist}), with results aligned with the simulation outcomes.

Moreover, we successfully deployed our three controllers in real-world settings on the Crazyflie quadrotor, providing a safer approach to real-world deployment. The results obtained were once again aligned with our simulation outcomes and confirmed the capabilities of our controllers to transfer into real-world conditions without limitations on the control module (Figures \ref{fig:deploy-cf-sim} and \ref{fig:deploy-cf-dist}).

\chapter{Conclusion}
\label{cha:conclusion}
This paper offered a thorough investigation into the control systems of autonomous quadrotors, with a particular focus on enhancing robustness against cyber-physical attacks, specifically false data injections. After providing evidence of the necessity of designing secure control schemes for autonomous quadrotor systems, we applied deep reinforcement learning techniques to enhance the security of such systems against cyber threats. This paper has two critical improvements. \\

The first is a proposed nominal controller capable of reaching a target location through optimal trajectories and stabilising the quadrotor with minimal fluctuations upon arrival. Our experimental analysis proved the effectiveness of our nominal controller in doing so with great use of its dynamics over diverse types of trajectories, both in simulation and real-world scenarios. 

The second lies in the design of an attack and a secure controller. Although some secure algorithms have been proposed in the literature, most cannot be applied to underactuated nonlinear complex systems, i.e., quadrotors, or do not commit to preserving a confident level of stability in agile settings. Consequently, the effectiveness of existing results does not apply to quadrotors in the way this paper suggests. 
By taking a deep reinforcement learning approach, an optimal false data injection attack was established to outperform random data injection methods and deteriorate the tracking performance of a quadrotor under nominal controls. Furthermore, it proposed a secure countermeasure framework following the same approach, under which the performance of an attacked quadrotor can be recovered immediately.

The effectiveness of our proposed solutions were evaluated in simulation and real-world scenarios over a wide range of experiments. These revealed a significant enhancement in the quadrotor's robustness to cyber-attacks. By integrating the learning-based secure controller, we ensured the continuous stability and safety of the quadrotor, even when subjected to malicious cyber-attacks. The results highlighted the quadrotor's capability to adapt to different attack scenarios, thus validating our methodology's effectiveness in real-time threat mitigation. \\

Furthermore, our work leveraged the capabilities of the Agilicious quadrotor, a state-of-the-art platform which provides the best size-to-computing-power ratio for agile and autonomous flights. As the first team in the United Kingdom to deploy this quadrotor and implement reinforcement learning on its platform, our work introduced a comprehensive breakdown of this quadrotor, including software designs and hardware alternatives. Additionally, this paper provided a detailed reinforcement-learning framework to train autonomous controllers on Agilicious-based agents to support future research on this quadrotor. Finally, we introduced a new open-source environment that builds upon PyFlyt for future research on Agilicious platforms. These contributions promote easy reproducibility with minimal engineering overhead for future works.

\section{Limitations}
Despite providing significant implications, our investigations encountered several development and deployment limitations. \\

One significant limitation encountered was the inability to deploy the Agilicious quadrotor with controllers integrated outside its planning module. This state-of-the-art quadrotor for autonomous systems required specific deployment conditions that were not available in our laboratory settings. Facilities equipped with more advanced systems, such as bird nets and other specialised infrastructures, might offer the necessary environment for safe deployment. As a result, we were restricted in our ability to fully explore and validate the practical aspects of our proposed solutions under more realistic conditions.

Another critical limitation was the exploration of hyperparameters within our controllers. As we observed during our tuning process, hyperparameters indeed played a crucial role in the training and performance of the learning-based controllers. However, our study could only explore a relatively small hyperparameter space due to resource constraints. This limitation might have prevented us from achieving the optimal configuration that maximizes the performance of our policies. A more extensive hyperparameter search could enhance the robustness and accuracy of the proposed solutions.

\section{Future Works}
Building on the outcomes and insights gained from this study, we proposed several promising directions for future research that could significantly enhance the cyber-security of autonomous quadrotor systems. \\

First, the defender proposed in this paper must be activated upon detecting an attack. Thus, its effectiveness is contingent on the preliminary identification of an attack through a third-party framework. This detection could be achieved using anomaly detection methods or more straightforward mathematical strategies, as indicated in \cite{luo2020interval}. Nevertheless, implementing a continuously active defence system could eliminate the need for such initial detection, offering a more robust solution against cyber threats. This approach would, however, necessitate that the secure controller learns to refrain from modifying initial control commands when the quadrotor is not under attack.

Furthermore, the stability analysis conducted in our experiments is entirely empirical, lacking a mathematical demonstration of the system's stability. Although most current deep reinforcement learning approaches operate on empirical bases, introducing a mathematical proof could affirm stability from a theoretical standpoint. This would confirm that deep reinforcement learning methods are both practically and theoretically sound solutions for addressing cyber threats in quadrotor systems. For reference, \cite{wu2023secure} provided such an analysis on a two-dimensional system.

Looking forward, this research lays a solid foundation for further exploration into the security of autonomous systems. 
The optimal false data injection algorithm developed in this study provides a baseline for future research on exposing significant vulnerabilities in quadrotor systems and developing alternative secure controllers.
Moreover, future studies could expand upon this work by exploring additional types of cyber-attacks, integrating multi-agent systems, or applying the methodologies developed to different classes of autonomous vehicles. \\

In conclusion, this study not only achieved its stated objectives but also significantly advanced the field of autonomous system security through the innovative application of deep reinforcement learning to the secure problem of underactuated nonlinear complex systems. The implications of this project extend beyond academic inquiry, offering practical solutions to some of the most pressing challenges in robotic security.

\prefacesection{Acknowledgements}
I would like to express my deepest gratitude to my supervisor, Dr. Wei Pan, for his invaluable guidance and insightful critiques throughout this project. His expertise has been very helpful in shaping both the direction and success of this research. \\

Special thanks are also due to Dr. Louise Denis and Dr. Wai Pan for granting me access to the robotics laboratory of the faculty. The resources provided, including high-end hardware, equipment, and facilities, have been instrumental in the execution of this project. Their support has enabled me to explore and realize the practical aspects of my research in an environment that is as challenging as it is inspiring.

I am particularly grateful to PhD Rishabh Yadav for his knowledgeable support and hands-on assistance with the mechanical aspects of the project. His expertise in robotics has not only enhanced the quality of my work but also enriched my learning experience throughout the project.

I extend my thanks to the University of Manchester for providing access to the CSF3 server, which has been essential for training all models during the project. 

Lastly, I wish to acknowledge the collaborative efforts of Ribhav Ojha, which have been greatly appreciated. \\

The support and resources provided by all mentioned have not only contributed to this research but have also been vital in accomplishing the educational and professional growth I have experienced during this project.
\afterpreface




\printbibliography
\appendix

\end{document}